\pdfoutput=1
\RequirePackage{fix-cm}
\documentclass[twocolumn]{svjour3}          
\smartqed  
\usepackage{mathptmx}      
\usepackage{multirow}
\usepackage[export]{adjustbox}
\usepackage{latexsym}
\usepackage{ifsym}
\usepackage[toc,acronym,nonumberlist,numberline,shortcuts]{glossaries}

\usepackage[labelfont=bf]{subcaption}   
\DeclareCaptionLabelSeparator{nspace}{\hspace{4mm}} 
\usepackage[labelfont=bf,labelsep=nspace]{caption}
\usepackage{rotating}
\usepackage{placeins}     
\usepackage{wrapfig}    
\usepackage{float}      
\usepackage[authoryear]{natbib}
\usepackage{mathtools}
\usepackage{bm} 
\usepackage{amssymb}
\usepackage{mdwlist}        
\usepackage{paralist}
\usepackage{url}
\usepackage{tabularx}
\usepackage{xcolor}

\usepackage[bookmarks,breaklinks=true]{hyperref}
\hypersetup{colorlinks=true,
allcolors=blue,
pdfpagemode=UseOutlines,
bookmarksdepth=3}
\usepackage{hypcap}
\usepackage[nottoc,numbib]{tocbibind}

\newcommand{\figref}[1]{Fig.~\ref{#1}}

\newcommand{\tabref}[1]{Tab.~\ref{#1}}
\newcommand{\secref}[1]{Sect.~\ref{#1}} 
\newcommand{\eqnref}[1]{Eq.~\ref{#1}} 

\DeclareMathAlphabet{\mathcal}{OMS}{cmsy}{m}{n}

\mathcode`\*="8000
{\catcode`\*\active\gdef*{\cdot}}

\journalname{}
\newacronym{FOR}{f.o.r.}{frame of reference}
\newacronym{GT}{GT}{Ground-Truth}
\newacronym{1D}{1D}{1-Dimensional}
\newacronym{2D}{2D}{2-Dimensional}
\newacronym{3D}{3D}{3-Dimensional}
\newacronym{MSL}{MSL}{Mean Sea Level}
\newacronym{DOF}{DOF}{Degrees of Freedom}
\newacronym{CG}{CG}{Center of Gravity}
\newacronym{NED}{NED}{North-East-Down}
\newacronym{SD}{SD}{Standard Deviation}
\newacronym{UWB}{UWB}{Ultra Wide-Band}
\newacronym{MCS}{MCS}{Motion Capture System}

\newacronym{ER}{ER}{Evolutionary Robotics}
\newacronym{GA}{GA}{Genetic Algorithm}
\newacronym{ACO}{ACO}{Ant Colony Optimization}
\newacronym{PSO}{PSO}{Particle Swarm Optimization}
\newacronym{FSM}{FSM}{Finite State Machine}
\newacronym{PFSM}{PFSM}{Probabilistic Finite State Machine}
\newacronym{NN}{NN}{Neural Network}
\newacronym{RNN}{RNN}{Recurrent Neural Network}
\newacronym{FNN}{FNN}{Feed-forward Neural Network}
\newacronym{BT}{BT}{Behavior Tree}
\newacronym{RL}{RL}{Reinforcement Learning}

\newacronym{LS}{LS}{Least Squares}
\newacronym{RLS}{RLS}{Recursive Least Squares}
\newacronym{KF}{KF}{Kalman Filter}
\newacronym{EKF}{EKF}{Extended Kalman Filter}
\newacronym{UKF}{UKF}{Unscented Kalman Filter}
\newacronym{PF}{PF}{Particle Filter}
\newacronym{IAEKF}{IAEKF}{Iterative Adaptive EKF}
\newacronym{KCF}{KCF}{Kalman Consensus Filter}
\newacronym{LPF}{LPF}{Low-Pass Filter}
\newacronym{BPF}{BPF}{Band-Pass Filter}
\newacronym{HPF}{HPF}{High-Pass Filter}
\newacronym{MAF}{MAF}{Moving Average Filter}
\newacronym{MDS}{MDS}{Multi-Dimensional Scaling}

\newacronym{CRR}{CRR}{Conflict Resolution Rate}
\newacronym{FT}{FT}{Flight Time}
\newacronym{RMSE}{RMSE}{Root Mean Squared Error}
\newacronym{ZMGN}{ZMGN}{Zero-Mean Gaussian Noise}

\newacronym{GNSS}{GNSS}{Global Navigation Satellite System}
\newacronym{IR}{IR}{Infra-Red}
\newacronym{IMU}{IMU}{Inertial Measurement Unit}
\newacronym{SLAM}{SLAM}{Simultaneous Localization and Mapping}

\newacronym{AOA}{AOA}{Angle of Arrival}
\newacronym{TOA}{TOA}{Time of Arrival}
\newacronym{TDOA}{TDOA}{Time Difference of Arrival}
\newacronym{RTOA}{RTOA}{Round-trip Time of Arrival}

\newacronym{WSN}{WSN}{Wireless Sensor Network}
\newacronym{WLAN}{WLAN}{Wireless Local Area Network}
\newacronym{RSS}{RSS}{Received Signal Strength}
\newacronym{RSSI}{RSSI}{Received Signal Strength Indication}
\newacronym{FSL}{FSL}{Free Space Loss}
\newacronym{BLE}{BLE}{Bluetooth Low Energy}
\newacronym{GRPR}{GRPR}{Golden Receiver Power Range}
\newacronym{ISM}{ISM}{Industrial, Scientific and Medical}
\newacronym{AP}{AP}{Access Point}
\newacronym{MAC}{MAC}{Media Access Control}
\newacronym{IoT}{IoT}{Internet of Things}
\newacronym{LD}{LD}{Log-Distance}
\newacronym{LQI}{LQI}{Link Quality Indicator}

\newacronym{SQC}{SQC}{Sum Quadratic Constraint}
\newacronym{RANSAC}{RANSAC}{RANdom SAmpling and Consensus}
\newacronym{RGB}{RGB}{Red-Green-Blue}
\newacronym{LED}{LED}{Light-Emitting Diode}
\newacronym{LoG}{LoG}{Laplacian of Gaussian}
\newacronym{SIFT}{SIFT}{Scale-Invariant Feature Transform}
\newacronym{SURF}{SURF}{Speeded Up Robust Feature}
\newacronym{OF}{OF}{Optical Flow}
\newacronym{FAST}{FAST}{Features from Accelerated Segment Test}
\newacronym{CenSurE}{CenSurE}{Center Surround Extremas for Realtime Feature Detection and Matching}

\newacronym{CC}{CC}{Collision Cone}
\newacronym{VO}{VO}{Velocity Obstacle}
\newacronym{RVO}{RVO}{Reciprocal Velocity Obstacle}
\newacronym{HRVO}{HRVO}{Hybrid Reciprocal Velocity Obstacle}
\newacronym{ORCA}{ORCA}{Optimal Reciprocal Collision Avoidance}
\newacronym{HL}{HL}{Human-Like}
\newacronym{CALU}{CALU}{Collision Avoidance under Localization Uncertainty}
\newacronym{COCALU}{COCALU}{Convex Outline Collision Avoidance under Localization Uncertainty}

\newacronym{ROS}{ROS}{Robotics Operating System}
\newacronym{SIDPAC}{SIDPAC}{System Identification Programs for Aircraft}
\newacronym{STDMA}{STDMA}{Self-Organized Time Division Multiple Access}

\newacronym{UAV}{UAV}{Unmanned Air Vehicle}
\newacronym{MAV}{MAV}{Micro Air Vehicle}

\begin{document}

\title{On-board Range-based Relative Localization for Micro Aerial Vehicles in indoor Leader-Follower Flight}

\titlerunning{On-board Range-based Relative Localization for Micro Aerial Vehicles in indoor Leader-Follower Flight}

\author{Steven van der Helm$^{1}$ \and
    Kimberly N. McGuire$^1$ \and \\
    Mario Coppola$^{1,2}$ \and
    Guido C.H.E. de Croon$^1$
}

\authorrunning{S. van der Helm \and
    K.N. McGuire \and 
    M. Coppola \and 
    G.C.H.E. de Croon}

\institute{S. van der Helm \at
        \email{\href{mailto:stevenhelm@live.nl}{stevenhelm@live.nl}}
       \and
       K.N. McGuire \at
            \email{\href{mailto:k.n.mcguire@tudelft.nl}{k.n.mcguire@tudelft.nl}}            
       \and
       M. Coppola \at
            \email{\href{m.coppola@tudelft.nl}{m.coppola@tudelft.nl}}
       \and
       G.C.H.E de Croon \at 
            \email{\href{g.c.h.e.decroon@tudelft.nl}{g.c.h.e.decroon@tudelft.nl}}\\ \newline
    ${^1}$ Delft University of Technology, Faculty of Aerospace Engineering,  
    \emph{Department of Control and Simulation (Micro Air Vehicle Laboratory)}.
    Kluyverweg 1, 2629HS, Delft, The Netherlands.\\
    ${^2}$ Delft University of Technology, Faculty of Aerospace Engineering, 
    \emph{Department of Space Systems Engineering}.
    Kluyverweg 1, 2629HS, Delft, The Netherlands. \\
}

\date{}

\maketitle

\begin{abstract}
  We present a range-based solution for indoor relative localization by Micro Air Vehicles (MAVs), achieving sufficient accuracy for leader-follower flight.
  Moving forward from previous work, we removed the dependency on a common heading measurement by the MAVs, making the relative localization accuracy independent of magnetometer readings.
  We found that this restricts the relative maneuvers that guarantee observability, and also that higher accuracy range measurements are required to rectify the missing heading information, yet both disadvantages can be tackled.
  Our implementation uses Ultra Wide Band, for both range measurements between MAVs and sharing their velocities, accelerations, yaw rates, and height with each other.
  We used this on real MAVs and performed leader-follower flight in an indoor environment.
  The follower MAVs could follow the leader MAV in close proximity for the entire durations of the flights.
  The followers were autonomous and used only on-board sensors to track and follow the leader.
\end{abstract}


\section{Introduction}
  Swarm robotics offer to make \gls{MAV} applications more robust, flexible, and scalable \citep{Sahin2005,Brambilla2013}.
  These properties pertain to a group's ability to remain operable under loss of individual members and to reconfigure for different missions.
  Furthermore, one can imagine that, through cooperation, a swarm of MAVs could execute tasks faster than any single MAV.
  The envisioned applications of such multi-agent robotic systems are plentiful.
  Examples of interest are: 
  cooperative surveillance and/or mapping \citep{Saska2016,Schwager2009-2,Achtelik2012},
  localization of areas of sensory interest (e.g. chemical plumes) \citep{Hayes2003,Schwager2009},
  the detection of forest fires \citep{Merino2006},
  or search missions in hazardous environments \citep{Beard2003}.
  In order to deploy a team of MAVs for such applications, there are certain behaviors that the MAVs should be capable of, such as collision avoidance \citep{Coppola2016,Roelofsen2015} 
  or leader-follower/ formation flight \citep{Vasarhelyi2014,Cheng2014,Gu2006}.
  These tasks are accomplished by the MAVs through knowledge of the relative location of (at least) the neighboring MAVs in the group, for which several solutions can be found in literature.

  Often used are external systems that provide a global reference frame within which agents can extract their own, and the other MAVs', position.
  One example is \glspl{MCS}
  \citep{Schwager2009,Mulgaonkar2015,Kushleyev2013,Michael2010,Turpin2012,Chiew2015,Hayes2002}.
  These systems provide highly accurate location data, but only within the limited coverage provided by the system.
  Alternatively, \gls{GNSS} can be used to provide similar location data 
  \citep{Gu2006,Saska2016,Vasarhelyi2014,Quintero2013,Hauert2011}.
  Although \gls{GNSS} is widely available, it has relatively low accuracy if compared to \gls{MCS} and therefore large inter-MAV separation is required to guarantee safe flight \citep{Nageli2014}.
  Furthermore, \gls{GNSS} cannot reliably be used indoors due to signal attenuation \citep{Liu2007} and can also be subject to multi-path issues in some urban environments or forests \citep{Nguyen2016}.

  To increase the versatility of the solution, MAVs should thus use on-board sensors to determine the locations of neighboring MAVs.
  Often, vision based methods are employed, such as:
  onboard camera based systems \citep{Nageli2014,Iyer2013,Conroy2014,Roelofsen2015}, or
  infrared sensor systems \citep{Kriegleder2015,Stirling2012,Roberts2012}. 
  A drawback of these systems is that they have a limited field of view. 
  This issue can be tackled by creating constructs with an array of sensors \citep{Roberts2012} or by actively tracking neighboring agents \citep{Nageli2014} to keep them in the field of view.
  The first solution introduces a weight penalty, while the second solution severely limits freedom of motion and scalability as a consequence of the need for active tracking of neighbors.
  Therefore, neither solution is ideal for MAVs; 
  a natively omni-directional sensor would be more advantageous.
  One such sensor is a wireless radio transceiver.

  \cite{Guo2017} recently implemented an \gls{UWB} radio-based system for this.
  Range measurements are fused with displacement information from each MAV to estimate the relative location between MAVs.
  However, their method suggests that each MAV must keep track of their own displacement with respect to an initial launching point.
  If this measurement is obtained through on-board sensors 
  (for example, by integrating velocities) 
  then this measurement can be subject to drift over time.
\begin{figure}[t!]
	\centering
	\def\svgwidth{\linewidth}\footnotesize
	\input{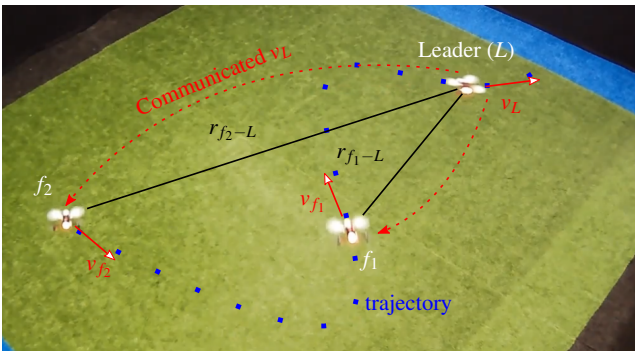}
	\caption{Leader-follower flight with 3 Parrot Bebops, equipped with UWB modules. By estimating and communicating their relative range ($R$) and ego-motion ($v$), follower 1 ($f_1$) and follower 2 ($f_2$) are able to localize the leader and able to follow its trajectory with a certain time delay. }
	\label{fig:frontimage}
\end{figure}
  Alternatively, \cite{Coppola2016} demonstrated a Bluetooth based relative localization method.
  Rather than using displacement information, the velocities of the MAVs, the orientation, and the height were communicated between each other, and the signal strength was used as a range measurement.

  Despite the promising results of range-based solutions, a drawback of the solutions by \cite{Coppola2016} and by \cite{Guo2017} is that the MAVs need knowledge of a common frame orientation.
  This is established by having each MAV measure their heading with respect to North, which would be typically done with magnetometers.
  Magnetometers are notoriously susceptible to the local disturbances in the magnetic field.
  In indoor environments, disturbances upwards of 80$^\circ$ can occur \cite{Afzal2010}. 
  The difficulty of establishing a reliable direction towards North in an indoor environment is a well known problem.
  Solutions are found in the form of complementary filters \citep{Roetenberg2005,Roetenberg2007,Afzal2011,Yuan2015},
  or the use of redundant magnetic sensors to compensate the local disturbances \citep{Afzal2010,Xiaomeng2006}.
  These solutions, however, may be unnecessary for the purpose of relative localization, since a shared reference frame is not theoretically necessary when performing range based relative localization \citep{Zhou2008,Martinelli2005}.

  The main contribution of this paper is an analysis of the consequences of removing the heading dependency in range based relative localization, leading to the development and implementation of a heading-independent relative localization and tracking method that is accurate enough for full on-board indoor leader-follower flight, as shown in \figref{fig:frontimage}.
  The analysis is provided by a formal observability analysis and by performing limit-case simulations.
  Differently from the work of \cite{Zhou2008} and \cite{Martinelli2005}, the analysis also considers the inclusion of acceleration information, since this is commonly known by MAVs from their \gls{IMU}.
  Furthermore, our analysis specifically focuses on the implications of removing a heading dependency on the performance of the relative localization filters and on the relative maneuvers that the agents can perform in order to guarantee that the filter remains observable.
  The observability analysis will show that the task of leader-follower flight is especially difficult with range-based relative localization methods, because it does not allow for the MAVs to fly parallel trajectories.
  We then use the insights gathered for the development and implementation of a heading-independent leader-follower system that we are able to use on-board of autonomous MAVs operating indoors.
  The MAVs rely only on on-board sensors, using \gls{UWB} for both communication and relative ranging.

  The structure of the paper is as follows.
  First, in \secref{sec:Obs}, we compare the theoretical observability of range based relative localization systems both with and without a reliance on a common heading.
  The findings from \secref{sec:Obs} are verified through simulation in \secref{sec:Simulation}, where we also evaluate the difference in performance that can be expected.
  We carry this information forward in \secref{sec:Experiment}, where a heading-independent system is implemented on real MAVs, and where we show the results of our leader-follower experiments.
  The results are further discussed in \secref{sec:Discussion}.
  Finally, the overall conclusions are drawn in \secref{sec:Conclusion}.
  Future work is discussed in \secref{sec:futurework}.

\section{Observability of the Relative Localization Filter}
\label{sec:Obs}

  In this section, an observability analysis is performed that specifically focuses on the practical implications of performing range based relative localization both with and without reliance on a common heading reference.
  The purpose of the eventual relative localization filter is for an MAV
  (say MAV 1)
  to be able to track the position of another MAV
  (say MAV 2).
  Despite our focus on MAVs in particular, the conclusions that follow hold for any general system that can provide the same sensory information.
  Furthermore, the results can be extrapolated to more than two MAVs, as will be demonstrated in \secref{sec:Experiment}.
  For clarity, only MAVs 1 and 2 are considered in the coming analysis.

\subsection{Preliminaries}
\label{sec:prelim}
  We will conduct the analysis by studying the local weak observability of the systems 
  \citep{Hermann1977}. 
  With an analytical test, briefly introduced in the following, local weak observability can be used to extract whether a specific state can be distinguished from other states in its neighborhood.

  Consider a generic non-linear state-space system $\mathbf{\sum}$:
  \begin{align}
  \mathbf{\dot{x}}&=\mathbf{f(x,u)}\\
  \mathbf{y} &= \mathbf{h(x)}
  \end{align}
  The system $\sum$ has state vector 
  $\mathbf{x}=[x_1,x_2,\dots x_n]^\intercal\in\mathbb{R}^n$, 
  an input vector $\mathbf{u}\in\mathbb{R}^l$, 
  and an output vector $\mathbf{y}\in\mathbb{R}^m$.
  The vector function $\mathbf{f(x,u)}$ contains the definitions for the time derivatives of all the states in $\mathbf{x}$ and the vector function $\mathbf{h(x)}$ contains the observation equations for the system.
  The Lie derivatives of this system are:
  \begin{align}
  \mathcal{L}_{\mathbf{f}}^0 \mathbf{h} &= \mathbf{h}\\
  \mathcal{L}_{\mathbf{f}}^1 \mathbf{h} &= \nabla \otimes \mathcal{L}_{\mathbf{f}}^0 \mathbf{h} \cdot \mathbf{f}\\\nonumber
  \vdots \\
  \mathcal{L}_{\mathbf{f}}^i \mathbf{h} &= \nabla \otimes \mathcal{L}_{\mathbf{f}}^{i-1} \mathbf{h}\cdot \mathbf{f}
  \end{align}

  Where $\otimes$ is the Kronecker product and $\nabla$ is the differential operator, defined as 
  $\nabla=[\frac{\partial}{\partial x_1},\frac{\partial}{\partial x_2},\dots,\frac{\partial}{\partial x_n}]$.
  Note that, accordingly, $\nabla \otimes \mathbf{h}$ is equivalent to the Jacobian matrix of $\mathbf{h}$.
  Using these definitions, an observability matrix $\mathcal{O}$ can be constructed, as in \eqnref{eq:Omatrix}.
  \begin{equation}
  \label{eq:Omatrix}
  \mathcal{O} = \left[ {\begin{array}{*{20}{c}}
  {\nabla \otimes \mathcal{L}_{\mathbf{f}}^0 \mathbf{h} }\\
  {\nabla \otimes \mathcal{L}_{\mathbf{f}}^1 \mathbf{h} }\\
  {\vdots}\\
  {\nabla \otimes \mathcal{L}_{\mathbf{f}}^i \mathbf{h} }
  \end{array}} \right], \ i\in \mathbb{N}
  \end{equation}
A system is locally weakly observable if the observability matrix is full rank \citep{Hermann1977}.

\subsection{Reference Frames}
\label{sec:refframes}
\begin{figure}
    \centering
      \def\svgwidth{0.6\linewidth}\footnotesize
      \input{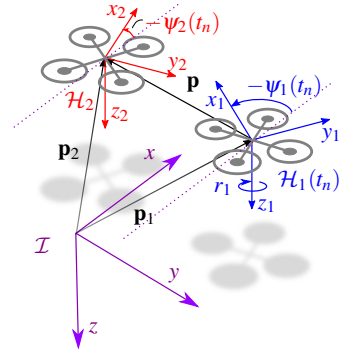}
    \caption{Reference frames used in this paper.
    Frame $\mathcal{I}$ in purple is the earth-fixed North East Down frame (assumed to be inertial).
    Frames $\mathcal{H}_1$ (blue) and $\mathcal{H}_2$ (red) are body fixed reference frames for MAVs 1 and 2, respectively.
    }
    \label{fig:frames}
  \end{figure}

  For the analyses that follow, consider the reference frames schematically depicted in \figref{fig:frames}.
  Denoted by $\mathcal{I}$ is the Earth-fixed North-East-Down (NED) reference frame, which is assumed to be an inertial frame of reference.
  Denoted by $\mathcal{H}_i (i=1,2)$ is a body-fixed reference frame belonging to MAV $i$.
  Its origin is coincident with MAV $i$'s centre of gravity,
  and its location with respect to the $\mathcal{I}$ frame is represented by the vector $\mathbf{p_i}$.
  $\mathcal{H}_i$ is a horizontal frame of reference, such that the z-axis of the $\mathcal{H}_i$ frame remains parallel to that of the $\mathcal{I}$ frame.
  The $\mathcal{H}_i$ frame, however, is rotated with respect to the $\mathcal{I}$ frame about the positive z-axis by an angle $\psi_i$, where $\psi_i$ is the heading that MAV $i$ has with respect to North, also referred to as its yaw angle.
  The rate of change of $\psi_i$ is represented by $r_i$.

  Note that the $\mathcal{H}_i$ frame is different from a typical body-fixed frame $\mathcal{B}_i$, which uses the three Euler angles for roll, pitch, and yaw to represent its orientation with respect to the $\mathcal{I}$ frame.
  The reason for using $\mathcal{H}_i$ rather than $\mathcal{B}_i$ is that it simplifies the kinematic relations without having to impose additional assumptions, such as the roll and pitch angle of the MAV being small.

\subsection{Nonlinear System Description}
  We shall study the case where MAV 1 attempts to estimate the relative position of MAV 2.
  We use $\mathbf{p}$ to denote this relative position, such that $\mathbf{p}=\mathbf{p_2}-\mathbf{p_1}$ (see \figref{fig:frames}).
  Furthermore, let $\mathbf{v_i}$ and $\mathbf{a_i}$ be the linear velocities and accelerations of frame $\mathcal{H}_i$ with respect to frame $\mathcal{I}$ expressed in frame $\mathcal{H}_i$, respectively.
  Finally, let $\Delta\psi$ represent the difference in heading between MAVs $1$ and $2$, such that $\Delta\psi = \psi_2 - \psi_1$.

  From this point on, we shall assume that the MAVs are capable of measuring their own height.
  Since the horizontal plane of $\mathcal{H}_i$ matches the horizontal plane of $\mathcal{I}$, the height can be treated as a decoupled dimension that does not influence the observability, provided that it is measured.
  Therefore, for the sake of brevity, the height is not included in the system description.
  The vectors for the relative position $\mathbf{p}$, the velocity $\mathbf{v_i}$, and the acceleration $\mathbf{a_i}$ can thus be expanded as 2D vectors: 
  $\mathbf{p}^\intercal = [p_x,p_y]^\intercal$, 
  $\mathbf{v_i}=[v_{x,i},v_{y,i}]^\intercal$, 
  $\mathbf{a_i}=[a_{x,i},a_{y,i}]^\intercal$, $i=1,2$.

  The rate of change of $\Delta \psi$ is $\Delta \dot{\psi} = r_2-r_1$.
  Note that the value for $r_i$ is not equal to the yaw rate as would commonly be measured by an on-board rate gyroscope in the body frame $\mathcal{B}_i$. Instead, $r_i$ is expressed as:
  \begin{equation}
  \label{eq:r_itransform}
  r_i = \frac{sin(\phi_i)}{cos(\theta_i)} \tilde{q}_i + \frac{cos(\phi_i)}{cos(\theta_i)} \tilde{r}_i
  \end{equation}

  \noindent where $\tilde{q}_i$ and $\tilde{r}_i$ represent the true pitch and yaw rate as would be measured by a rate gyroscope, and $\phi_i$ and $\theta_i$ are the roll and pitch angles of the MAV.
  However, for the sake of simplicity, $r_i$ will be referred to as the MAV's yaw rate.

  Similarly, $\mathbf{a_i}$, which is the value for the linear acceleration of the $\mathcal{H}_i$ frame expressed in coordinates of the $\mathcal{H}_i$ frame, is not equal to what is measured by the on-board accelerometer.
  Instead, it is equal to:
  \begin{equation}
  \label{eq:a_itransform}
  \mathbf{a_i} = \left[ {\begin{array}{*{20}{c}}
  {c(\theta_i)}&{s(\phi_i)s(\theta_i)} & {c(\phi_i) s(\theta_i)})\\
  {0}&{c(\phi_i)} & {-s(\phi_i)}
  \end{array}} \right]
   \mathbf{s_{i}}
  \end{equation}

  \noindent where $\mathbf{s_{i}}$ is the specific force measured in the body frame $\mathcal{B}_i$ by the accelerometer of MAV $i$.
  Furthermore, $c(\alpha)$ and $s(\alpha)$ represent short hand notation for $cos(\alpha)$ and $sin(\alpha)$, respectively.
  The matrix in this equation consists of the first two rows of the rotation matrix from the $\mathcal{B}_i$ frame to the $\mathcal{H}_i$ frame.

  Following the above, the complete state vector of the system is given by 
  $\mathbf{x} = [\mathbf{p}^\intercal, \Delta\psi, \mathbf{v_1}^\intercal, \mathbf{v_2}^\intercal]^\intercal$,
  and the input vector is
  $\mathbf{u}^\intercal = [\mathbf{a_1}^\intercal,\mathbf{a_2}^\intercal,r_1,r_2]^\intercal$.
  The continuous time state differential equations can be written as:
  \begin{equation}
  \label{eq:ssmodel}
  \mathbf{\dot{x}} = \mathbf{f(x,u)}= \left[ {\begin{array}{*{20}{c}}
  {-\mathbf{v_1}+\mathbf{R}\mathbf{v_2}-\mathbf{S_1 p}}\\
  {r_2-r_1}\\
  {\mathbf{a_1-S_1 v_1}}\\
  {\mathbf{a_2-S_2 v_2}}
  \end{array}} \right]
  \end{equation}

  \noindent where $\mathbf{R}$ is the 2D rotation matrix from frame $\mathcal{H}_2$ to $\mathcal{H}_1$:
  \begin{equation}
  \label{eq:rotmat}
  \mathbf{R} = \mathbf{R}(\Delta\psi) = \left[ {\begin{array}{*{20}{c}}
  {cos(\Delta\psi)}&{-sin(\Delta\psi)}\\
  {sin(\Delta\psi)}&{cos(\Delta\psi)}
  \end{array}} \right]
  \end{equation}

  The matrices $\mathbf{S_1}$ and $\mathbf{S_2}$ are the skew-symmetric matrix equivalent of the cross product, adapted to the 2D case.
  The matrix $\mathbf{S_i}$ is equal to:
  \begin{equation}
  \mathbf{S_i} = \mathbf{S_i}(r_i) = \left[ {\begin{array}{*{20}{c}}
  {0}&{-r_i}\\
  {r_i}&{0}
  \end{array}} \right], i = 1,2
  \end{equation}

  The variables $\mathbf{a_i}$ and $r_i$ are \textit{inputs} into the system and MAV 1 must thus have knowledge of these values.
  However, these are typically available from accelerometer and gyroscope data in combination with the appropriate relations given in \eqnref{eq:r_itransform} and \eqnref{eq:a_itransform}.

  Finally, \eqnref{eq:ssmodel} needs to be complemented with an observation model. Apart from the height, which must be measured but is not included in this analysis, the MAVs should be able to measure the relative range between each other, along with their own and the other's velocities.
  Then, the analysis that follows aims to study the difference between the following two scenarios:
    a scenario where the above measurements are the only measurements and
    a scenario where the MAVs are additionally capable of observing each other's headings.
  The situation where the MAVs can observe a heading is referred to as $\sum_A$ and the situation where a heading is not observed is referred to as $\sum_B$.
  \begin{enumerate*} 
    \item [$\sum_{A}$:]
      The scenario where $\psi_1$ and $\psi_2$ are observed is equivalent to $\Delta\psi$ (the difference in headings) being observed.
      Therefore, for $\sum_A$, the observation model is:
      \begin{equation}
        \label{eq:obsmodel1}
        \mathbf{ y_A} = \mathbf{h_A(x)} = \left[ {\begin{array}{*{20}{c}}
        {h_{A1}(\mathbf{x})}\\
        {h_{A1}(\mathbf{x})}\\
        {\mathbf{h_{A3}}(\mathbf{x})}\\
        {\mathbf{h_{A4}}(\mathbf{x})}
        \end{array}} \right] =
        \left[ {\begin{array}{*{20}{c}}
        {\frac{1}{2}\mathbf{p}^\intercal \mathbf{p}}\\
        {\Delta\psi}\\
        {\mathbf{v_1}}\\
        {\mathbf{v_2}}
        \end{array}} \right]
      \end{equation}

      Note that the observation equation $h_{A1}(\mathbf{x})$ is slightly modified with regards to the previously mentioned measurements.
      Rather than observing the range between the two MAVs (i.e. $||\mathbf{p}||_2$), half the squared range is observed
      (i.e. $\frac{1}{2}\mathbf{p}^\intercal \mathbf{p}$).
      This change makes the observability analysis more convenient without affecting its result.
      Both $||\mathbf{p}||_2$ and $\frac{1}{2}\mathbf{p}^\intercal \mathbf{p}$ contain the same information as far as observability of the system is concerned \citep{Zhou2008}.

    \item [$\sum_{B}$:]
      In this case, the headings of the MAVs are not measured, and it is thus not not possible to observe the difference in heading $\Delta\psi$ directly.
      For $\sum_B$, the observation model is:
      \begin{equation}
      \label{eq:obsmodel2}
      \mathbf {y_B} = \mathbf{h_B(x)} = \left[ {\begin{array}{*{20}{c}}
      {h_{B1}(\mathbf{x})}\\
      {\mathbf{h_{B2}}(\mathbf{x})}\\
      {\mathbf{h_{B3}}(\mathbf{x})}
      \end{array}} \right] =
      \left[ {\begin{array}{*{20}{c}}
      {\frac{1}{2}\mathbf{p}^\intercal \mathbf{p}}\\
      {\mathbf{v_1}}\\
      {\mathbf{v_2}}
      \end{array}} \right]
      \end{equation}
    \end{enumerate*}

  	The effect of the difference in the observation equations is studied in the following sections.

\subsection{Observability Analysis with a Common Heading Reference}

For system $\sum_A$, which uses the observation model from \eqnref{eq:obsmodel1}, the first entry in the observability matrix is equal to:
\begin{align}
\label{eq:Omatterm11}
  \nonumber
  \nabla \otimes \mathcal{L}_{\mathbf{f}}^0 \mathbf{h}_A = \nabla \otimes \mathbf{h}_A &= \left[ {\begin{array}{*{20}{c}}
  {\mathbf{p}^\intercal}&{0}&{\mathbf{0}_{1\mathtt{x}2}}&{\mathbf{0}_{1\mathtt{x}2}}\\
  {\mathbf{0}_{1\mathtt{x}2}} & {1} & {\mathbf{0}_{1\mathtt{x}2}} & {\mathbf{0}_{1\mathtt{x}2}}\\
  {\mathbf{0}_{2\mathtt{x}2}}&{\mathbf{0}_{2\mathtt{x}1}}&{\mathbf{I}_{2\mathtt{x}2}}&{\mathbf{0}_{2\mathtt{x}2}}\\
  {\mathbf{0}_{2\mathtt{x}2}}&{\mathbf{0}_{2\mathtt{x}1}}&{\mathbf{0}_{2\mathtt{x}2}}&{\mathbf{I}_{2\mathtt{x}2}}
  \end{array}} \right] \\
  &= \left[ {\begin{array}{*{20}{c}}
  {\mathbf{p}^\intercal}&{\mathbf{0}_{1\mathtt{x}5}}\\
  {\mathbf{0}_{5\mathtt{x}2}}&{\mathbf{I}_{5\mathtt{x}5}}
  \end{array}} \right]
\end{align} 

\noindent where $\mathbf{I}_{n\mathtt{x}n}$ represents an identity matrix of size $n\mathtt{x} n$ and $\mathbf{0}_{m\mathtt{x}n}$ represents a null matrix of size $m\mathtt{x}n$.
We can already deduce simplifying information from \eqnref{eq:Omatterm11} that will aid the subsequent analysis.
First, note that, for the higher order terms in the observability matrix, the last 5 columns do not contribute to increasing its rank, because these columns are populated with an identity matrix.
Furthermore, these higher order terms in the observation matrix
(corresponding to the observations of $\Delta\psi$, $\mathbf{v_1}$, and $\mathbf{v_2}$) 
only have terms in those last 5 columns because none of the higher order Lie derivatives corresponding to those observations depend on the state $\mathbf{p}$.
For this reason, these need not be computed and we can thus omit them for brevity.
The remainder of this analysis considers only the terms corresponding to observation $h_{A1}(\mathbf{x})=\frac{1}{2}\mathbf{p}^\intercal\mathbf{p}$. 

The first order Lie derivative corresponding to the observation $h_{A1}(\mathbf{x})=\frac{1}{2}\mathbf{p}^\intercal\mathbf{p}$ is equal to:
\begin{equation}
\label{eq:Lieder1}
\mathcal{L}_\mathbf{f}^1 h_{A1} = \mathbf{p}^\intercal (-\mathbf{v_1}+\mathbf{R} \mathbf{v_2}-\mathbf{S_1}\mathbf{p})
\end{equation}
Next, remembering that $\mathbf{S_1}$ is a skew symmetric matrix, such that $\mathbf{S_1} + \mathbf{S_1}^\intercal = \mathbf{0}_{2\mathtt{x}2}$, the following identity is obtained:
\begin{equation}
\frac{\partial \mathbf{p}^\intercal\mathbf{S_i} \mathbf{p}}{\partial\mathbf{p}} = \mathbf{p}^\intercal (\mathbf{S_i}+\mathbf{S_i}^\intercal) = \mathbf{p}^\intercal (\mathbf{0}_{2\mathtt{x}2}) = \mathbf{0}_{1\mathtt{x}2}
\end{equation}
Using this identity, it is can be verified that the second term in the observation matrix corresponding to $h_{A1}(\mathbf{x})$ is:
\begin{equation}
\label{eq:Omatterm2}
\nabla \mathcal{L}_\mathbf{f}^1 h_{A1}=\left[ {\begin{array}{*{20}{c}}
{-\mathbf{v_1}+\mathbf{R}\mathbf{v_2}} \\
{\mathbf{p}^\intercal \frac{\partial \mathbf{R}}{\partial \Delta\psi} \mathbf{v_2}}\\
{-\mathbf{p}}\\
{\mathbf{R}^\intercal \mathbf{p} }
\end{array}} \right]^\intercal
\end{equation}

At this point, it would be possible to continue calculating higher order terms for the observability matrix, but in practice this is not necessary.
The first term of the observability matrix as shown in \eqnref{eq:Omatterm11}  already presents a matrix of rank 6.
Since the state is of size 7, this means that only 1 more linearly independent row needs to be added to the observability matrix to provide local weak observability of the system.
Furthermore, it is of practical interest to study the scenarios in which the system is locally weakly observable with a minimum amount of Lie derivatives involved in the analysis.
This is due to the fact that in practice all signals are noisy, and differentiation of a noisy signal inevitably leads to increasingly noisy signals.
It will be demonstrated that the terms presented in \eqnref{eq:Omatterm2} are sufficient, under certain conditions, to make the observability matrix full rank.

As mentioned, \eqnref{eq:Omatterm11} already shows that the last five columns of the observability matrix are no longer of interest to increase its rank. Furthermore, only the observation of $h_{A1}(\mathbf{x})=\frac{1}{2}\mathbf{p}^\intercal\mathbf{p}$ provides non-zero terms in the first two columns of the observability matrix. Therefore, the following matrix can be constructed by collecting the terms of the first two columns in the observation matrix belonging to observation $h_{A1}(\mathbf{x})$:
\begin{equation}
\label{eq:Opart1}
\mathbf{M_A} = \left[ {\begin{array}{*{20}{c}}
{\mathbf{p}^\intercal}\\
{-\mathbf{v_1}^\intercal+\mathbf{v_2}^\intercal \mathbf{R}^\intercal}
\end{array}} \right]
\end{equation} 
\noindent where the first term is from the zeroth order Lie derivative (see \eqnref{eq:Omatterm11}) and the second term from the first order Lie derivative (see \eqnref{eq:Omatterm2}).
The system is thus observable with a minimum amount of Lie derivatives if the matrix given by \eqnref{eq:Opart1} has two linearly independent rows. 
By the definition of linear independence, this means that the following condition must hold to guarantee local weak observability of the system:
\begin{equation}
\label{eq:sys1cond}
-\mathbf{v_1} + \mathbf{R}\mathbf{v_2} \neq c \mathbf{p}
\end{equation}
\noindent where $c$ is an arbitrary constant.

The condition in \eqnref{eq:sys1cond} essentially tells us that the relative velocity of the two MAVs should not be a multiple of the relative position vector between the two. 
For more practical insight, we can extract more intuitive conditions that must also be met for \eqnref{eq:sys1cond} to hold.
These conditions are:
\begin{itemize*}
\item
    $\displaystyle \mathbf{p} \neq \mathbf{0}_{2\mathtt{x}1}$%
  \hfill\refstepcounter{equation}\label{eq:sys1cond1}\textup{(\theequation)}%
\item
    $\displaystyle \mathbf{v_1} \neq \mathbf{0}_{2\mathtt{x}1} \  \mathtt{or} \   \mathbf{v_2} \neq \mathbf{0}_{2\mathtt{x}1}$%
  \hfill\refstepcounter{equation}\label{eq:sys1cond2}\textup{(\theequation)}%
\item
    $\displaystyle \mathbf{v_1} \neq \mathbf{R}\mathbf{v_2}$ %
  \hfill\refstepcounter{equation}\label{eq:sys1cond3}\textup{(\theequation)}%
\end{itemize*}
The first condition tells us that the $x$ and $y$ coordinates of the relative position of MAV 2 with respect to MAV 1 should not be equal to 0.
In practice, this would only be possible if the MAVs were separated by height, for otherwise their physical dimension would prevent this condition from occurring. 
The second condition tells us that one of the two MAVs needs to be moving to render the filter observable, and that the observability is indifferent to which of the MAVs is moving (hence the \texttt{or} operator).
The third condition tells us that the MAVs should not be moving in parallel at the same speed (note the rotation matrix $\mathbf{R}$ that transforms $\mathbf{v_2}$ to the $\mathcal{H}_1$ frame).

Whilst these three conditions are easier to consider, it should be noted that they form only a subset of the conditions imposed by \eqnref{eq:sys1cond}.
For example, the scenario where MAV 2 is stationary, and MAV 1 flies straight towards MAV 2, does not violate any of these three conditions.
It does, however, violate \eqnref{eq:sys1cond}.
Therefore, the observability of a state and input combination should be checked against the full condition in \eqnref{eq:sys1cond}.

\subsection{Observability Analysis Without a Common Heading Reference}\label{sec:obsernonorth}

  After determining the conditions under which system $\sum_A$ is locally weakly observable, we compare it to the system where the heading measurements are no longer present.
  We now consider system $\sum_B$, whose observation equation (\eqnref{eq:obsmodel2}) does not include the state $\Delta\psi$.
  For this system, the first term in the observability matrix is:
  \begin{equation}
  \label{eq:Omatterm21}
  \nabla \otimes \mathcal{L}_{\mathbf{f}}^0 \mathbf{h}_B = \nabla \otimes \mathbf{h}_B = \left[ {\begin{array}{*{20}{c}}
  {\mathbf{p}^\intercal}&{0}&{\mathbf{0}_{1\mathtt{x}2}}&{\mathbf{0}_{1\mathtt{x}2}}\\
  {\mathbf{0}_{2\mathtt{x}2}}&{\mathbf{0}_{2\mathtt{x}1}}&{\mathbf{I}_{2\mathtt{x}2}}&{\mathbf{0}_{2\mathtt{x}2}}\\
  {\mathbf{0}_{2\mathtt{x}2}}&{\mathbf{0}_{2\mathtt{x}1}}&{\mathbf{0}_{2\mathtt{x}2}}&{\mathbf{I}_{2\mathtt{x}2}}
  \end{array}} \right]
  \end{equation} 

  \eqnref{eq:Omatterm21} is very similar to \eqnref{eq:Omatterm11}, but with the important difference that the row corresponding to the observation of $\Delta\psi$ is null. 
  Consequently, the matrix is only of rank 5, rather than rank 6.
  Since the state size is still 7, a minimum of two more independent rows must be added to the observability matrix to make the system locally weakly observable.
  Once again only the terms corresponding to the observation $h_{B1}(\mathbf{x})=\frac{1}{2}\mathbf{p}^\intercal\mathbf{p}$ have terms that could increase the rank of the observability matrix.
  This means that this time a minimum of two more Lie derivatives must be calculated.

  It can be verified that the first derivative $\mathcal{L}_\mathbf{f}^1 h_{B1}$, and thus its state-derivative $\nabla \mathcal{L}_\mathbf{f}^1 h_{B1}$, are exactly the same as for $\sum_A$.
  Therefore, these need not be calculated anymore and are given by \eqnref{eq:Lieder1} and \eqnref{eq:Omatterm2}, respectively.
  The second order Lie derivative is equal to:
  \begin{align}
  \label{eq:Lf2ne}
    \mathcal{L}_\mathbf{f}^2 h_{B1} &= (-\mathbf{v_1}^\intercal + \mathbf{v_2}^\intercal \mathbf{R}^\intercal)(-\mathbf{v_1}+\mathbf{R} \mathbf{v_2} - \mathbf{S_1} \mathbf{p}) \\\nonumber
    &  + \mathbf{p}^\intercal \frac{\partial \mathbf{R}}{\partial \Delta\psi} \mathbf{v_2}(r_2-r_1) -\mathbf{p}^\intercal(\mathbf{a_1} - \mathbf{S_1}\mathbf{v_1}) \\\nonumber
    &+ \mathbf{p}^\intercal \mathbf{R}^\intercal (\mathbf{a_2} - \mathbf{S_2}\mathbf{v_2})
  \end{align}

  Some terms in \eqnref{eq:Lf2ne} can be seen to drop out when the equation is expanded.
  For example, the yaw rate of MAV 1 ($r_1$) cancels out completely.
  Therefore, \eqnref{eq:Lf2ne} reduces to:
  \begin{align}
  \mathcal{L}_\mathbf{f}^2 h_{B1} &= \mathbf{v_1}^\intercal \mathbf{v_1} + \mathbf{v_2}^\intercal \mathbf{v_2} - 2\mathbf{v_1}^\intercal \mathbf{R} \mathbf{v_2} + \mathbf{p}^\intercal \frac{\partial \mathbf{R}}{\partial \Delta\psi} \mathbf{v_2} r_2 \\\nonumber
  & - \mathbf{p}^\intercal \mathbf{a_1}  + \mathbf{p}^\intercal \mathbf{R} \mathbf{a_2} - \mathbf{p}^\intercal \mathbf{R}^\intercal \mathbf{S_2} \mathbf{v_2} 
  \end{align}

  The state derivative of $\mathcal{L}_\mathbf{f}^2 h_{B1}$ can then be shown to be equal to \eqnref{eq:Omatterm23}.
  Once again, note that some terms drop out (this step has been omitted for brevity).
  \begin{equation}
  \label{eq:Omatterm23}
  \nabla \mathcal{L}_\mathbf{f}^2 h_{B1}=\left[ {\begin{array}{*{20}{c}}
  {\mathbf{a_1}+\mathbf{R}\mathbf{a_2} }\\
  {-2 \mathbf{v_1}^\intercal \frac{\partial \mathbf{R}}{\partial \Delta\psi} \mathbf{v_2} + \mathbf{p}^\intercal \frac{\partial \mathbf{R}}{\partial \Delta\psi}\mathbf{a_2}} \\
  {2 \mathbf{v_1}- 2 \mathbf{R}\mathbf{v_2} }\\
  {-2 \mathbf{R}^\intercal \mathbf{v_1} +2 \mathbf{v_2}}
  \end{array}} \right]^\intercal
  \end{equation}

  Just as for $\sum_A$, a part of the observation matrix can be extracted for analysis.
  This time, the first three columns in the observation matrix (as opposed to two) are collected for the observation $h_{B1}(\mathbf{x})=\frac{1}{2}\mathbf{p}^\intercal\mathbf{p}$.
  Also, this time the terms up to and including the second order Lie derivative are minimally needed to obtain a full rank observability matrix.
  The following matrix is obtained:
  \begin{equation}
  \label{eq:Opart2}
  \mathbf{M_B} = \left[ {\begin{array}{*{20}{c}}
  {\mathbf{p}^\intercal}&{0}\\
  {-\mathbf{v_1}^\intercal+\mathbf{v_2}^\intercal \mathbf{R}^\intercal}&{\mathbf{p}^\intercal \frac{\partial \mathbf{R}}{\partial \Delta\psi}\mathbf{v_2}}\\
  {-\mathbf{a_1}^\intercal+\mathbf{a_2}^\intercal \mathbf{R}^\intercal}&{-2 \mathbf{v_1}^\intercal \frac{\partial \mathbf{R}}{\partial \Delta\psi} \mathbf{v_2} + \mathbf{p}^\intercal \frac{\partial \mathbf{R}}{\partial \Delta\psi} \mathbf{a_2}}
  \end{array}} \right]
  \end{equation} 

  In this case, obtaining the conditions for which this is a full rank matrix is less obvious due to the plethora of terms.
  Rather than directly demonstrating linear independence of the three rows in  \eqnref{eq:Opart2}, the determinant $\mathbf{|M_B|}$ may be computed and demonstrated to be non-zero.
  This is done as follows.
  Recall that $\mathbf{p}^\intercal = [p_x,p_y]$.
  Furthermore, suppose $-\mathbf{v_1}^\intercal+\mathbf{v_2}^\intercal \mathbf{R}^\intercal = [a,b]$ and $-\mathbf{a_1}^\intercal+\mathbf{a_2}^\intercal \mathbf{R}^\intercal = [c,d]$.
  Then, matrix $\mathbf{M_B}$ can be written as:
  \begin{equation}
  \label{eq:Opartsimp}
  \mathbf{M_B} = \left[ {\begin{array}{*{20}{c}}
  {p_x} & {p_y} &{0}\\
  {a} & {b} &{\mathbf{p}^\intercal \frac{\partial \mathbf{R}}{\partial \Delta\psi} \mathbf{v_2}}\\
  {c} & {d} &{-2 \mathbf{v_1}^\intercal \frac{\partial \mathbf{R}}{\partial \Delta\psi} \mathbf{v_2} + \mathbf{p}^\intercal \frac{\partial \mathbf{R}}{\partial \Delta\psi}\mathbf{a_2}}
  \end{array}} \right]
  \end{equation} 
  The determinant of $\mathbf{M_B}$ can be computed using a cofactor expansion along the last column of $\mathbf{M_B}$. This results in:
  \begin{align}
  |\mathbf{M_B}| &= -\mathbf{p}^\intercal \frac{\partial \mathbf{R}}{\partial \Delta\psi} \mathbf{v_2} (d p_x-c p_y) + \\\nonumber
  & (-2 \mathbf{v_1}^\intercal \frac{\partial \mathbf{R}}{\partial \Delta\psi} \mathbf{v_2} + \mathbf{p}^\intercal \frac{\partial \mathbf{R}}{\partial \Delta\psi} \mathbf{a_2}) (b p_x-a p_y)
  \end{align}
  Now, the following identity can be used:
  \begin{align}
  \label{eq:Amult}
  b p_x-a p_y = \left[ {\begin{array}{*{20}{c}}
  {a}&{b}
  \end{array}} \right]\left[ {\begin{array}{*{20}{c}}
  {-p_y}\\
  {p_x}
  \end{array}} \right] =  \left[ {\begin{array}{*{20}{c}}
  {a}&{b}
  \end{array}} \right]
  \mathbf{A}
  \left[ {\begin{array}{*{20}{c}}
  {p_x}\\
  {p_y}
  \end{array}} \right],
  \end{align}
  where $\mathbf{A}=\left[ {\begin{array}{*{20}{c}}
  {0}&{-1}\\
  {1}&{0}
  \end{array}} \right]$.

  Substituting back the original expressions for $[a,b]$, $[c,d]$, and $[p_x,p_y]$, the determinant of $\mathbf{M_B}$ becomes:
  \begin{align}
  \label{eq:mb_orig}
  \nonumber
  |\mathbf{M_B}| &= -\mathbf{p}^\intercal \frac{\partial \mathbf{R}}{\partial \Delta\psi}\mathbf{v_2} (-\mathbf{a_1}^\intercal+\mathbf{a_2}^\intercal \mathbf{R}^\intercal) \mathbf{A} \mathbf{p} +\\
  &(-2 \mathbf{v_1}^\intercal \frac{\partial \mathbf{R}}{\partial \Delta\psi} \mathbf{v_2} + \mathbf{p}^\intercal \frac{\partial \mathbf{R}}{\partial \Delta\psi} \mathbf{a_2}) (-\mathbf{v_1}^\intercal+\mathbf{v_2}^\intercal \mathbf{R}^\intercal)\mathbf{A} \mathbf{p}
  \end{align}

  This can be simplified and written as:
  \begin{align}
  \nonumber
  |\mathbf{M_B}| &=\left[ \mathbf{p}^\intercal \frac{\partial \mathbf{R}}{\partial \Delta\psi} \left( {-\mathbf{a_2} \mathbf{v_1}^\intercal + \mathbf{v_2} \mathbf{a_1}^\intercal  }  \right)\right. + \\
  \label{eq:sufobs}
  &2 \left. \mathbf{v_1}^\intercal \frac{\partial \mathbf{R}}{\partial \Delta\psi} \left( {\mathbf{v_2} \mathbf{v_1}^\intercal - \mathbf{v_2} \mathbf{v_2}^\intercal \mathbf{R}^\intercal}  \right) \right] \mathbf{A} \mathbf{p}
  \end{align}

  This system is thus locally weakly observable with a minimum amount of Lie derivatives if $\mathbf{|M_B|}$ is non-zero.
  Due to the specific properties of the $\mathbf{A}$ matrix in this determinant
  (see \eqnref{eq:Amult}),
  the following equation must hold to render the determinant $|\mathbf{M_B}|$ non-zero:
  \begin{align}
  \nonumber
  \mathbf{p}^\intercal \frac{\partial \mathbf{R}}{\partial \Delta\psi} \left( {-\mathbf{a_2} \mathbf{v_1}^\intercal + \mathbf{v_2} \mathbf{a_1}^\intercal  }  \right) + \\
  \label{eq:sys2cond}
  2  \mathbf{v_1}^\intercal \frac{\partial \mathbf{R}}{\partial \Delta\psi} \left( {\mathbf{v_2} \mathbf{v_1}^\intercal - \mathbf{v_2} \mathbf{v_2}^\intercal \mathbf{R}^\intercal}  \right)  \neq  k \mathbf{p}^\intercal
  \end{align}
  \noindent where $k$ is an arbitrary constant.

  It is difficult to find an intuitive interpretation for \eqnref{eq:sys2cond}.
  Just as for \eqnref{eq:sys1cond}, we can extract a more intuitive subset of conditions that also definitely must be met for the system to be observable.
  These conditions are:

  \begin{itemize*}
  \item
      $\displaystyle \mathbf{p} \neq \mathbf{0}_{2\mathtt{x}1}$%
    \hfill\refstepcounter{equation}\label{eq:sys2cond1}\textup{(\theequation)}%
  \item
      $\displaystyle (\mathbf{v_1} \neq \mathbf{0}_{2\mathtt{x}1} \ \mathtt{or} \  \mathbf{a_1} \neq \mathbf{0}_{2\mathtt{x}1}) \ \mathtt{and}$%
    \hfill
  \item[] 
      $\displaystyle (\mathbf{v_2} \neq \mathbf{0}_{2\mathtt{x}1} \ \mathtt{or} \ \mathbf{a_2} \neq \mathbf{0}_{2\mathtt{x}1})$%
    \hfill\refstepcounter{equation}\label{eq:sys2cond2}\textup{(\theequation)}%
  \item
      $\displaystyle \mathbf{v_1} \neq s\mathbf{R v_2} \ \mathtt{or} \ (\mathbf{a_1} \neq \mathbf{0}_{2\mathtt{x}1} \ \mathtt{or} \ \mathbf{a_2} \neq \mathbf{0}_{2\mathtt{x}1})$ %
    \hfill\refstepcounter{equation}\label{eq:sys2cond3}\textup{(\theequation)}%
  \end{itemize*}

  \noindent where $s$ an arbitrary constant.

\begin{figure}[t!]
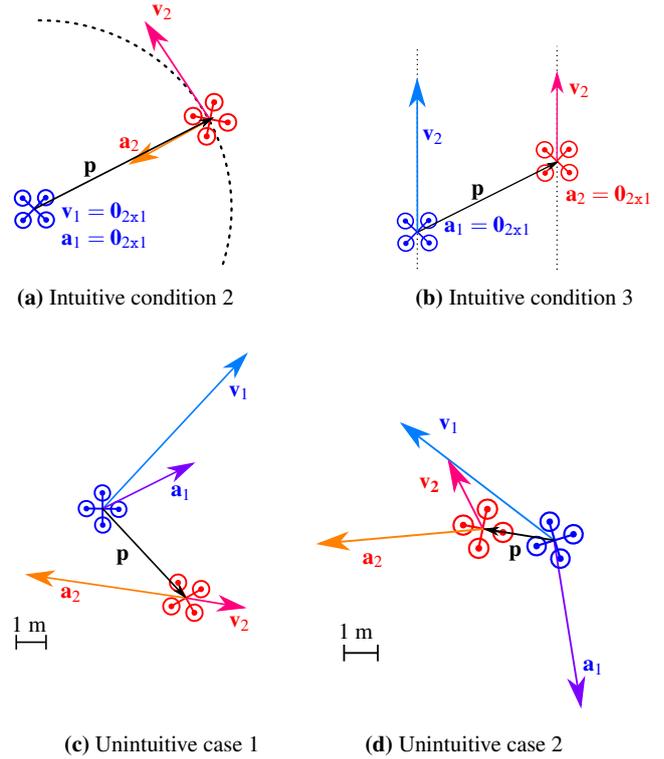

  	
  	\begin{subfigure}[t]{0.35\linewidth}
  		\centering
  		\def\svgwidth{\linewidth}\footnotesize
  		\input{Fig3a.tex}
  		\caption{Intuitive condition 2}
  		\label{fig:vect2i}
  	\end{subfigure} \hfill \begin{subfigure}[t]{0.40\linewidth}
  	\centering
  	\def\svgwidth{\linewidth}\footnotesize
  	\input{Fig3b.tex}
  	\caption{Intuitive condition 3}
  	\label{fig:vect3i}
  \end{subfigure}
  
  \vspace{0.5cm}
  
  \begin{subfigure}[t]{39mm}
  	\centering
  	\def\svgwidth{\linewidth}\footnotesize
  	\input{Fig3c.tex}
  	\caption{Unintuitive case 1}
  	\label{fig:vect13}
  \end{subfigure}
  \begin{subfigure}[t]{39mm}
  	\centering
  	\def\svgwidth{\linewidth}\footnotesize
  	\input{Fig3d.tex}
  	\caption{Unintuitive case 2}
  	\label{fig:vect17}
  \end{subfigure}
  \caption{Representations of four unobservable state and input combinations.
  	The relative position $\mathbf{p}$, the velocities $\mathbf{v_i}$, and the accelerations $\mathbf{a_i}$ of MAVs 1 and 2 are depicted.
  }
  \label{fig:vect1}
\end{figure}

  The first condition tells us that the determinant $|\mathbf{M_B}|$ is zero if the $x$ and $y$ coordinates of the origins of frames $\mathcal{H}_1$ and $\mathcal{H}_2$ coincide.
  This is the same as for $\sum_A$.
  The second condition tells us that both MAVs need to be moving.
  This movement may be either through having a non-zero velocity, or through having a non-zero acceleration (the violation of which is shown in \figref{fig:vect2i}).
  The third condition tells us that the MAVs may not move in parallel, as in \figref{fig:vect3i}, \emph{unless} at least one of the MAVs is also accelerating at the same time.
  Note that this time the MAVs are not allowed to move in parallel regardless of whether they are moving at the same speed or not (notice the scalar multiple $s$).
  By comparison, the equivalent condition for $\sum_A$ only specified that the MAVs may not move in parallel at the same speed.

\begin{figure}[t!]
	
	\begin{subfigure}[t]{39mm}
		\centering
		\includegraphics[width=\columnwidth]{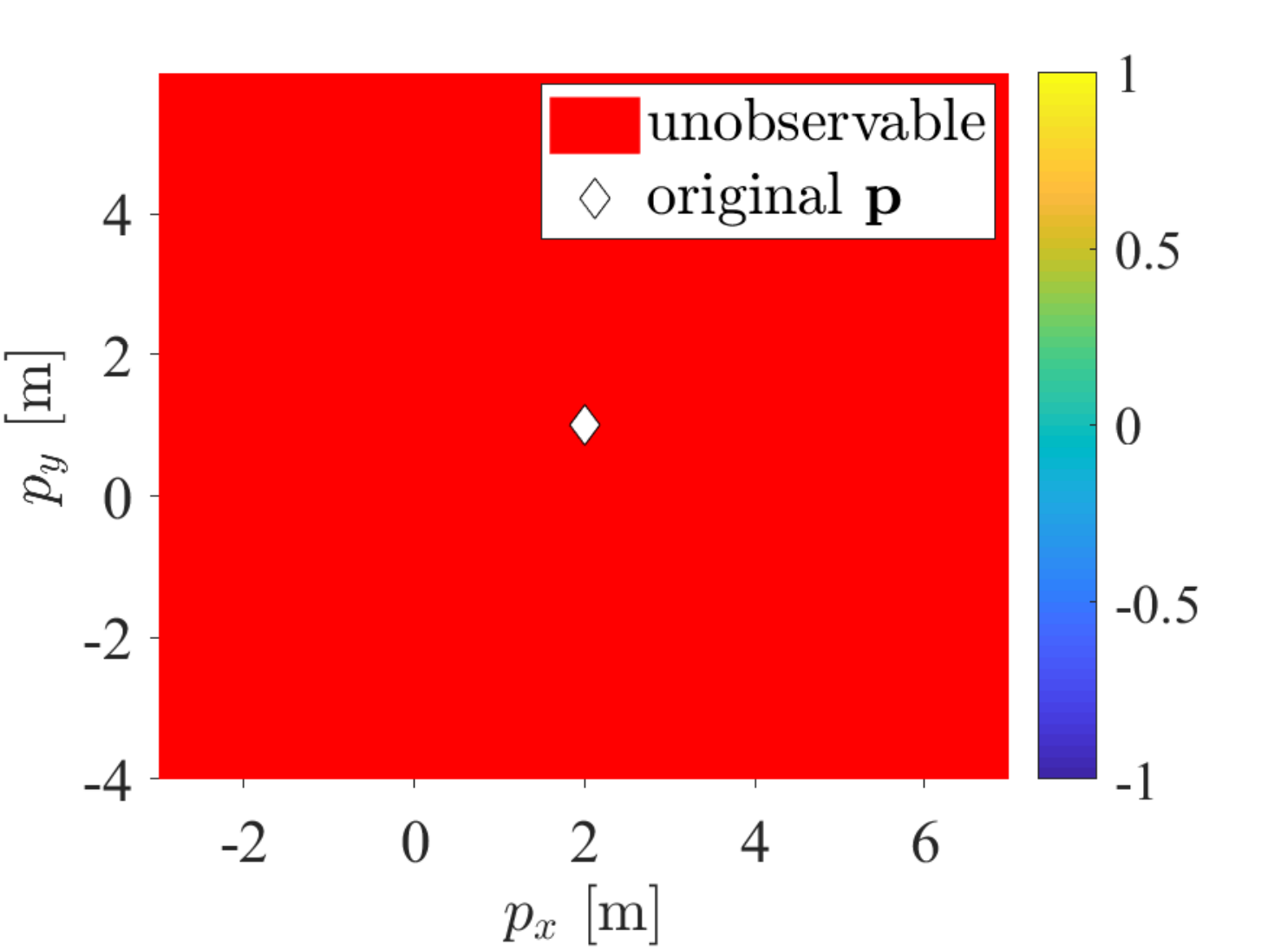}
		\caption{\centering Intuitive condition 2 \newline Fully unobservable}
		\label{fig:heati21}
	\end{subfigure}
	~
	\begin{subfigure}[t]{39mm}
		\centering
		\includegraphics[width=\columnwidth]{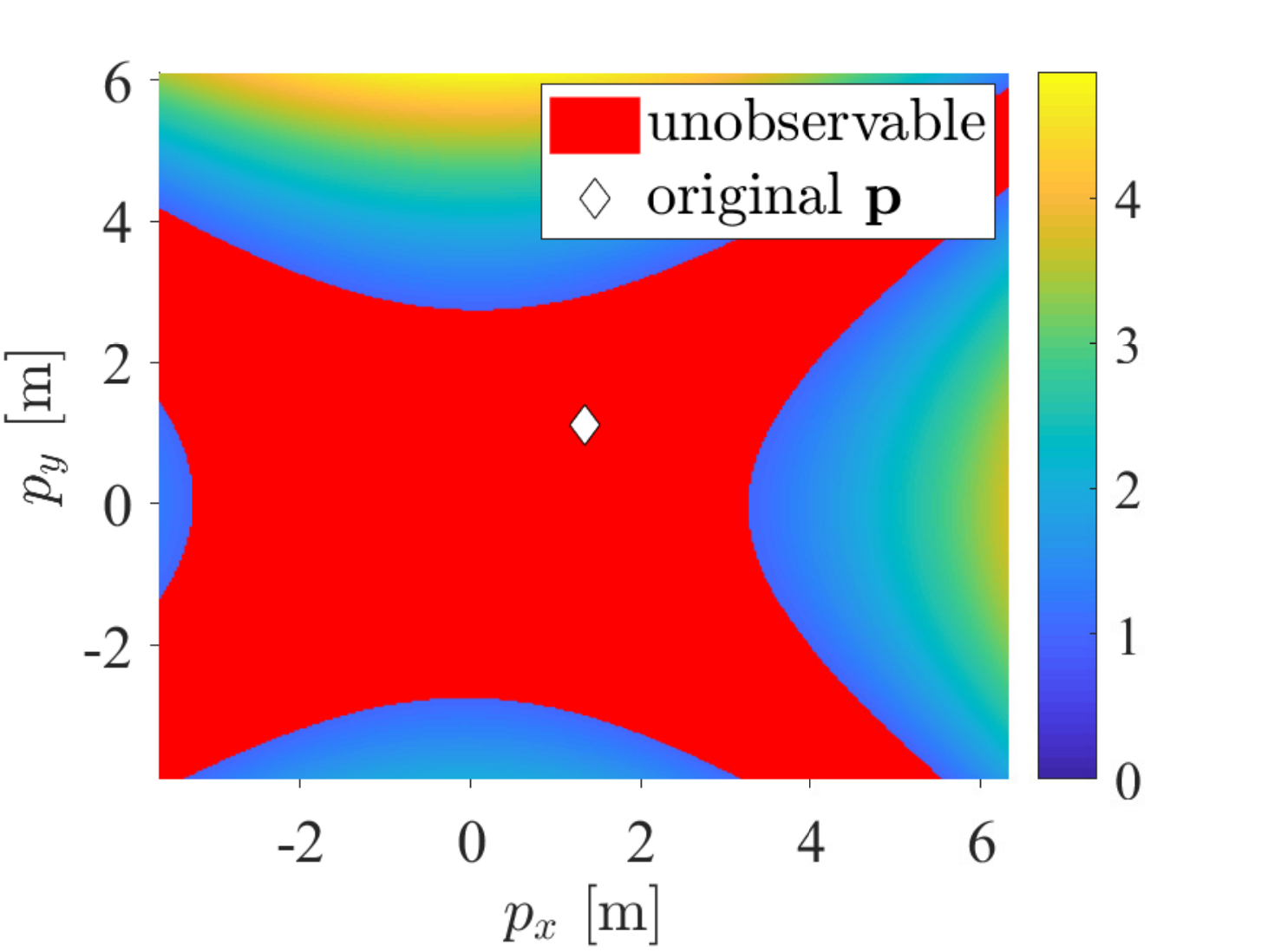}
		\caption{\centering Intuitive condition 2 \newline Partially unobservable}
		\label{fig:heati22}
	\end{subfigure}
	
	\begin{subfigure}[t]{39mm}
		\centering
		\includegraphics[width=\columnwidth]{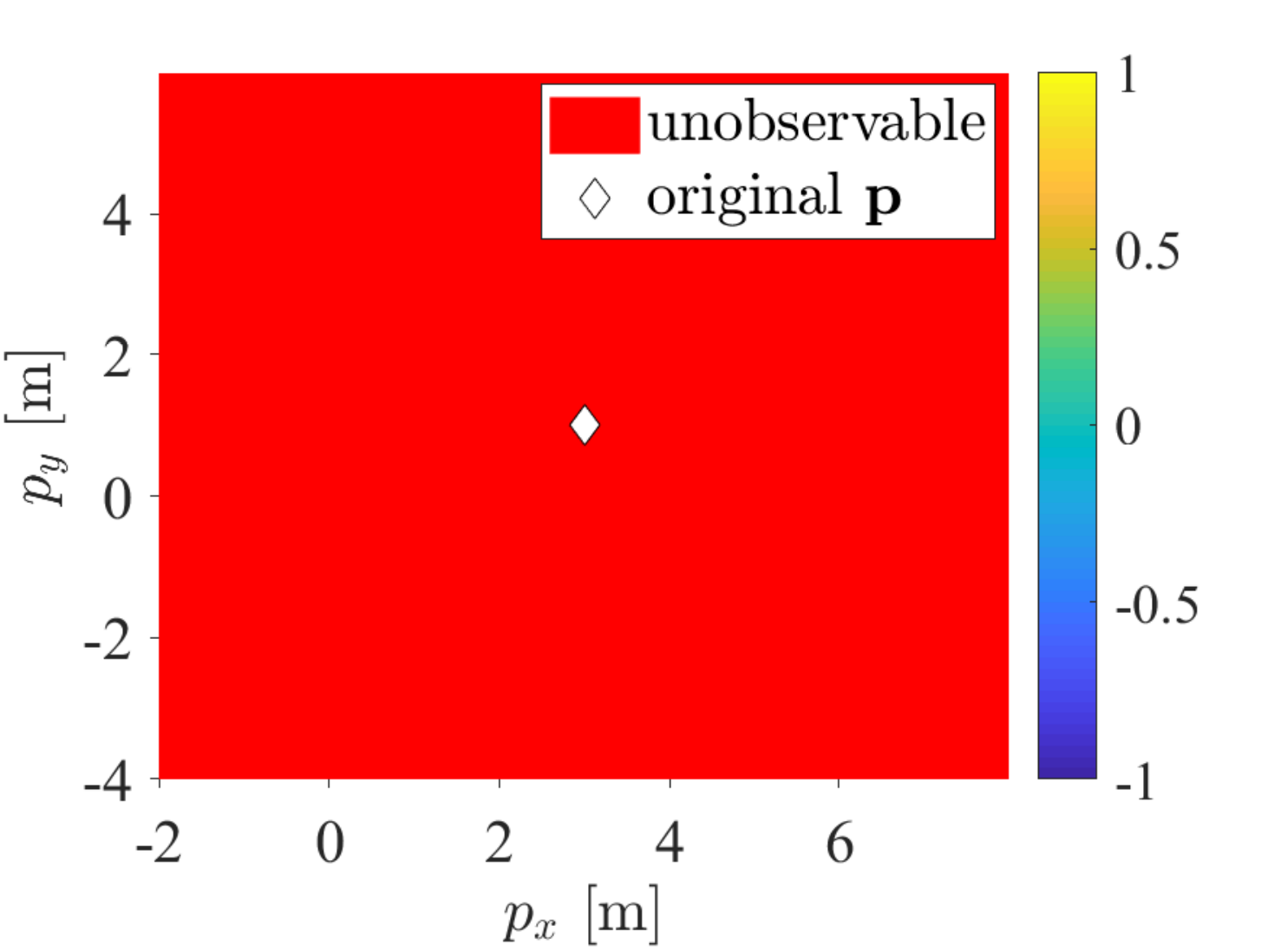}
		\caption{\centering Intuitive condition 3 \newline Fully unobservable}
		\label{fig:heati31}
	\end{subfigure}
	~
	\begin{subfigure}[t]{39mm}
		\centering
		\includegraphics[width=\columnwidth]{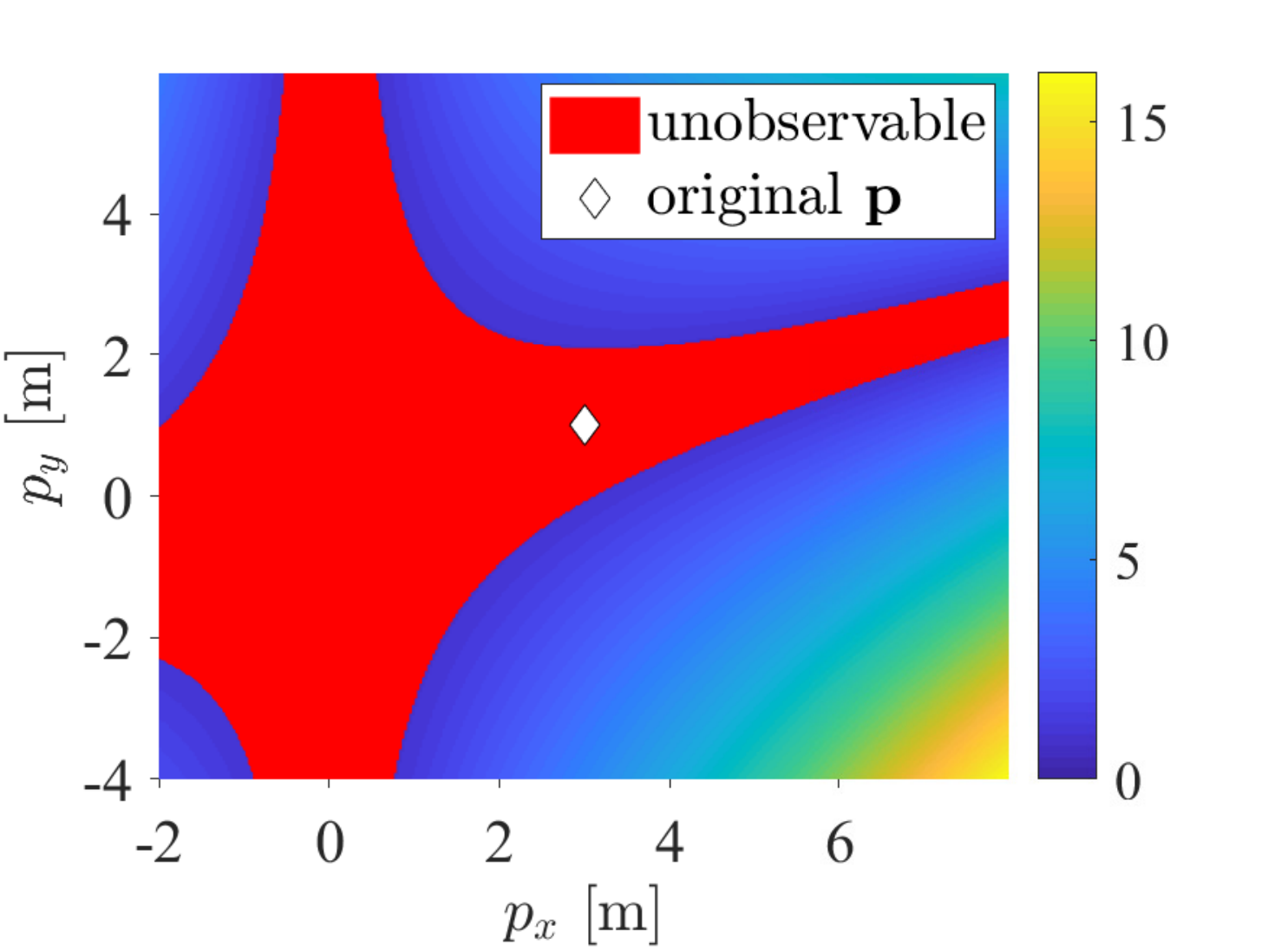}
		\caption{\centering Intuitive condition 3 \newline Partially unobservable}
		\label{fig:heati32}
	\end{subfigure}
	
	\begin{subfigure}[t]{39mm}
		\centering
		\includegraphics[width=\columnwidth]{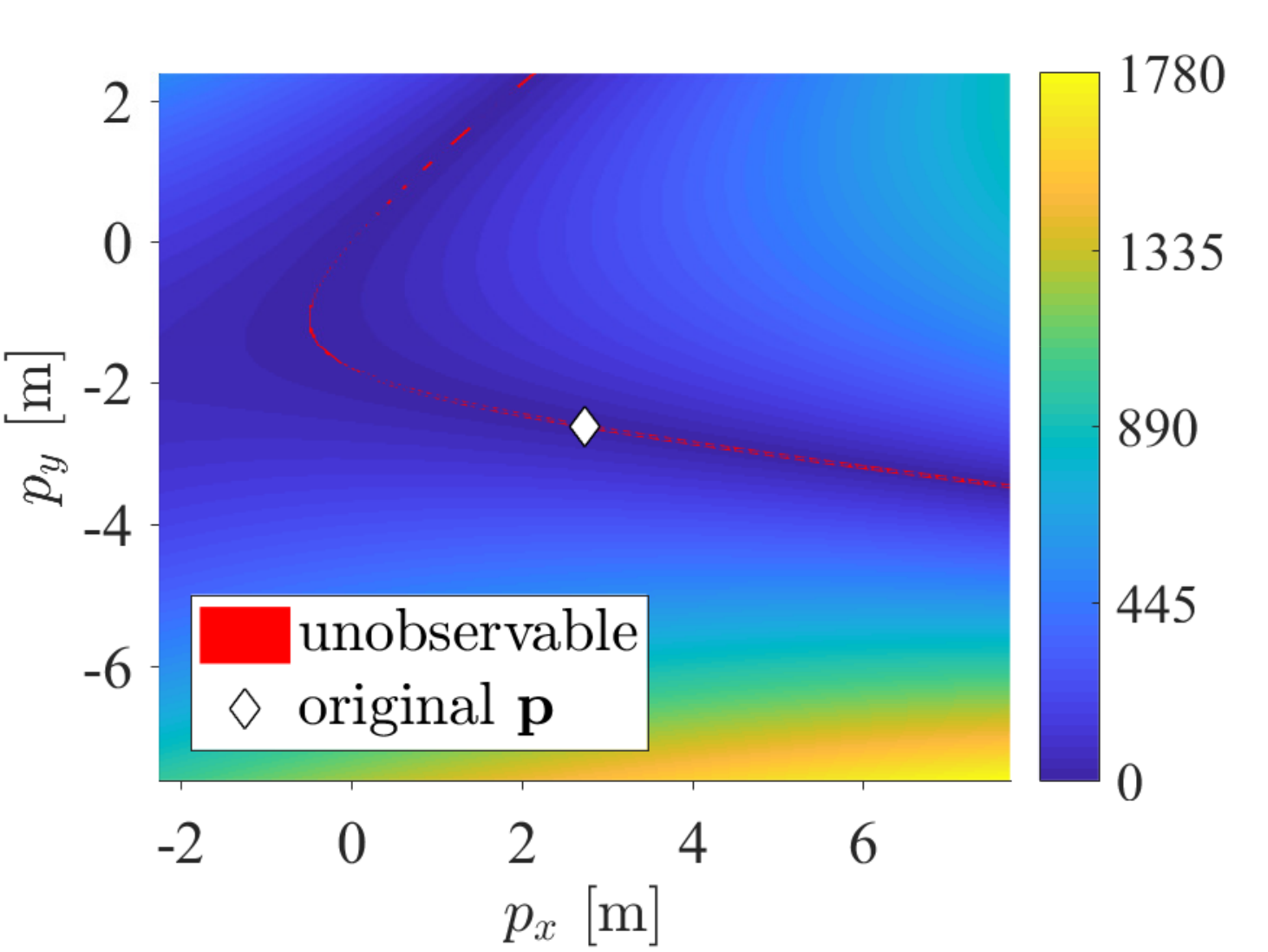}
		\caption{Unintuitive case 1}
		\label{fig:heat13}
	\end{subfigure}
	~
	\begin{subfigure}[t]{39mm}
		\centering
		\includegraphics[width=\columnwidth]{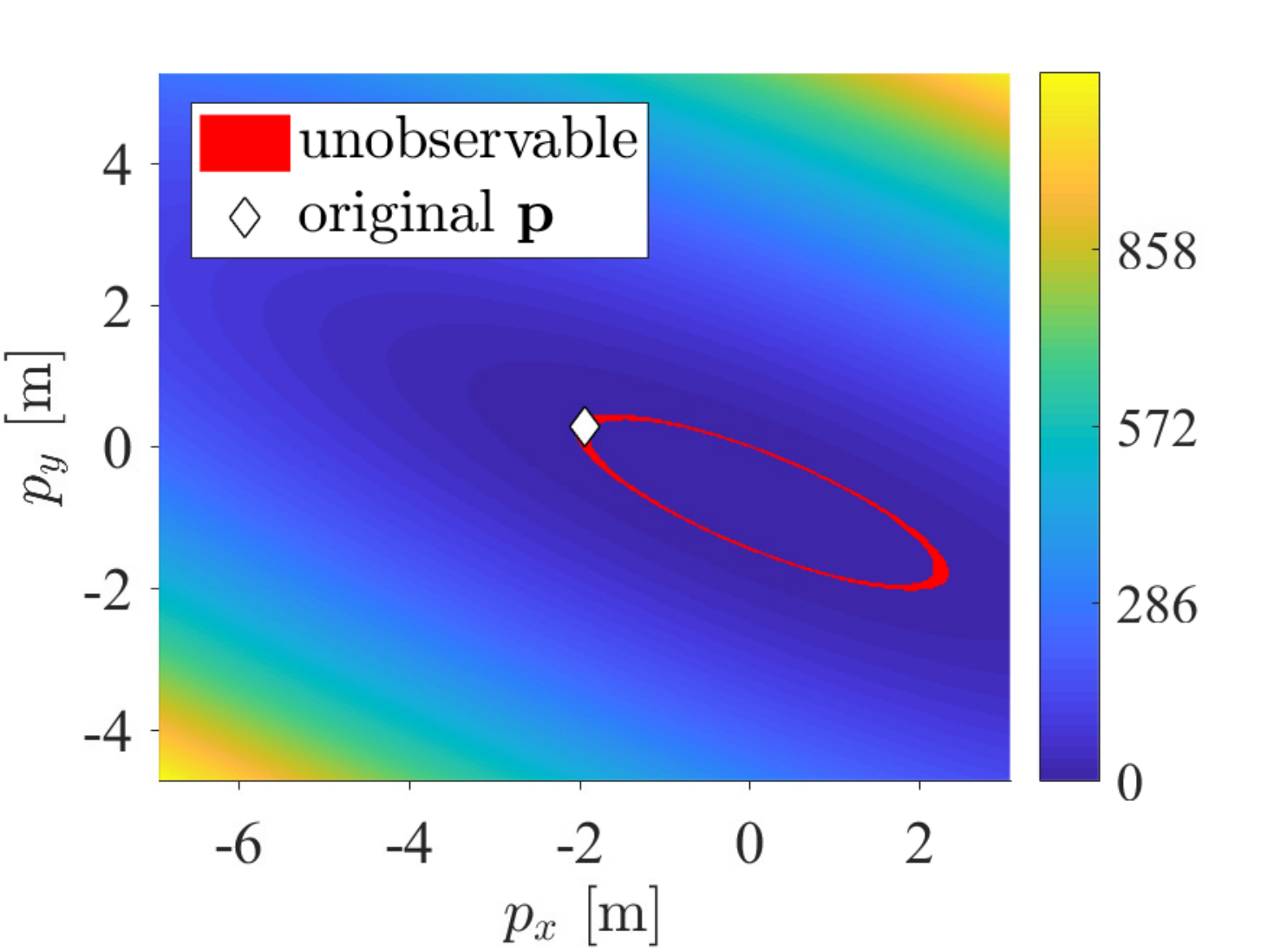}
		\caption{Unintuitive case 2}
		\label{fig:heat17}
	\end{subfigure}  
	\caption{Color map of observability for different relative positions. The velocities and accelerations of the MAVs are kept as depicted by figure \ref{fig:vect1} and the values for $\mathbf{p}^\intercal=[p_x,p_y]^\intercal$ are varied over a 10 m range.
	}
	\label{fig:heat1}
\end{figure}
  
  In order to study these  intuitive conditions in further detail, we evaluated how the observability of the system is affected once the relative position $\mathbf{p}$ between the MAVs changes.
  By varying the $p_x$ and $p_y$ values of the vector $\mathbf{p}$ around  the originally set values for $\mathbf{p}$ (as in \figref{fig:vect1}), we analyzed the observability of the system for different relative positions, while keeping the velocities and accelerations constant.
  The measure for observability was obtained by interpreting the meaning of \eqnref{eq:sys2cond}.
  It essentially tells that the left hand side of the equation should not be parallel to the relative position vector $\mathbf{p}$.
  Therefore, a practical measure of observability is how far away the left hand side of equation \eqnref{eq:sys2cond} is from being parallel to $\mathbf{p}$, which can be tested with the cross product.
  The absolute value of the cross product is then used as a measure of the observability of the system.   
  This paper considers a  cross product less than a value of 1 to be unobservable.\footnote{
  In reality, only if the cross product is truly 0 does it represent an unobservable condition, however the threshold does enable their visibility on the plot.}

  For the case of the second  (Eq. \eqref{eq:sys2cond2}, \figref{fig:heati21}) and the third intuitive condition (Eq. \eqref{eq:sys2cond3}, \figref{fig:heati31}) it can be seen that a varying $\mathbf{p}$ does not affect the unobservability in the color map.
   Once an acceleration vector is added to the state of MAV 1 in both cases, specifically $\mathbf{a}_1~=[0.3~0.3]^\intercal$, the color plots show that for a set of relative positions, the system does become observable again. 
   However, the chances of the MAVs ending up in an unobservable state are still significant within an operating area of 100 m$^2$.

The three intuitive conditions we extracted are only a subset of all conditions imposed by \eqnref{eq:sys2cond}.
  This means that there exist state and input combinations that satisfy the three intuitive conditions, but that do not satisfy \eqnref{eq:sys2cond}.
  In order to study what the implications of the full unobservability condition in \eqnref{eq:sys2cond} are, we used the Nelder-Mead simplex method to find other points in the state and input space that violate the full observability condition.
  Two examples are shown in \figref{fig:vect13} and \figref{fig:vect17}.
  These scenarios do not violate any of the intuitive conditions given by \eqnref{eq:sys2cond1}-\ref{eq:sys2cond3}.
  The relative position is non-zero, both MAVs have non-zero velocities and accelerations, and the velocity vectors are not parallel.
  Nevertheless, they violate \eqnref{eq:sys2cond}.
  Based on this, color maps for the unobservable conditions in \figref{fig:vect13} and \figref{fig:vect17} are given in \figref{fig:heat13} and \figref{fig:heat17}, respectively.

Both color maps of \figref{fig:heat13} and \figref{fig:heat17} clearly show a non-linear relationship between the relative position vector $\mathbf{p}$ and the observability of the system. 
Moreover, both maps show a different non-linear relationship. \figref{fig:heat13} shows more of a hyperbolic relationship, whereas the unobservable region in \figref{fig:heat17} looks more elliptical. 
It can be shown that different conditions show yet other relationships between the observability of the system for different relative positions $\mathbf{p}$. 
Moreover, these relationships only show what happens in two dimensions (for the two entries in the vector $\mathbf{p}$). 
In reality, the observability condition in \eqnref{eq:sys2cond} presents an 11 dimensional problem. 
It is therefore still difficult to deduce general rules from these results. 
What the latter two color maps  do have in common is that the unobservable relative positions are in all cases vastly outnumbered by the observable relative positions. 
This is different than what was observed for situations that would violate any of the more intuitive conditions in \eqnref{eq:sys2cond2} and \eqnref{eq:sys2cond3}.

\subsection{Comparison of the Two Systems}

Finally, the results from the observability analysis of both systems will be compared.
This will answer the question of what practical implications there are when moving from a system that relies on a common heading reference to a system that does not.

A primary result of the analysis is that removing the relative heading measurement results in a system that requires at least 
\emph{one extra Lie derivative} 
in the range observation to make the system locally weakly observable.
This is an important result, because it tells us that the heading-independent system $\sum_B$ relies more heavily on the range equation than $\sum_A$.
Without a heading observation, the range measurement serves to estimate a total of three states, as opposed to two in $\sum_A$.
Some of this information is contained in the second derivative of the range observation, and it is a well known fact that derivatives of a noisy signal become increasingly noisy.
In practice, this means that any system that wishes to perform range-based relative localization without a heading dependency needs an 
\emph{accurate and low-noise range measurement}.

Another important result is that the criteria posed for $\sum_B$ specify that 
\emph{both MAVs must be moving}.
Contrarily, the criteria for $\sum_A$ specify that only one of the MAVs must be moving.
Whilst this result might not be as relevant for MAV teams 
(as the MAVs will typically be moving anyway), this result can be important for other applications of range-based relative localization.
Think, for example, of the case where a single static beacon is used to estimate the position of a flying MAV using only range sensing and communication.
The results of our analysis show that $\sum_B$ is not observable in this case, and thus a common heading reference must be known for such a system to work 
(or, alternatively, the MAV must track the beacon and then communicate its estimate back to the beacon).
Note that, in the case where one of the participants is not moving, it can be shown that even the higher order Lie derivatives in $\sum_B$ will not succeed in making the observability matrix full rank, so that this statement generally holds.

A third difference is found in the condition for parallel movement of the two MAVs.
$\sum_A$ requires that the MAVs should not move in parallel at the same speed (which can be translated to mean that there should be a non-zero relative velocity between the two MAVs).
Instead, $\sum_B$ requires that \textit{the MAVs should not be moving in parallel regardless of speed}.
Therefore, even if the second MAV were to be moving twice as fast as the first, the filter would not be observable as long as the direction of movement is the same.
However, $\sum_B$, can bypass this condition in some cases if either of the MAVs is also simultaneously accelerating.
Similarly, it can be shown that $\sum_A$ is able to bypass the parallel motion condition with acceleration, although a second order Lie derivative would be necessary in that case.

\section{Verification through Simulations}
\label{sec:Simulation}

In this section, we further investigate the conclusions drawn from the analytical observability analysis.
At first, a kinematic, noise-free study is performed to verify and confirm the differences in the observability conditions for $\sum_A$ and $\sum_B$. Afterwards, the influence of noise and disturbances on the filter are studied.

\subsection{Filter Design}
The filter of choice, used throughout the rest of this paper, is an Extended Kalman Filter (EKF).
This choice was made because this type of filter fits intuitively with how the state-space system was described in \secref{sec:Obs}.
The EKF also uses a state differential model and an observation model.
The state differential model can thus be kept exactly as the one given earlier in \eqnref{eq:ssmodel}.
The observation models for $\sum_A$ and $\sum_B$ are also kept almost the same as given in \eqnref{eq:obsmodel1} and \eqnref{eq:obsmodel2}, 
with the only adjustment that mow the full range $||\mathbf{p}||_2$ is observed, rather than half the squared range $\frac{1}{2}\mathbf{p}^\intercal \mathbf{p}$.
Furthermore, using the EKF is in line with earlier research on range-based relative localization \citep{Coppola2016}.

An EKF has parameters that need to be tuned, namely:
the initial state,  
the system and measurement noise matrices, 
and the initial state covariance matrix. 
The initial state is an important setting that will be described where appropriate in the next sections.
The matrices are always tuned to correspond to the actual expected values.
The measurement noise matrix is tuned based on the expected noises on the measurements, and similarly for the system noise matrix.
However, since some of the simulations also make use of perfect measurements 
(with zero noise) and since a noiseless entry in the measurement noise matrix is not possible, the corresponding entries are then given a small value of 0.1.

\subsection{Kinematic, noise-free study of unobservable situations}
In the first simulated study, the two MAVs that are studied have kinematic trajectories that can be described analytically.
The MAVs also have perfect noise-free knowledge of the inputs and measurements.
The kinematic and noise-free situation is used to confirm conclusions drawn in the observability analysis performed in \secref{sec:Obs}.

The two MAVs involved in the EKF are designated MAV 1 and MAV 2. 
MAV 1 shall be the host of the EKF and shall attempt to track the relative position of MAV 2.
For clarity, this MAV is denoted as the \emph{host} of the filter. 
MAV 2 is the one whose position is tracked by MAV 1.
It does not run an EKF.
For clarity, this MAV is denoted as the \emph{tracked} MAV.
The following three scenarios are studied:
\begin{enumerate}
\item MAV 1 (host) is moving and MAV 2 (tracked) is stationary.
\item MAV 1 (host) is stationary and MAV 2 (tracked) is moving.
\item MAV 1 (host) and MAV 2 (tracked) are both moving in parallel to each other at different speeds.
\end{enumerate}

These scenarios have been chosen because they match the intuitive conditions where $\sum_A$ is observable, but $\sum_B$ is not. 
These are limit cases and therefore provide valuable verification of the analytically found differences between the two systems. 

The simulations will show whether these different scenarios have convergent EKFs or not.
The focus of this analysis is on the estimation of the relative position $\mathbf{p}$ and the relative heading $\Delta\psi$.
Since the velocities are observed directly, these are observable regardless of the situation, and are thus not shown. 

The initial velocities of MAVs 1 and 2 are initialized to their true value, since these are not the variables of interest in this analysis.
The initial position and relative heading are initialized with an error, the specifics of which will be given in the respective scenarios.
The yaw rates and headings of both MAVs are kept at 0 rad/s and 0 rad, respectively.
The EKF runs at a frequency of 50 Hz.

The error measure throughout this paper is the Mean Absolute Error (MAE). The separate $x$ and $y$ errors in the relative location estimate $\mathbf{p}$ are combined according to the norm $||\mathbf{p}||_2$. This choice was made because the separate errors in $x$ and $y$ directions offer little additional insight and are mostly very similar.

\subsubsection{MAV 1 (host) moving, MAV 2 (tracked) stationary}

Previous analytical analysis has shown that $\sum_A$ is locally weakly observable, while $\sum_B$ is not observable.
This result is therefore expected to be reflected in the simulation as well.

In the simulation, MAV 1 (the host) is positioned at $\mathbf{p}_{1,0}^\intercal = [0,0]^\intercal$ and has a constant velocity $\mathbf{v_1}^\intercal = [1,0]^\intercal$.
MAV 2 (the tracked MAV) is positioned at $\mathbf{p}_{2,0}^\intercal = [1,1]^\intercal$ with no velocity or acceleration.
The initial guess of MAV 1 for the relative position and heading of MAV 2 is $[\hat{\mathbf{p}}_0^\intercal,\hat{\Delta\psi}_0]^\intercal = [0.1,0.1,1]^\intercal$.
This means that the initial estimation error in $p_x$, $p_y$, and $\Delta\psi$ is thus equal to 0.9, 0.9, and 1, respectively.

\begin{figure}[t!]
  \centering
  \begin{subfigure}[t]{39mm}
    \centering
    \includegraphics[width=\textwidth]{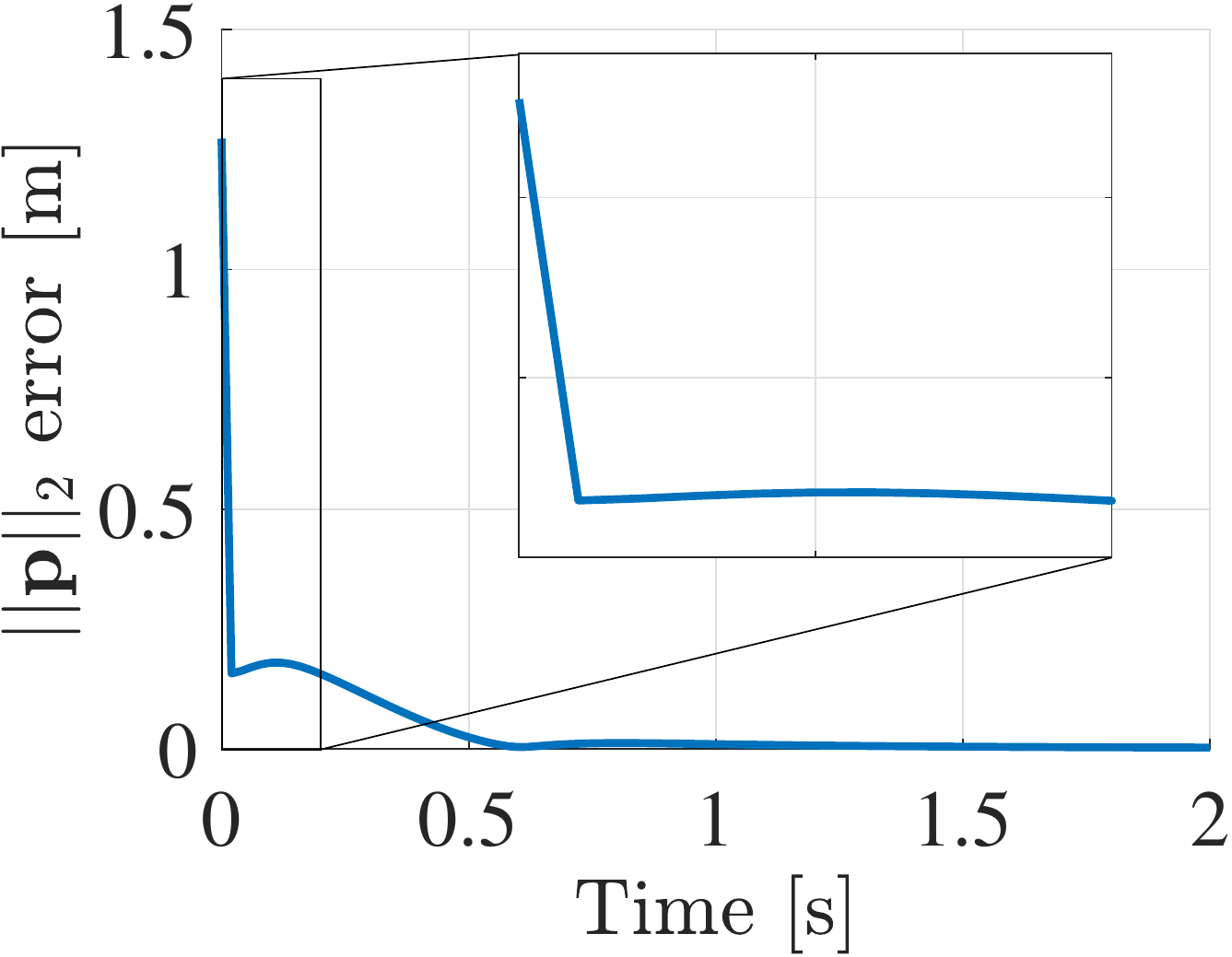}
    \label{fig:case11_1}
  \end{subfigure}
  ~
  \begin{subfigure}[t]{39mm}
    \centering
    \includegraphics[width=\textwidth]{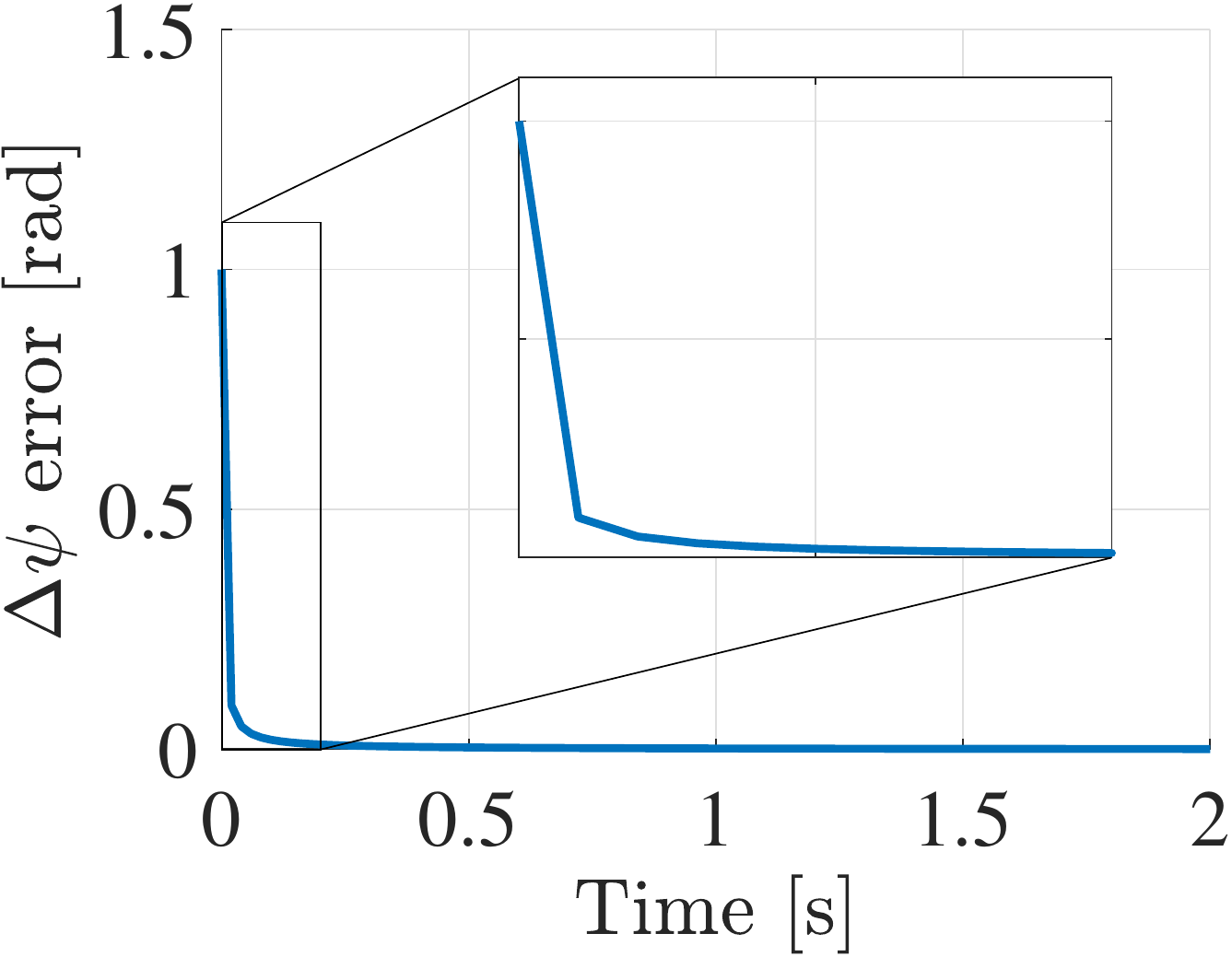}%
\label{fig:case11_2}
  \end{subfigure}
\caption{$\sum_A$ EKF convergence for case 1: MAV 1 (host) moving, MAV 2 (tracked) stationary}
\label{fig:case11}
  \end{figure}

\begin{figure}[t!]
  \centering
  \begin{subfigure}[t]{39mm}
    \centering
    \includegraphics[width=\textwidth]{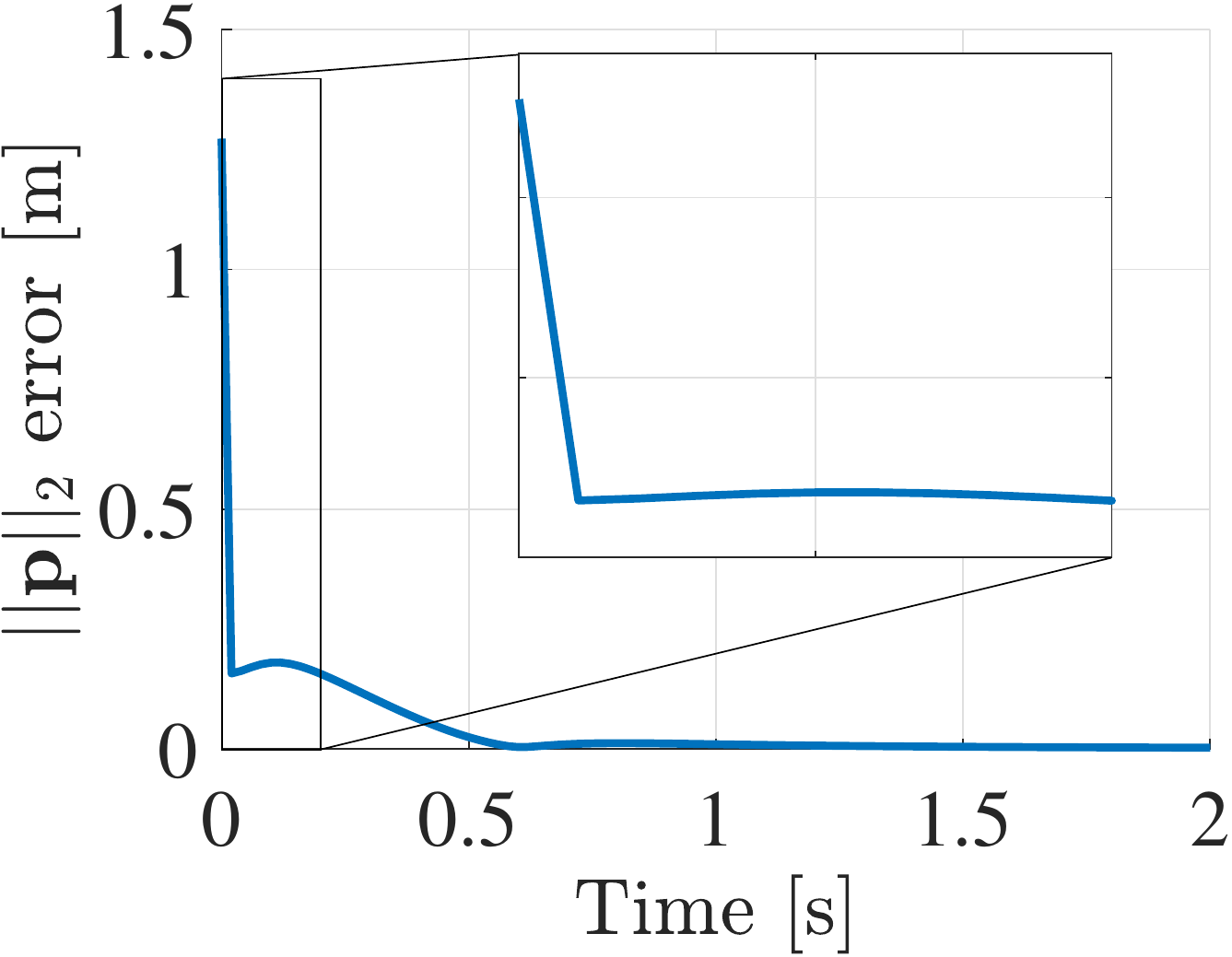}
    \label{fig:case12_1}
  \end{subfigure}
  ~
  \begin{subfigure}[t]{39mm}
    \centering
    \includegraphics[width=\textwidth]{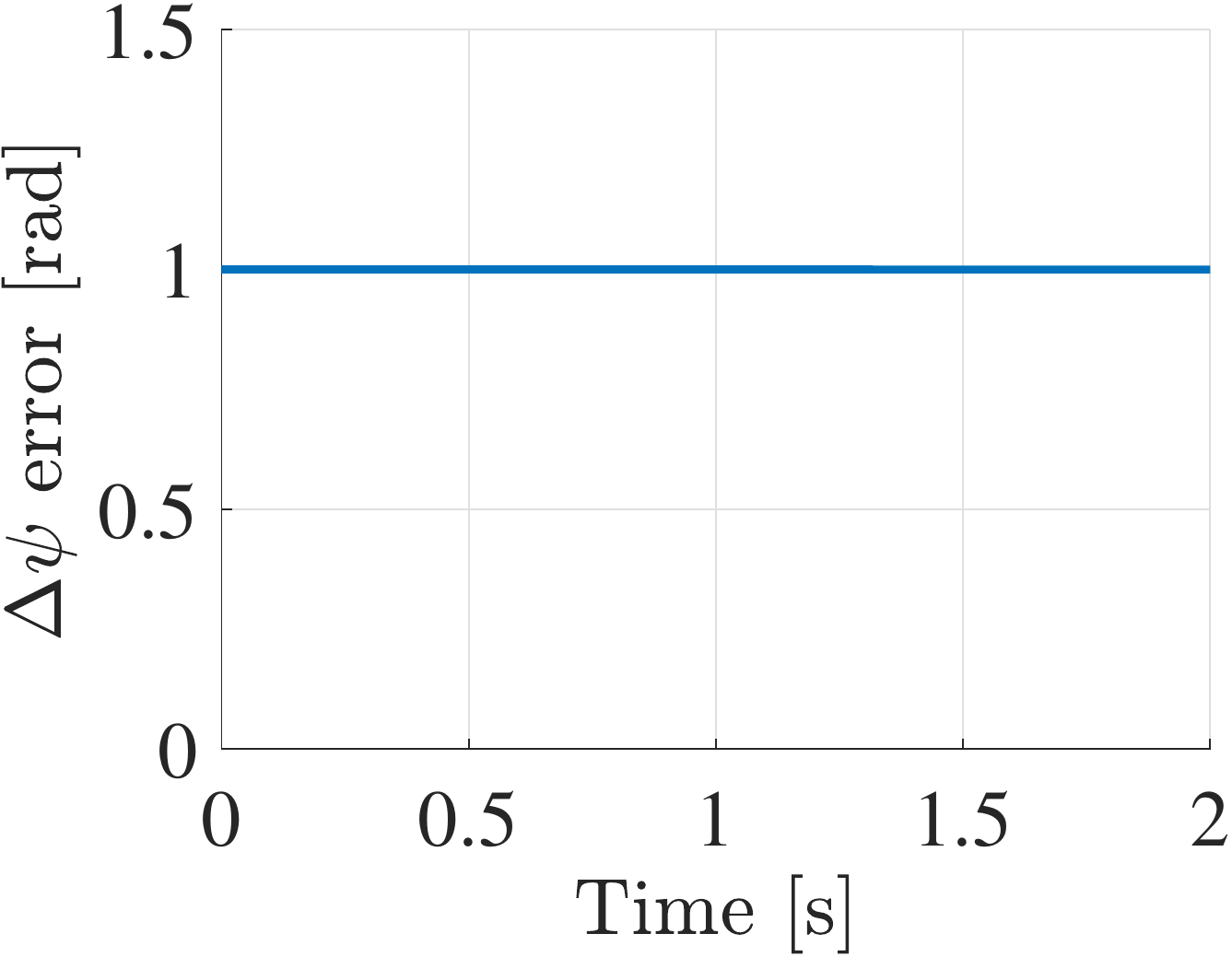}
    \label{fig:case12_2}
  \end{subfigure}
  \caption{$\sum_B$ EKF convergence for case 1: MAV 1 (host) moving, MAV 2 (tracked) stationary}
  \label{fig:case12}
  \end{figure}

As can be seen in \figref{fig:case11}, both the relative position $\mathbf{p}$ error and the relative heading $\Delta\psi$ error quickly converge to 0. 
Contrarily, the observability analysis of $\sum_B$ has shown that this scenario is not locally weakly observable, because the second condition is violated, i.e., one of the MAVs is not moving.
However, \figref{fig:case12} shows that the $||\mathbf{p}||_2$ error converges to 0 just as rapidly as for $\sum_A$.
A more thorough inspection shows that the unobservable state of the system is in fact $\Delta\psi$, which is the one that does not converge.
This is a favorable result, since the relative position is typically the variable of interest, rather than the difference in heading. 

The reason that this occurs lies in the information provided by the first state differential equation.
This equation tells us that $\dot{\mathbf{p}} = -\mathbf{v_1}+\mathbf{R}\mathbf{v_2}-\mathbf{S_1 p}$.
The only dependency that this equation has on the relative heading $\Delta\psi$ is in the rotation matrix $\mathbf{R}$.
Therefore, as long as $\mathbf{v_2}$ is equal to $\mathbf{0}$, the differential equation for $\dot{\mathbf{p}}$ has no dependency on the relative heading between the two MAVs.
The convergence of $\mathbf{p}$ therefore remains unaffected.
The situation changes when it is $\mathbf{v_2}$ that is non-zero and $\mathbf{v_1}$ that is zero. This case is studied next.

\subsubsection{MAV 1 (host) stationary, MAV 2 (tracked) moving}
\begin{figure}[t!]
  \centering
  \begin{subfigure}[t]{39mm}
    \centering
    \includegraphics[width=\textwidth]{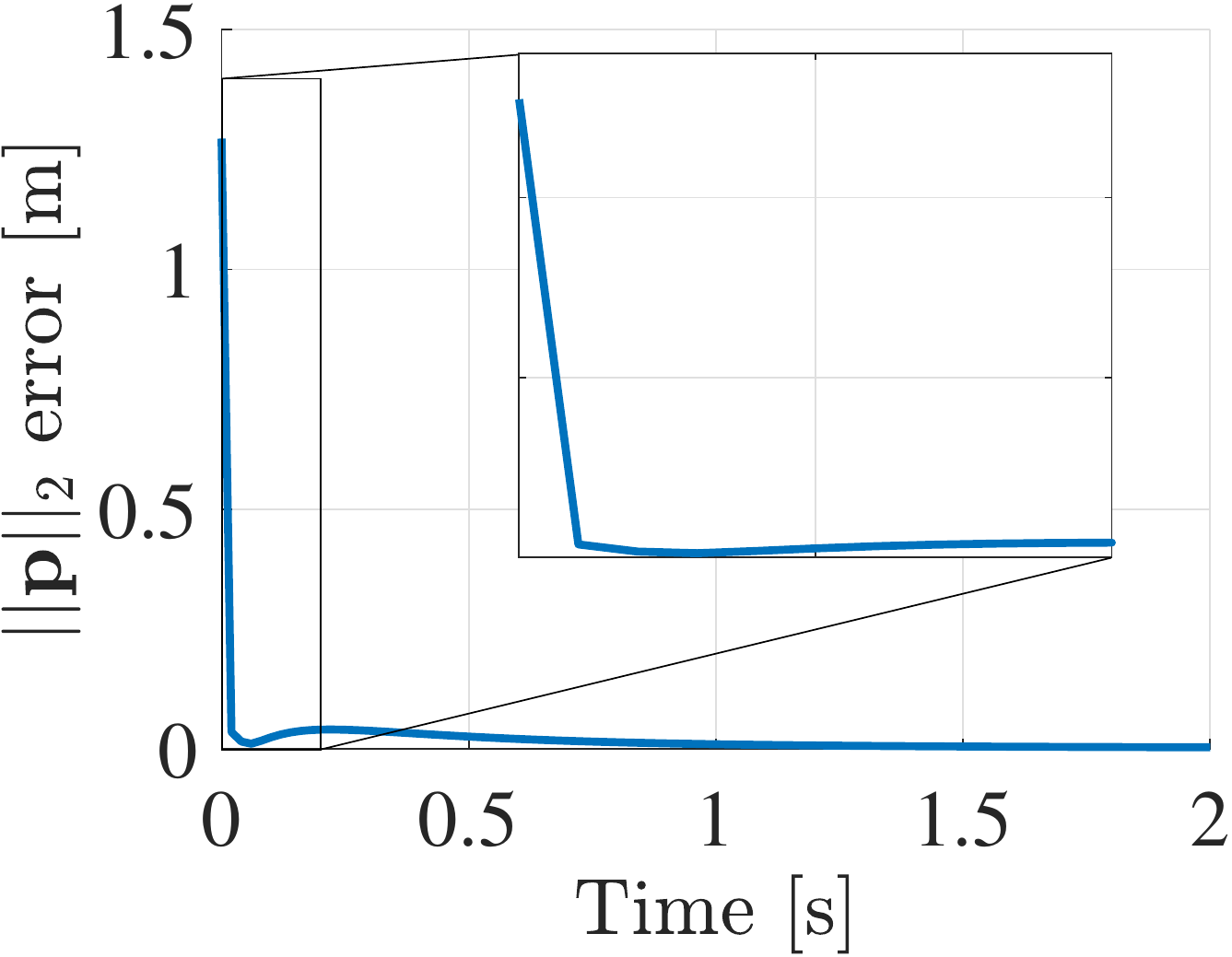}
\label{fig:case21_1}
  \end{subfigure}
  ~
  \begin{subfigure}[t]{39mm}
    \centering
    \includegraphics[width=\textwidth]{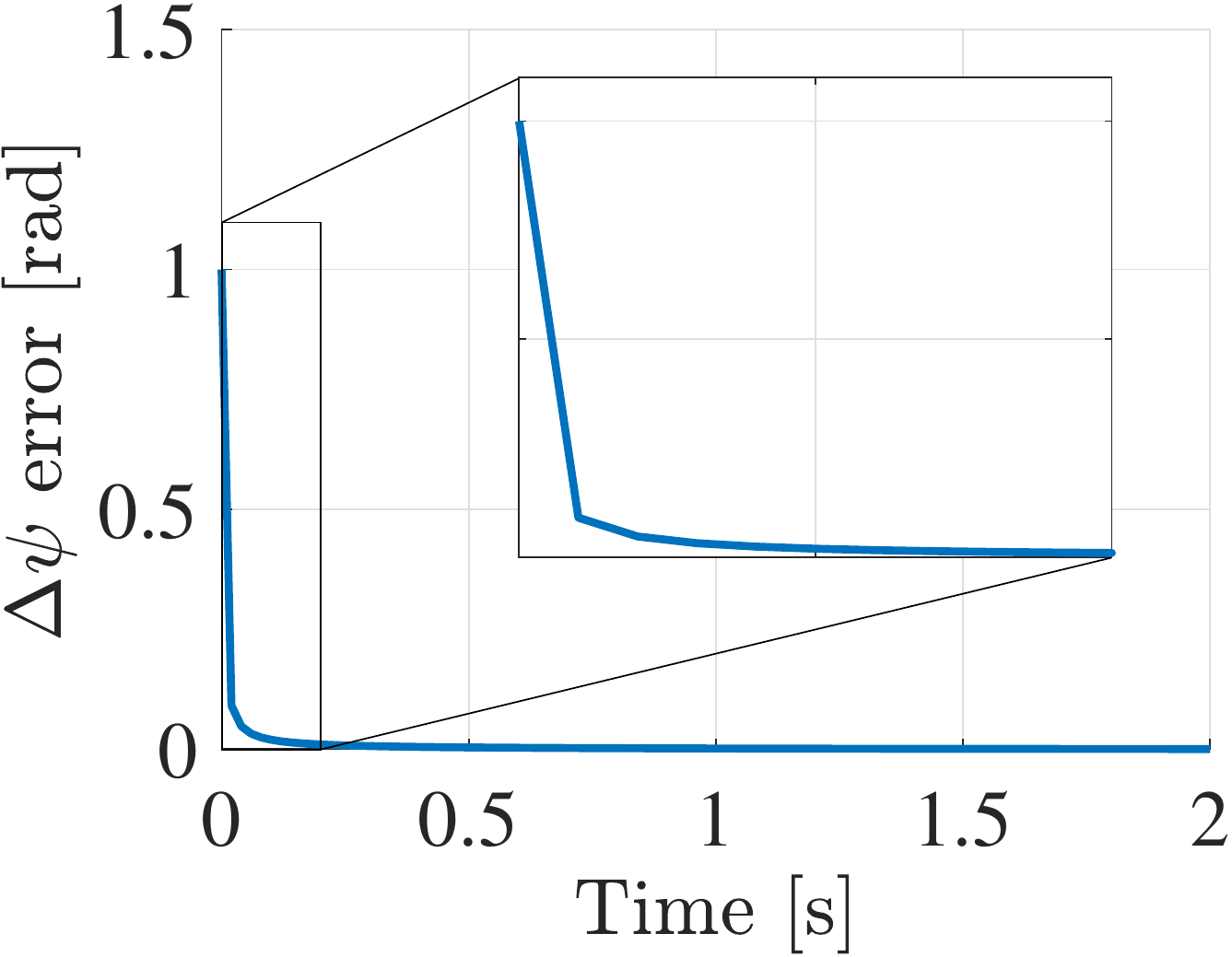}
\label{fig:case21_2}
  \end{subfigure}
\caption{$\sum_A$ EKF convergence for case 2: MAV 1 (host) stationary, MAV 2 (tracked) moving}
\label{fig:case21}
  \end{figure}

\begin{figure}[t!]
  \centering
  \begin{subfigure}[t]{39mm}
    \centering
    \includegraphics[width=\textwidth]{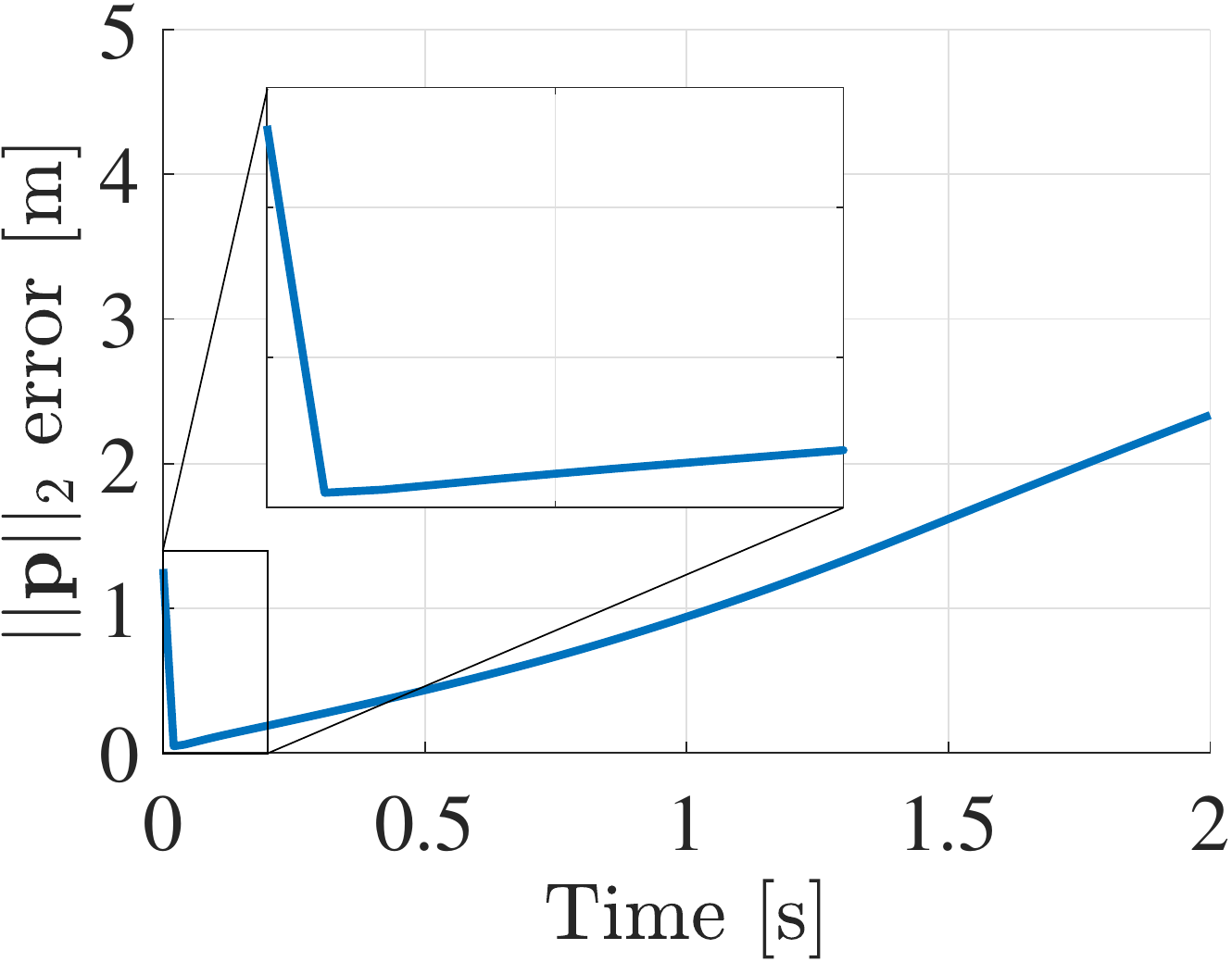}%
\label{fig:case22_1}
  \end{subfigure}
  ~
  \begin{subfigure}[t]{39mm}
    \centering
    \includegraphics[width=\textwidth]{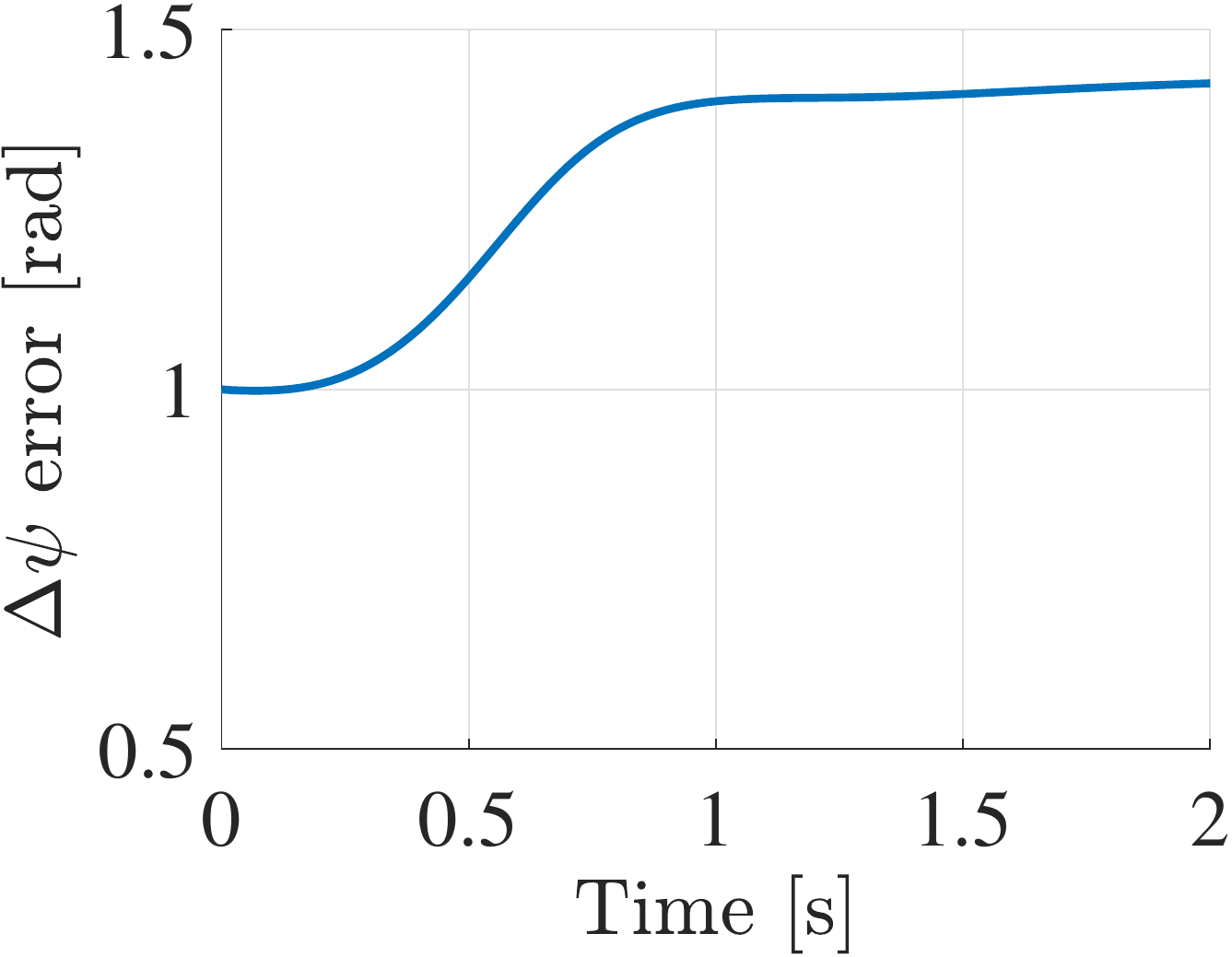}
\label{fig:case22_2}
  \end{subfigure}
\caption{$\sum_B$ EKF convergence for case 2: MAV 1 (host) stationary, MAV 2 (tracked) moving}
\label{fig:case22}
  \end{figure}

For this case, all of the parameters are the same as for case 1, with the only difference being that now $\mathbf{v_1}=\mathbf{0}$ and $\mathbf{v_2}^\intercal = [1,0]^\intercal$. 
The analytical observability analysis has shown that this scenario is locally weakly observable for $\sum_A$.
As expected, it can be seen in \figref{fig:case21} that both the errors for $\mathbf{p}$ and $\Delta\psi$ converge rapidly to 0. 
The observability analysis has then shown that $\sum_B$ is not locally weakly observable in this scenario.
Indeed, \figref{fig:case22} shows that both $||\mathbf{p}||_2$ and $\Delta\psi$ do not converge and that $||\mathbf{p}||_2$ even diverges. 

This time, because $\mathbf{v_2}$ is not equal to $\mathbf{0}$, the state differential equation for the relative position of MAV 2 has a dependency on the relative heading state $\Delta\psi$.
Because $\Delta\psi$ does not converge to its true value, and eventually settles at an error of approximately 1.5 rad, there is a large inaccuracy in the state differential equation for $\dot{\mathbf{p}}$.
This consequently results in an ever increasing error in $\mathbf{p}$, since MAV 1 essentially `thinks' that MAV 2 is flying in a different direction than it really is.

This shows the reason as to why it is generally not possible for a stationary vehicle (or beacon) to be tracking a moving vehicle using range-only measurements and velocity information without a common heading reference. 
Contrarily, it \textit{is} possible for a moving vehicle to be tracking a stationary vehicle or beacon's position.
This is entirely caused by the fact that a vehicle will always be `aware', in its own body frame, of the direction it is moving in and hence does not need a convergent estimate of the relative heading with respect to the vehicle it is tracking.
However, when the vehicle it is tracking does move, it needs this convergent estimate of the relative heading to know which direction the other is moving in.

\subsubsection{MAV 1 (host) and MAV 2 (tracked) moving in parallel at different speeds}
Finally, the case where both MAVs are moving in parallel, but at different speeds, is studied.
Once more, most of the parameters are kept the same as those presented under case 1.
This time, the velocity of MAV 2 is set to $\mathbf{v_2}^\intercal=[1,0]^\intercal$ and the velocity of MAV 1 is set in a parallel direction, but with twice the magnitude ($\mathbf{v_1}^\intercal=2\mathbf{v_2}^\intercal=[2,0]^\intercal$).

\begin{figure}[t!]
  \centering
  \begin{subfigure}[t]{39mm}
    \centering
    \includegraphics[width=\textwidth]{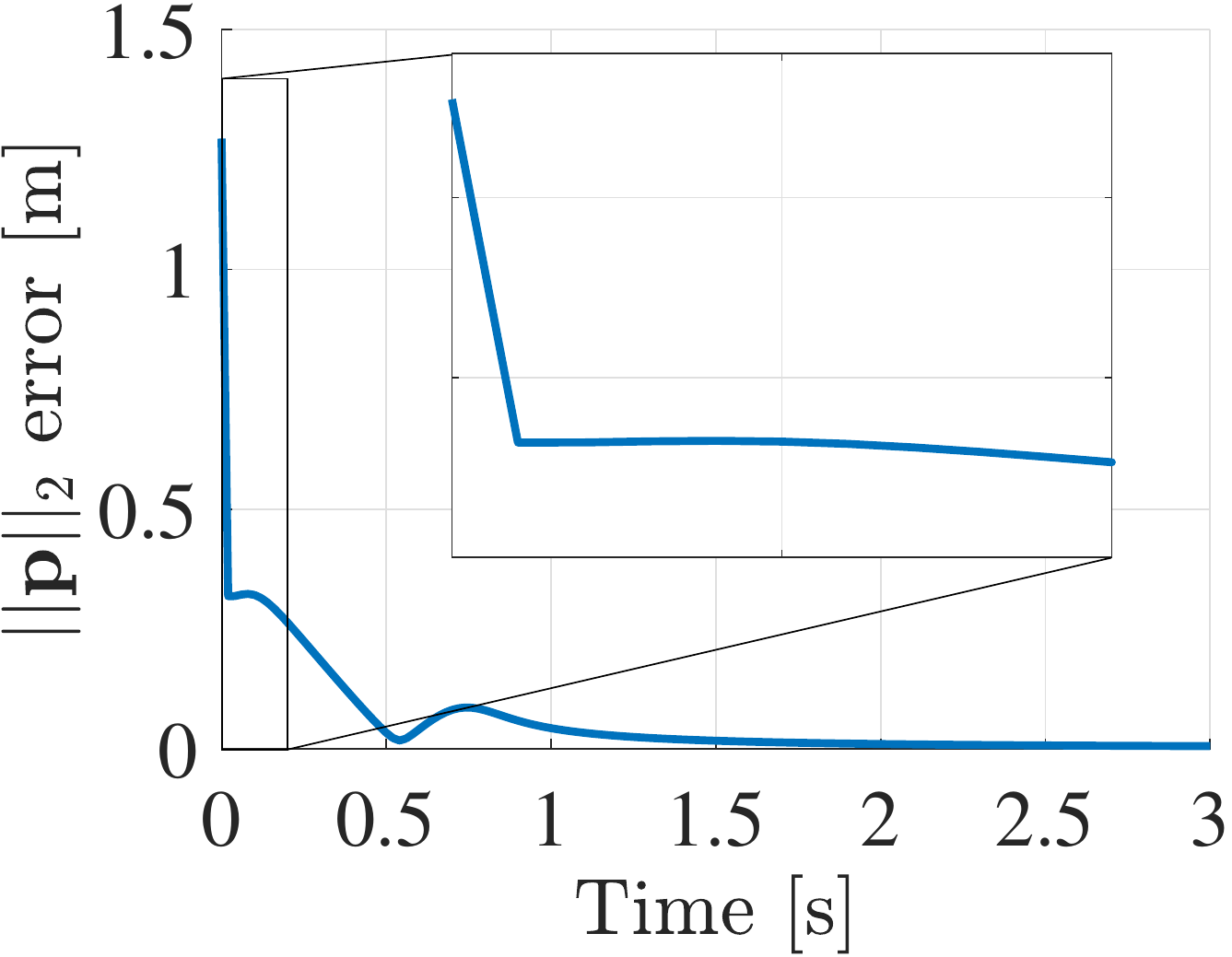}
\label{fig:case31_1}
  \end{subfigure}
  ~
  \begin{subfigure}[t]{39mm}
    \centering
    \includegraphics[width=\textwidth]{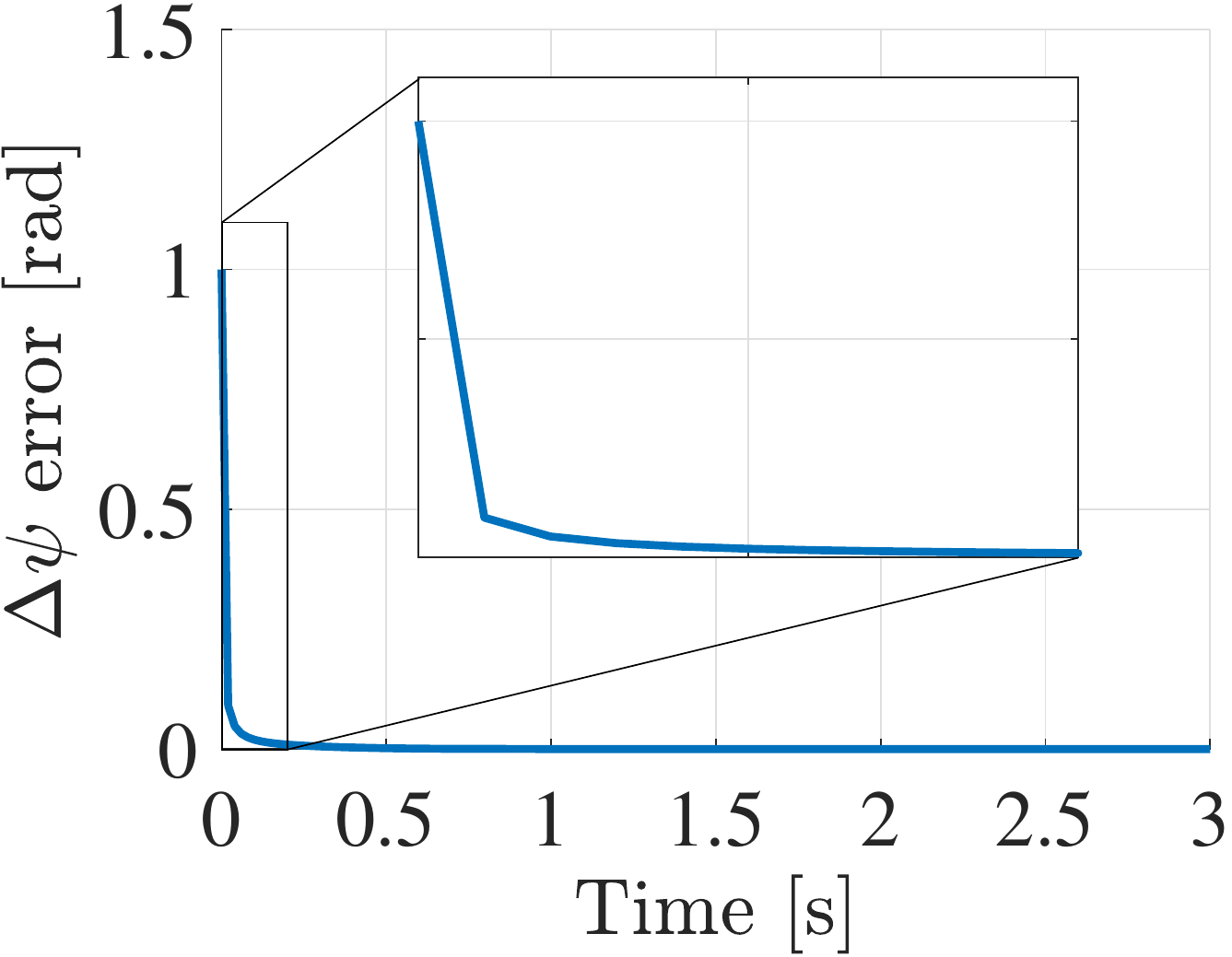}
\label{fig:case31_2}
  \end{subfigure}
\caption{$\sum_A$ EKF convergence for case 3: MAV 1 (host) and MAV 2 (tracked) moving in parallel}
\label{fig:case31}
  \end{figure}

\begin{figure}[t!]
  \centering
  \begin{subfigure}[t]{39mm}
    \centering
    \includegraphics[width=\textwidth]{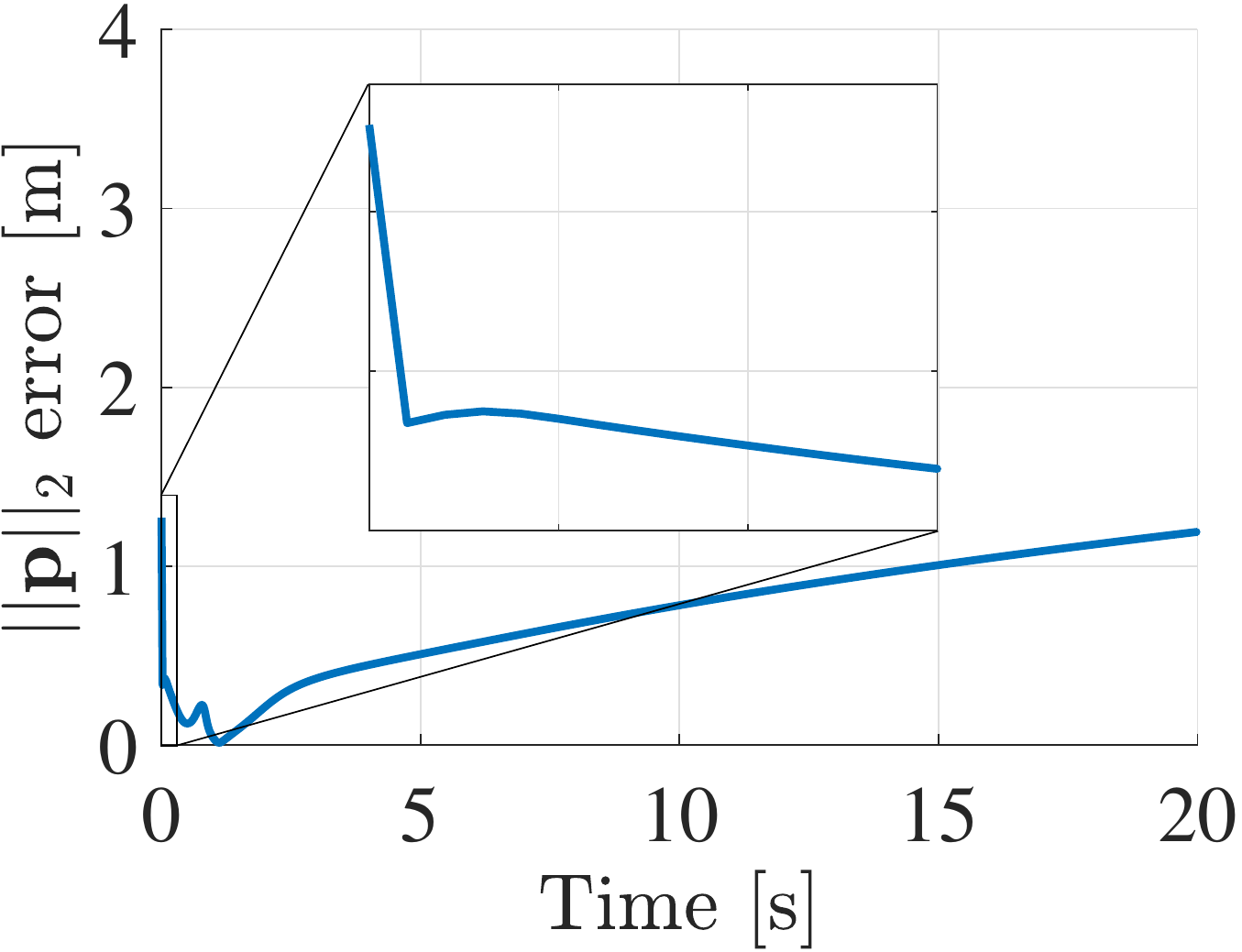}
    \label{fig:case32_1}
  \end{subfigure}
  ~
  \begin{subfigure}[t]{39mm}
    \centering
    \includegraphics[width=\textwidth]{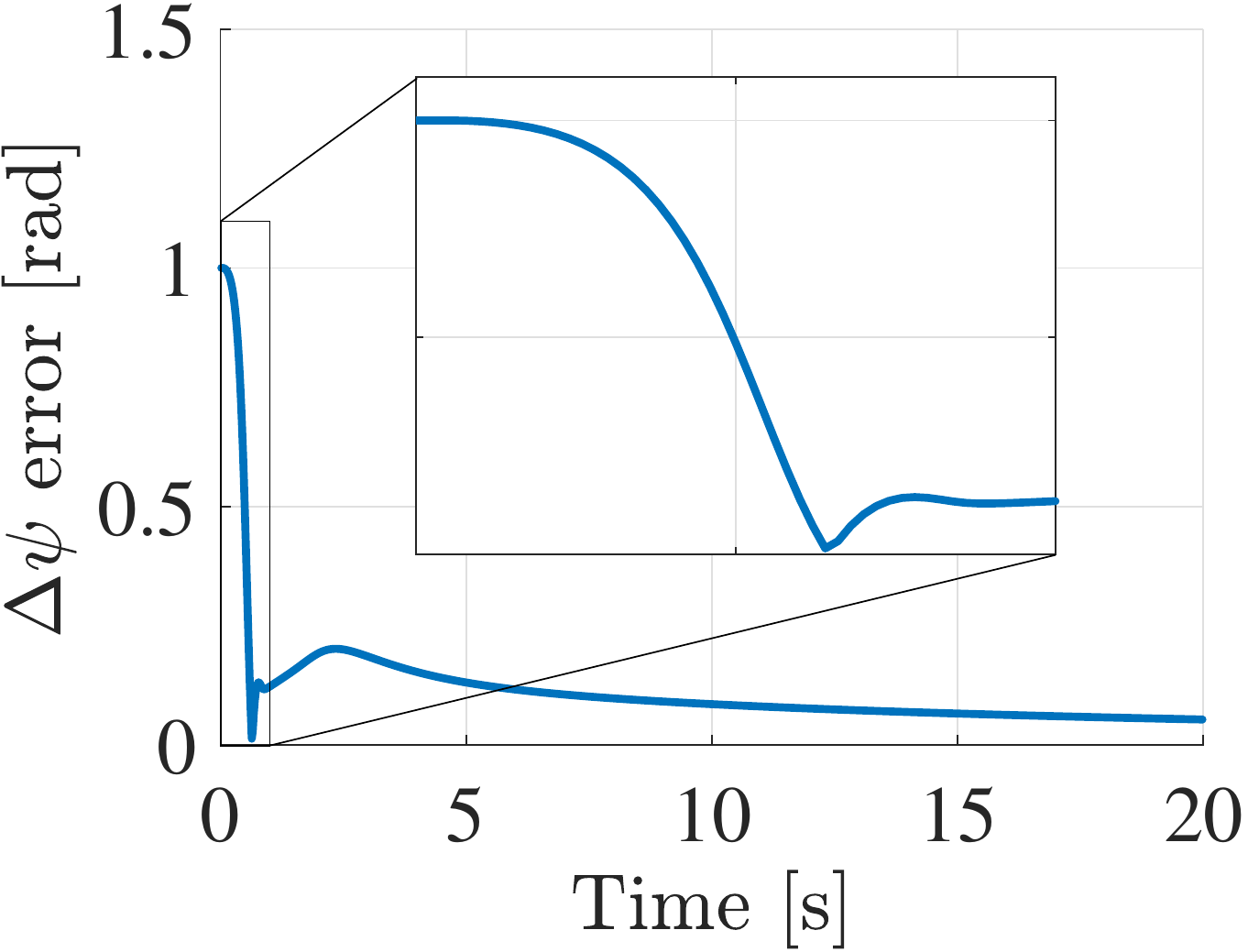}
\label{fig:case32_2}
  \end{subfigure}
\caption{$\sum_B$ EKF convergence for case 3: MAV 1 (host) and MAV 2 (tracked) moving in parallel}
\label{fig:case32}
  \end{figure}

According to the observability analysis, this is one of the limit cases where $\sum_A$ is still just observable, but $\sum_B$ is not.
Indeed, \figref{fig:case31} shows convergent behavior for $\sum_A$, whereas \figref{fig:case32} shows divergence for $\sum_B$. 
Note that the filter for $\sum_B$ has a decreasing error in $\Delta \psi$. However, the convergence for $\Delta \psi$ is very slow (notice how this situation has been simulated for a much longer time than the previous cases). Furthermore, the error for $\mathbf{p}$ continues to rise indefinitely.

This result concludes the noise-free simulations that compare the performance of the filters for $\sum_A$ and $\sum_B$. These simulations verify that the conclusions regarding the differences between the two filters in \secref{sec:Obs} also hold true when translated to a simulation environment.

\subsection{Kinematic, noisy range measurements study of observable situation}

Whilst a noise-free study demonstrates the feasibility of the proposed filter and can verify the differences between $\sum_A$ and $\sum_B$, it is also important to study the filter's performance when presented with noisy data.
Not only is this more representative of the filter's performance in practice, but it also can be used to verify one of the main conclusions that were drawn in the observability study, namely that $\sum_B$ needs information present in the second derivative of the range data to be observable, compared to only a first derivative for $\sum_A$.
It is consequently expected that, with all other parameters fixed, $\sum_B$ will perform increasingly worse as the range data becomes more noisy.

In this study, we steer away from unobservable scenarios.
The intent now is to study both filter's performances for the case where the filters \textit{are} known to be observable, in order to compare their performance.
For this reason, the trajectories of MAV 1 (host) and MAV 2 (tracked) are designed so as to stay clear of the unobservable situations and to excite the filter properly through relative motion.
The trajectories that we devised for this study are perfectly circular, and we assume that the MAVs fly at the same height.

\begin{figure}[t!]
\centering
      \def\svgwidth{0.5\linewidth}\footnotesize
      \input{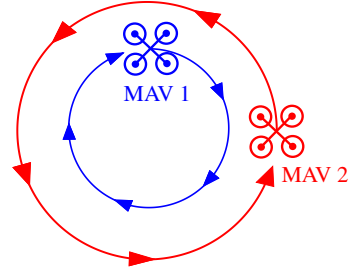}
\caption{Two circular trajectories for MAV 1 and MAV 2}
\label{fig:dcirctraj}
\end{figure}

The trajectories, depicted in \figref{fig:dcirctraj}, can be described in polar coordinates $[\rho, \theta]$.
MAV 1 flies a circular motion at an angular velocity $\dot{\theta_1}=\omega_1$ with radius $\rho_1$, and MAV 2 flies at angular velocity $\dot{\theta_2}=\omega_2$ with radius $\rho_2$.
To ensure that both MAVs have sufficient relative motion, one MAV flies clockwise and the other counter clockwise, such that $\omega_1=-\omega_2$. Moreover, the radius of MAV 2's trajectory is 1 meter larger than MAV 1's trajectory, and is offset by 90$^{\circ}$ in angle, such that $\rho_1=\rho_2-1$ and $\theta_1=\theta_2+\frac{\pi}{2}$. 

The radius difference in the trajectories ensures that the situation $\mathbf{p}=\mathbf{0}$ is avoided, and the angle offset ensures that the relative velocities are distributed more or less equally in $x$ and $y$ directions. Note that, for simplicity, both MAVs keep a steady heading such that $\psi_1=\psi_2$ and $r_1=r_2=0$.
Switching back to Cartesian coordinates, the trajectories can thus be analytically described as follows.
MAV 2's position vector in time is given by:

\begin{equation}
\label{eq:p2traj}
\mathbf{p_{2}}(t)=\left[ {\begin{array}{*{20}{c}}
{\rho_2 cos(\omega_2 t)}\\
{\rho_2 sin(\omega_2 t)}
\end{array}} \right]
\end{equation}

MAV 1's position vector in time can be described by:

\begin{align}
\nonumber
\mathbf{p_{1}}(t)&=\left[ {\begin{array}{*{20}{c}}
{(\rho_2-1) cos(-\omega_2 t + \frac{\pi}{2})}\\
{(\rho_2-1) sin(-\omega_2 t+\frac{\pi}{2})}
\end{array}} \right] \\
\label{eq:p1traj}
&= \left[ {\begin{array}{*{20}{c}}
{-(\rho_2-1) sin(-\omega_2 t)}\\
{(\rho_2-1) cos(-\omega_2 t)}
\end{array}} \right]
\end{align} 

The equations for $\mathbf{v_i}(t)$ and $\mathbf{a_i}(t)$ can be obtained by taking the time derivatives with respect to $\mathbf{p_i}(t), \ i=1,2$.
Note that this is not true for the general case, since $\mathcal{H}_i$ is a rotating frame of reference, but in this case it is possible because the MAVs keep a constant heading equal to 0 rad.

By setting $\rho_2=4$ m and $\omega_2=\frac{2\pi}{20}$ rad, the trajectory of MAV 2 becomes a circle with a radius of 4 m that is traversed in 20 s.
To comply with the previously defined constraints, $\rho_1$ and $\omega_1$ are 3 m and $-\frac{2\pi}{20}$ rad/s, respectively.
These values are representative of what a real MAV should easily be capable of and result in relative velocities of about 1 m/s in $x$ and $y$ directions between the two MAVs.

The study will test the performance of the relative localization filter as seen from the perspective of MAV 1, who is thus tracking MAV 2. The filter is fed perfect information on all state and input values, except for the measurement of the range $||\mathbf{p}||_2$ between the two MAVs.
The range measurement are artificially distorted with increasingly heavy Gaussian white noise.
The measured range fed to the filter is thus $||\mathbf{p}||_{2,m}= ||\mathbf{p}||_2 + n(\sigma_{R})$, where $n(\sigma_{R})$ is a Gaussian white noise signal with zero mean and standard deviation $\sigma_{R}$. The standard deviations that are tested are 0 (noise free), 0.1, 0.25, 0.5, 1, 2, 4, and 8 m.
In practice, a standard deviation of 8 m could be consider quite high, but this is intentionally chosen with the intent to observe a significant difference in the error.
Since this study keeps all the other measurements and inputs noise free, the noise on the range measurement needs to be higher to get a significant increase in the localization error.

This time the EKF runs at 20 Hz, which is more representative of our real-world set-up, discussed later in \secref{sec:Experiment}.
The described flight trajectory is simulated for 20 seconds each run (which is thus one complete revolution of the circular trajectory).
The EKF is initialized to the true state to exclude the effects of initialization.

For each particular noise standard deviation, both the filter for $\sum_A$ and for $\sum_B$ are simulated with 1000 different noise realizations.
For each realization the MAE of the estimated $\mathbf{p}$ with respect to its true value is computed, again by considering the combined error in the estimate of $||\mathbf{p}||_2$.
After 1000 realizations, the Average MAE (AMAE) is computed to extract the average performance for all noise realizations. 

\begin{figure}[t!]
\centering
\includegraphics[width=0.8\columnwidth]{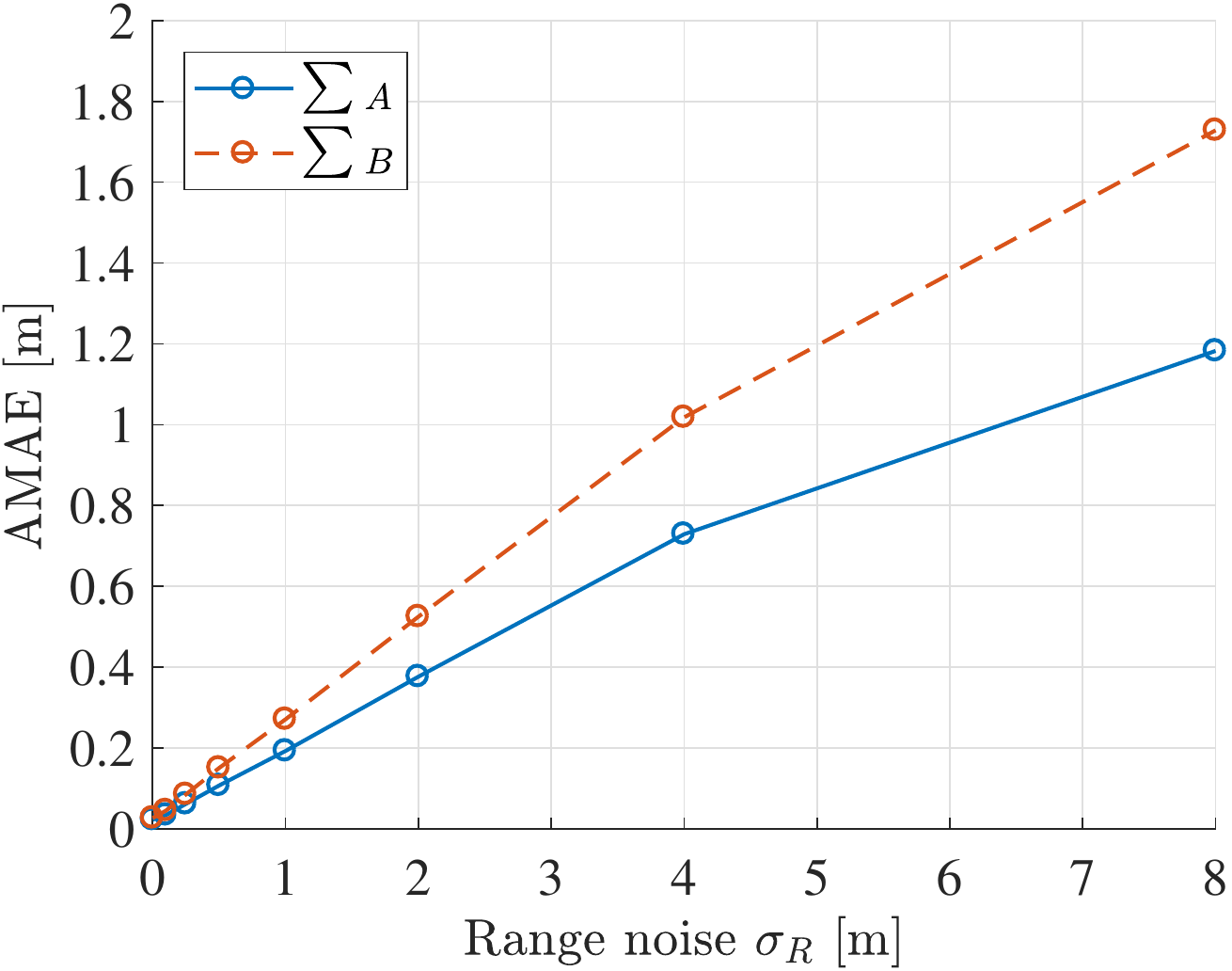}%
\caption{AMAE in estimate of $||\mathbf{p}||_2$ for $\sum_B$ and $\sum_A$}
\label{fig:circsimerr1}
\end{figure}

The resulting AMAE values for systems $\sum_A$ and $\sum_B$ are given in \tabref{tab:rangenoisecomp} and are plotted in \figref{fig:circsimerr1}. As expected, at very low noise values on the range measurement, both the filters for $\sum_A$ and $\sum_B$ have very similar error performance. With no noise on the range measurements, the difference between the two filters is only 4 mm.
However, since the filter for $\sum_B$ is more sensitive to noise on the range measurements, it quickly starts to perform worse than $\sum_A$ as the noise on the range measurement is increased. 


\begin{table}[]
\centering
\caption{Average Mean Absolute Error for $\sum_A$ and $\sum_B$ over 1000 runs with different noise standard deviation on the range measurement}
\label{tab:rangenoisecomp}
\begin{tabularx}{\columnwidth}{l|XXXXXXXX}
                  & \multicolumn{8}{c}{Range noise $\sigma_{R}$ [m]}\\
                  & 0   & 0.1 & 0.25 & 0.5 & 1    & 2    & 4    & 8     \\ \hline
$\sum_A$ AMAE [cm]   & 2.3 & 3.4 & 6.2  & 10.8 & 19.3 & 37.7 & 72.9 & 118.2  \\
$\sum_B$ AMAE [cm] & 2.7 & 4.5 & 8.5  & 15.1 & 27.1 & 52.5 & 101.8 & 172.8 \\
\end{tabularx}
\end{table}

This result is in line with the analytical results presented in \secref{sec:Obs}.
However, it also raises the question of whether removing the dependency on a common heading reference poses any advantage, since $\sum_A$ performs consistently better than $\sum_B$.
The reason for this result lies in the fact that the studied scenario uses perfect measurements for all the sensors \emph{except} for the measured range.
As mentioned in the introduction, the heading observation is notoriously troublesome and unreliable, especially in an indoor environment \citep{Afzal2010}.
Therefore, it would be valuable to study what would happen to this analysis in the case where the heading estimate is not perfect.
This is presented next.

\subsection{Kinematic, noisy range measurements, and heading disturbance study for observable situation}
In order to compare the results obtained with an imperfect heading measurement to those obtained in the previous section, the same trajectories are simulated (as in \eqnref{eq:p1traj} and \eqnref{eq:p2traj} for MAVs 1 and 2, respectively).
All the other simulation parameters are also kept the same, with one exception.
This time, a disturbance is introduced on the heading measurement.
The simulated disturbance is modeled to look similar to how a real local perturbation in the magnetic field would perturb a heading estimate.
The actual magnetic perturbation and the corresponding heading error are taken from the work of \cite{Afzal2010}, where indoor magnetic perturbations are studied.
It was found that the obtained disturbance on the heading estimate looks similar to a Gaussian curve, and in this analysis it is thus modeled as such. 

The disturbance on the heading estimate in time $d(t)$ is modeled as:

\begin{equation}
\label{eq:perturb}
d(t) = A_d \cdot e^{-\left(\epsilon (t-t_0)\right)^2}
\end{equation}

\begin{figure}[t!]
\centering
\includegraphics[width=0.7\columnwidth]{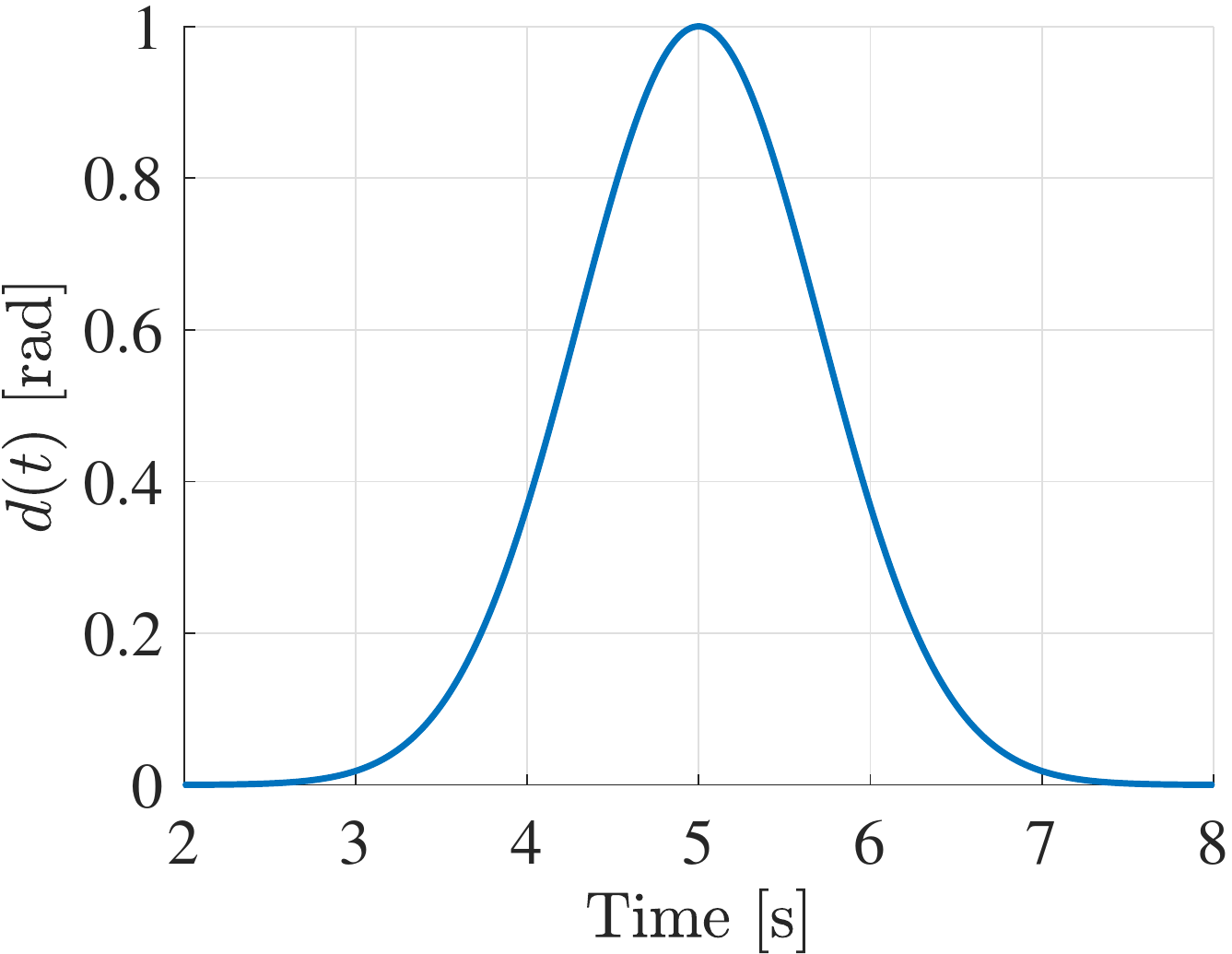}
\caption{Disturbance on the relative heading measurement in time, for an amplitude $A_d$ of 1 rad}
\label{fig:disthead}
\end{figure}

Here, the amplitude of the disturbance (in radians) is given by $A_d$, the parameter $\epsilon$ controls the width of the Gaussian curve, and $t_0$ controls the location of the curve in time.
For this study, $\epsilon=1$, resulting in a disturbance of approximately 4 seconds, and $t_0=5~s$, such that the disturbance occurs at around 5 seconds into the flight.
How such a disturbance looks is presented in \figref{fig:disthead} for an amplitude $A_d$ of 1 rad.

Several amplitudes of the disturbance are tested, namely 0, 0.25, 0.5, 1, and 1.5 rad.
The final amplitude of 1.5 results in a maximum heading estimate error of almost 85$^{\circ}$, which is approximately equal to the amplitude of the disturbance shown by \cite{Afzal2010}.
Note that the disturbance is introduced directly on the measurement of $\Delta \psi$ (the difference in headings between two MAVs).
This is the situation that would occur if one of the two MAVs would fly in a locally perturbed area. 

\begin{figure}[t!]
\centering
\includegraphics[width=0.8\columnwidth]{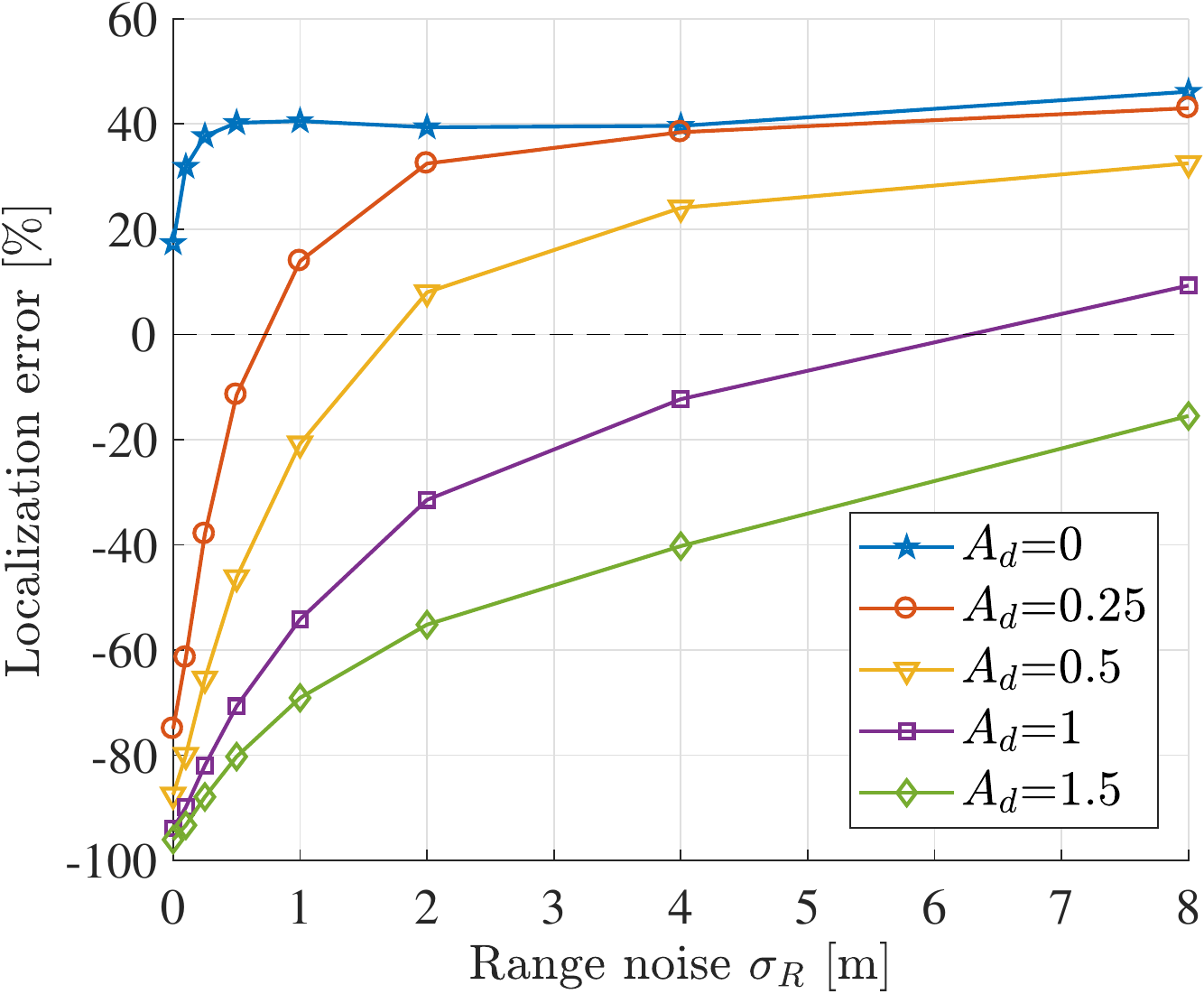}%
\caption{Percentage error comparison between $\sum_B$ and $\sum_A$ for different disturbance amplitudes $A_d$. Positive percentage means $\sum_B$ performs \textit{worse} than $\sum_A$.}
\label{fig:circsimerr2}
\end{figure}

Since the parameter of interest is how the filter for $\sum_B$ compares to the filter for $\sum_A$, the results are represented as a percentage comparison of the relative localization errors between the two filters.
This is visually presented in \figref{fig:circsimerr2}.
In the figure, a positive \% means that the filter for $\sum_B$ performs \textit{worse} than the filter for $\sum_A$.
At 0\%, marked by a dotted line, both filters perform equally well.

The comparison shows that as the applied disturbance amplitude on the heading measurement provided to system $\sum_A$ is increased, the region for which $\sum_B$ performs better than $\sum_A$ expands.
In the case of the largest disturbance, with $A_d$ equal to 1.5 radians, filter $\sum_B$ even performs better at a range noise $\sigma_R$ equal to 8.

This result reinforces the presumption that it is not always better to include a heading measurement in the filter, provided that the range measurement is of a high enough accuracy.
We will use this insight for the real-world implementation.
In the experimental set-up in \secref{sec:Experiment}, we will use Ultra Wide Band (UWB) radio modules to obtain range measurements between MAVs.
To give an idea of what type of range noise standard deviations can actually be achieved in practice, in the executed experiments with real MAVs, the UWB modules resulted in ranging errors with standard deviations between 0.1 and 0.3.
If we assume a normally distributed ranging error, based on the results hown in \figref{fig:circsimerr2}, it is then clear that the heading-independent system $\sum_B$ would be the preferred choice for all heading disturbance amplitudes
(except, trivially, for the situation where there is little to no heading disturbance at all).

\section{Leader-follower flight experiment}
\label{sec:Experiment}

In this section we demonstrate the heading-independent filter in practice, which is used for leader-follower flight in an indoor scenario.

\subsection{Leader-follower flight considerations}
\label{sec:flightconsid}

Before designing an actual control method to accomplish leader-follower flight, let's first reflect on the previous observability analysis results from \secref{sec:Obs} and their implications with respect to leader-follower flight.
We know that in order to have an observable, heading-independent, system, the combined motion of the leader and follower has to meet the observability condition presented in \eqnref{eq:sys2cond}.
We further know that in order to to meet this condition, the three intuitive conditions presented by \eqnref{eq:sys2cond1} to \eqnref{eq:sys2cond3} certainly have to be met.
Let's first consider these conditions:
\begin{enumerate*}
\item The first condition (\eqnref{eq:sys2cond1}) specifies that the relative position between leader and follower must be non-zero.
This condition has little implication to leader-follower flight, other than the fact that the follower must follow the leader at a non-zero horizontal distance, which typically is the objective.

\item The second conditions (\eqnref{eq:sys2cond2}) tells us that both MAVs must be moving. As far as leader-follower flight is concerned, this is automatically accomplished as long as the leader is not stationary.

\item The third condition (\eqnref{eq:sys2cond3}) is especially impactful for leader-follower flight.
It specifies that the MAVs should not be moving in parallel (regardless of speed), unless they are also accelerating.
A lot of research on leader-follower flight aims to design control laws that would result in fixed geometrical formations between different agents in the formation.
This is typically achieved by specifying desired formation shapes, or desired inter-agent distances for members in the swarm \citep{Turpin2012,Gu2006,Chiew2015,Saska2014}.
By the very nature of fixed geometries, that would result in parallel velocity vectors. 
\end{enumerate*}
The third condition requires a different approach to leader-follower flight. 
Rather than flying in a fixed formation, it is also possible for the follower to fly a delayed version of the leader's trajectory.
As long as the leader's trajectory is not a pure straight line for long periods of time, this will result in relative motion between the leader and follower. This is the approach taken in this paper.

This solution should also help to prevent the MAVs from getting stuck in an unobservable situation that is not covered by \eqnref{eq:sys2cond1} to \eqnref{eq:sys2cond3}, but that is covered by the full observability condition in \eqnref{eq:sys2cond}. We concluded that for the scenarios that are numerically found to be unobservable according to \eqnref{eq:sys2cond}, changing the relative position $\mathbf{p}$ only slightly can already result in an observable situation. In the proposed method of having the follower fly a time-delayed version of the leader's trajectory, the relative position vector $\mathbf{p}$ will naturally change if the leader's trajectory is not a straight line. 



\subsection{Leader-follower formation control design}
\label{sec:leadfolcd}

We want to construct a leader-follower control method that results in the follower flying a delayed version of the leader's trajectory.
As it turns out, this type of control can be directly accomplished with the information provided by the relative localization filter.

\begin{figure}
\centering
  \def\svgwidth{\linewidth}\footnotesize
  \input{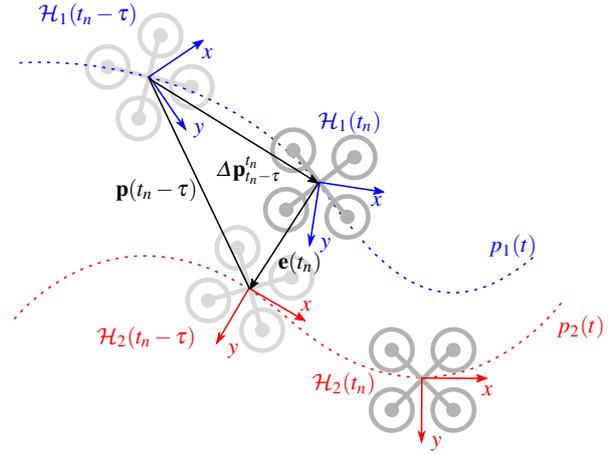}
\caption{Control problem for leader-follower flight. In blue is MAV 1's trajectory in time $\mathbf{p_1}(t)$. In orange is MAV 2's trajectory in time $\mathbf{p_2}(t)$. The desire is for MAV 1 to drive $\mathbf{e}(t)$ to $\mathbf{0}$ for $t\to \infty$. }
\label{fig:controlLF}
\end{figure}

Consider the schematic in \figref{fig:controlLF}.
It shows two arbitrary trajectories in dotted lines.
At the top, in blue, is the trajectory for MAV 1, which is represented by its position vector in time $\mathbf{p_1}(t)$.
On the bottom, in orange, is the trajectory for MAV 2, $\mathbf{p_2}(t)$. 
Suppose the desire is for the follower (MAV 1) to follow the leader's trajectory (MAV 2) with a time delay $\tau$. 
The control problem for MAV 1 can be expressed as the desire to accomplish $\mathbf{p_1}(t)=\mathbf{p_2}(t-\tau)$.

Let $t_n$ indicate the current time at which a control input must be calculated. At the current time, MAV 1 has a body fixed reference frame $\mathcal{H}_1(t_n)$, whose origin is $\mathbf{p_1}(t_n)$. At time $t_n-\tau$, MAV 1 knows the relative position of the leader in its own body fixed frame $\mathcal{H}_1(t_n-\tau)$, since this information is provided by the relative localization filter. However, for this control method to work, MAV 1 must have knowledge of where the leader's old position is at the current time $t_n$. This value of interest is depicted by the vector $\mathbf{e}(t_n)$ in \figref{fig:controlLF}; it is the positional error with respect to the desired follower's position at time $t_n$.

Let $\mathbf{R}_{\mathcal{H}_i(t_1) \mathcal{H}_i(t_2)}$ be the rotation matrix from frame $\mathcal{H}_i$ at time $t_2$, to frame $\mathcal{H}_i$ at time $t_1$, defined as:

\begin{equation}
\label{eq:rotmat2}
 \mathbf{R}_{\mathcal{H}_i(t_1) \mathcal{H}_i(t_2)} = \left[ {\begin{array}{*{20}{c}}
{cos(\Delta \psi_i |_{t_1}^{t_2})}&{-sin(\Delta \psi_i |_{t_1}^{t_2})}\\
{sin(\Delta \psi_i |_{t_1}^{t_2})}&{cos(\Delta \psi_i |_{t_1}^{t_2})}
\end{array}} \right],
\end{equation}

$\Delta \psi_i |_{t_1}^{t_2}$ is the change in heading angle for MAV $i$ from time $t_1$ to time $t_2$, which can be calculated as:

\begin{equation}
\Delta \psi_i |_{t_1}^{t_2} = \int\limits_{t_1}^{t_2} {r_i(t)dt} 
\end{equation}
 
The current positional error for the follower MAV 1, depicted in \figref{fig:controlLF}, can be defined as:

\begin{equation}
\mathbf{e}(t_n) = \mathbf{R}_{\mathcal{H}_1(t_n) \mathcal{H}_1(t_n-\tau)} \left (\mathbf{p}(t_n-\tau)-\Delta \mathbf{p}_{t_n-\tau}^{t_n} \right )
\end{equation}

The vector $\Delta \mathbf{p}_{t_n-\tau}^{t_n}$ represents how much the follower has moved from time $t_n-\tau$ until $t_n$ as defined in frame $\mathcal{H}_1(t_n-\tau)$. This vector can be calculated using information available to the follower:

\begin{equation}
\Delta \mathbf{p}_{t_n-\tau}^{t_n} = \int\limits_{t_n-\tau}^{t_n} {\mathbf{R}_{\mathcal{H}_1(t_n-\tau) \mathcal{H}_1(t)}\mathbf{v_1}(t)dt}
\end{equation}

Finally, one more piece of information is needed in order to be able to design a control law for the follower MAV, which is the model of the follower MAV and how it responds to control inputs. In this paper, it is assumed that the MAV already has stable inner loop control running on board, such that the MAV becomes an outer loop control system that directly can take velocity commands. It is further assumed that with the inner loops in place, the MAV responds like a very simple first order delay filter to velocity commands, such that the differential equation for the its velocity becomes:

\begin{equation}
\dot{\mathbf{v_1}} = \bm{\tau}^{-1}(\mathbf{v_{1c}}-\mathbf{v_1})
\end{equation}

\noindent Where $\bm{\tau}^{-1}$ is a diagonal matrix with on the diagonal the inverse values of the time constants that characterize the delay of the system with respect to a control input $\mathbf{v_{1c}}$.
This is only an approximation of how the actual MAV behaves, but it will be shown to be sufficient to accomplish the desired behavior.

With all this information in place, a control law can be designed. The control law is designed using Nonlinear Dynamic Inversion (NDI) principles. In order to use NDI, a state space model is required for the situation at hand. A very similar state space model to the one used for the relative localization filter can be used. Define the state vector as:

\begin{equation}
\mathbf{\bar{x}} = \left[\mathbf{e}^\intercal, \Delta \bar{\psi}, \mathbf{v_1}^\intercal, \mathbf{\bar{v_2}}^\intercal \right]^\intercal
\end{equation}

The state vector is similar to the one defined before for the relative localization filter, with a few small changes. First of all, $\mathbf{e}=\mathbf{e}(t)$ represents the current positional error for the follower MAV 1 with respect to the leader's old position. Secondly, $\Delta \bar{\psi}$ and $\mathbf{\bar{v_2}}^\intercal$ represent again the difference in heading between two MAVs and the velocity of MAV 2, except now $\Delta \bar{\psi}$ is the difference in heading between frame $\mathcal{H}_1(t)$ and $\mathcal{H}_2(t-\tau)$, and $\mathbf{\bar{v_2}}^\intercal$ is the delayed leader's velocity at time $t-\tau$, such that $\mathbf{\bar{v_2}}^\intercal = \mathbf{v_2}(t-\tau)$.

Similarly, define a new input vector as:

\begin{equation}
\mathbf{\bar{u}} = \left [\mathbf{v_{1c}}^\intercal, \mathbf{\bar{a_2}}^\intercal, r_1, \bar{r_2}     \right ] ^\intercal
\end{equation}

\noindent Where $\mathbf{v_{1c}}$ is the actual control input fed to MAV 1, and $\mathbf{\bar{a_2}}$ and $\bar{r_2}$ represent the same values as $\mathbf{a_2}$ and $r_2$, except delayed versions thereof. Therefore $\mathbf{\bar{a_2}}=\mathbf{a_2}(t-\tau)$ and $\bar{r_2} = r_2(t-\tau)$. 

Finally, a new set of state differential equations can be defined as:

\begin{equation}
\label{eq:ssmodel2}
\mathbf{\dot{\bar{x}}} = \mathbf{\bar{f}(\bar{x},\bar{u})}= \left[ {\begin{array}{*{20}{c}}
{-\mathbf{v_1}+\mathbf{\bar{R}}\mathbf{\bar{v_2}}-\mathbf{S_1 e}}\\
{\bar{r_2}-r_1}\\
{\bm{\tau}^{-1}(\mathbf{v_{1c}}-\mathbf{v_1})}\\
{\mathbf{\bar{a_2}-\bar{S_2} \bar{v_2}}}
\end{array}} \right]
\end{equation}

\noindent Where $\mathbf{\bar{R}}=\mathbf{R}(\Delta \bar{\psi})$ and  $\mathbf{\bar{S_2}}=\mathbf{S_2}(\bar{r_2})$.

The state that we wish to control is the current positional error that MAV 1 has with respect to the delayed leader's position, so the state $\mathbf{e}$. This state can be represented as:
\begin{equation}
\mathbf{e} = \mathbf{H} \mathbf{\bar{x}} 
\end{equation}

With $\mathbf{H}$ given by:
\begin{equation}
\mathbf{H} = \left[ {\begin{array}{*{20}{c}}
{\mathbf{I}_{2\mathtt{x}2} }&{\mathbf{0}_{2\mathtt{x}5}}
\end{array}} \right] 
\end{equation}

The derivative of the control variable with respect to time is equal to:
\begin{equation}
\dot{\mathbf{e}} = \mathcal{L}_{\mathbf{\bar{f}}}^1 \mathbf{e} =  \mathbf{H} {\mathbf{\bar{f}}} = -\mathbf{v_1}+\mathbf{\bar{R}}\mathbf{\bar{v_2}}-\mathbf{S_1 e}
\end{equation}

The second derivative of the control variable:
\begin{align}
\nonumber
\ddot{\mathbf{e}} &= \mathcal{L}_{\mathbf{\bar{f}}}^2 \mathbf{e} = (\nabla \otimes \dot{\mathbf{e}}) \cdot \mathbf{\bar{f}} \\\nonumber
&= \left[ {\begin{array}{*{20}{c}}
{-\mathbf{S_1}}&{\frac{\partial \mathbf{\bar{R}}}{\partial \Delta \bar{\psi}}\mathbf{\bar{v_2}}} & {-\mathbf{I}_{2\mathtt{x}2}} & {\mathbf{\bar{R}}}
\end{array}} \right]\cdot \mathbf{\bar{f}} \\\nonumber
&= -\mathbf{S_1} \left (  -\mathbf{v_1}+\mathbf{\bar{R}}\mathbf{\bar{v_2}}-\mathbf{S_1 e}   \right) + \frac{\partial \mathbf{\bar{R}}}{\partial \Delta \bar{\psi}}\mathbf{\bar{v_2}} \left(\bar{r_2}-r_1 \right) \\\nonumber
&-\mathbf{I}_{2\mathtt{x}2} \left(    \bm{\tau}^{-1}(\mathbf{v_{1c}}-\mathbf{v_1}) \right) + \mathbf{\bar{R}} \left (  \mathbf{\bar{a_2}-\bar{S_2} \bar{v_2}}   \right) \\
\label{eq:ddote}
& = \mathbf{D}\mathbf{v_{1c}} + \mathbf{b(x,u)}
\end{align}

With $\mathbf{D}$ equal to:
\begin{equation}
\mathbf{D} =  -\mathbf{I}_{2\mathtt{x}2} \bm{\tau}^{-1}
\end{equation}

and $\mathbf{b(x,u)}$ equal to:
\begin{align}
\nonumber
\mathbf{b(x,u)} &= -\mathbf{S_1} \left (  -\mathbf{v_1}+\mathbf{\bar{R}}\mathbf{\bar{v_2}}-\mathbf{S_1 p}   \right) \\\nonumber
&+ \frac{\partial \mathbf{\bar{R}}}{\partial \Delta \psi}\mathbf{\bar{v_2}} \left(\bar{r_2}-r_1 \right) \\
&+\mathbf{I}_{2\mathtt{x}2} \bm{\tau}^{-1}\mathbf{v_1} + \mathbf{\bar{R}} \left (  \mathbf{\bar{a_2}-S_2 v_2}   \right)
\end{align}

This can further be reduced to:

\begin{align}
\nonumber
\mathbf{b(x,u)} &= -\mathbf{S_1} \left (  -\mathbf{v_1}+\mathbf{\bar{R}}\mathbf{\bar{v_2}}-\mathbf{S_1 p}   \right) \\
&- \frac{\partial \mathbf{\bar{R}}}{\partial \Delta \psi}\mathbf{\bar{v_2}} r_1  +\mathbf{I}_{2\mathtt{x}2} \bm{\tau}^{-1}\mathbf{v_1} + \mathbf{\bar{R}}  \mathbf{\bar{a_2}}   
\end{align}

At this point the following control law can be chosen:
\begin{equation}
\label{eq:controllaw}
\mathbf{v_{1c}} = \mathbf{D}^{-1} (\mathbf{i} - \mathbf{b(x,u)})
\end{equation}

\noindent with $\mathbf{i}$ now a virtual control input.

This control law results in a fully linearized differential equation for the positional error of the follower, since substitution of the control law from  \eqnref{eq:controllaw} in \eqnref{eq:ddote} results in the following differential equation:
\begin{equation}
\mathbf{\ddot{e}} = \mathbf{i}
\end{equation}

Which can be shown to be exponentially stable if the following virtual control is implemented:
\begin{align}
\mathbf{i} = -K_p \mathbf{e} - K_d \mathbf{\dot{e}}\\
K_p, K_d > 0
\end{align}

\subsection{Experimental Set-Up}
\label{sec:expsu}


One of the main findings in the observability study and the simulation results is that the localization error scales more steeply with range noise for system $\sum_B$ than for $\sum_A$.
It is therefore important to use sensors that can provide accurate relative ranging measurements.

In this work, we chose to use Ultra Wide Band (UWB) based radio transceivers.
UWB has recently gained attention within the domain of ranging.
UWB signals are characterized by their fine temporal and spatial resolution \citep{Correal2003}, which leads UWB based systems to be able to, for example, resolve multipath effects more easily \citep{Win1998}.
Ultimately, this leads to an accurate ranging performance, which is important if using the heading independent filter.
Another advantage of UWB is its relative robustness to interference from other radio technologies due to the fact that it operates on an (ultra) wide range of frequencies \citep{Liu2007,Foerster2001,Molisch2006}.




The UWB ranging hardware used in the experiments is the ScenSor DWM1000 module sold by Decawave.\footnote{\url{https://www.decawave.com/products/dwm1000-module}}
The ranging algorithm that is employed is a particular implementation of the Two-Way Ranging (TWR) method \citep{Neirynck2016}.
In order to fuse ranging data with velocity, acceleration, height, and yaw rate data in the localization filter, these variables are also communicated between MAVs by using the UWB devices.
The same UWB messages used in the TWR protocol are also used to communicate these variables.

The UWB module transceiver has been installed on the Parrot Bebop 2 platform.
\footnote{\url{https://www.parrot.com/us/drones/parrot-bebop-2}}
The Bebop 2 runs custom autopilot software designed using the open-source autopilot framework Paparazzi UAV.
\footnote{\url{http://wiki.paparazziuav.org/wiki/Main_Page}}
Paparazzi UAV provides the stable inner loop control loops for the Bebop 2 using Incremental NDI (INDI).
This allows us to control the outer loop by giving the computed velocity commands to the INDI inner loops.


Velocity and height measurements are also necessary for the relative localization filter.
In the initial experiments, they are provided by an overhead Motion Capture System (MCS) by OptiTrack.\footnote{\url{http://optitrack.com/}}
In a second iteration of the experiment, they are fully provided by on-board sensors.
The velocity data is obtained from the MAVs' on-board bottom-facing camera using Lucas-Kanade optical flow.
Height is measured using an on-board ultrasonic sensor that the Bebop 2 is equipped with by default.
At all times, the acceleration and yaw rate measurements are obtained from the MAVs' on-board accelerometers and gyroscope, respectively.
The experiments are first conducted with two MAVs (one leader and one follower), detailed in \secref{sec:leadfolresults}, and then performed again with three MAVs (one leader and two followers), detailed in \secref{sec:threefollowers}.

\subsection{Leader-follower flight with one follower}
\label{sec:leadfolresults}
The experiment with one follower MAV consists of one Bebop 2 following another Bebop 2 using the control law presented in \secref{sec:leadfolcd}.
At first, right after take off, the MAVs fly concentric circles just like the ones shown in \figref{fig:dcirctraj}.
This procedure is there to make sure that the EKF running on-board the MAVs has time to converge to the correct result, such that by the time the follower MAV is instructed to start following the leader, the follower has a correct estimate of the relative location of the leader.

When leader-follower flight is engaged, the trajectory of the leader has been designed to sufficiently excite the the relative localization filter during the leader-follower flight and to decrease the likelihood of being stuck in unobservable states. This has been done by introducing frequent turns in the trajectory to have changing relative velocities and accelerations. The follower is instructed to follow the leader's trajectory with a time delay of $\tau=5$ seconds. 

It is important to note that, for safety reasons, the norm of the follower's commanded velocity $||\mathbf{v_{1c}}||_2$ during both experiments is saturated at 1.5 m/s.
The measure is taken because the MAVs were flying in a relatively small confined area (10 m by 10 m).
This change does however have consequences for the performance of the follower's tracking, which is discussed further in the next sections.

\subsubsection{Leader-follower flight with velocity and height information from a MCS} 
\begin{figure}[t!]
\centering
\includegraphics[width=0.7\columnwidth]{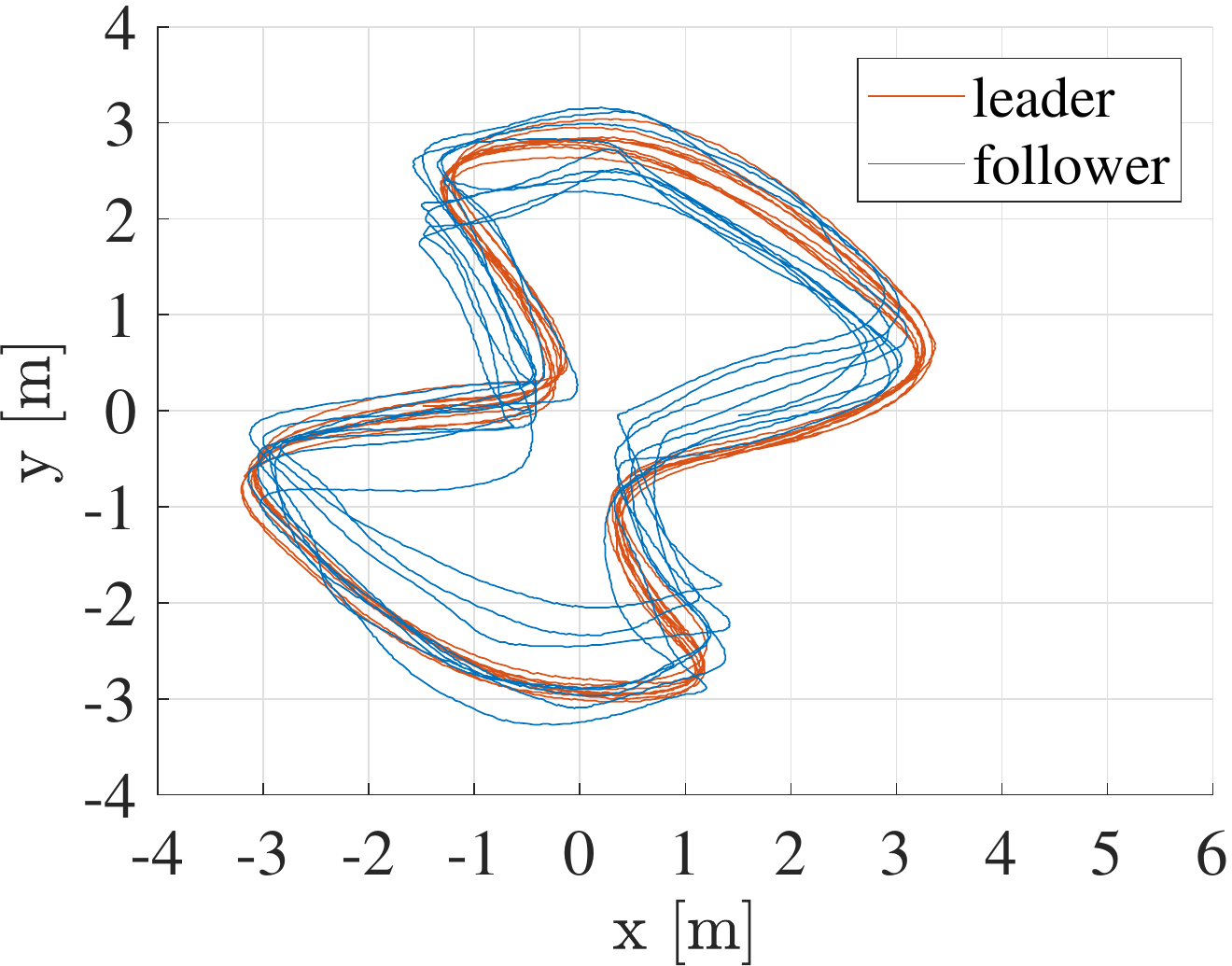}
\caption{The trajectories of leader and follower during experiment with MCS height and velocity}
\label{fig:NDI_GPS_traj}
\end{figure}

\begin{figure}[t!]
  \centering
  \begin{subfigure}[t]{39mm}
    \centering
    \includegraphics[width=\textwidth]{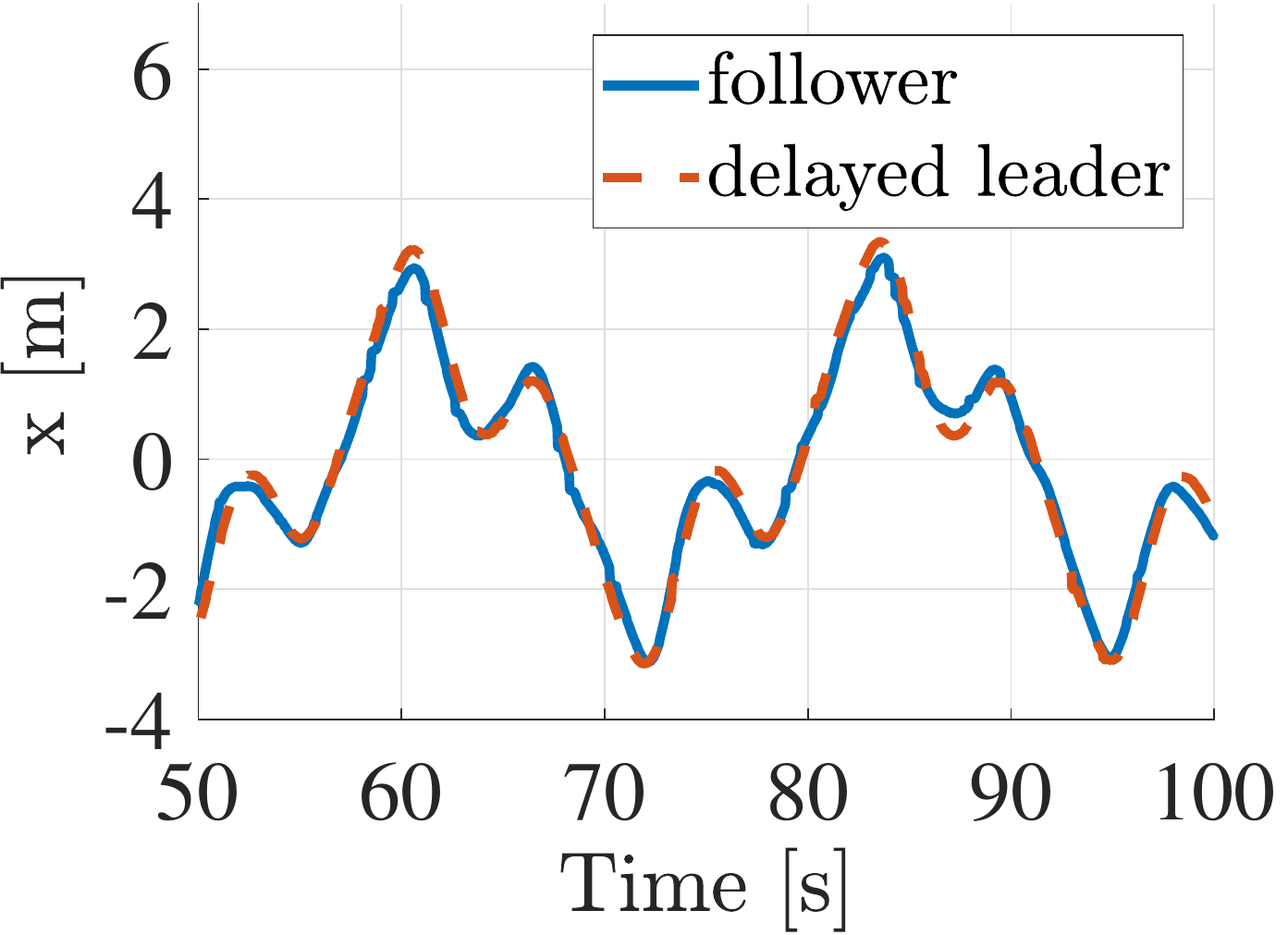}%
    \caption{}
  \label{fig:NDI_GPS_x}
  \end{subfigure}
  ~
  \begin{subfigure}[t]{39mm}
    \centering
    \includegraphics[width=\textwidth]{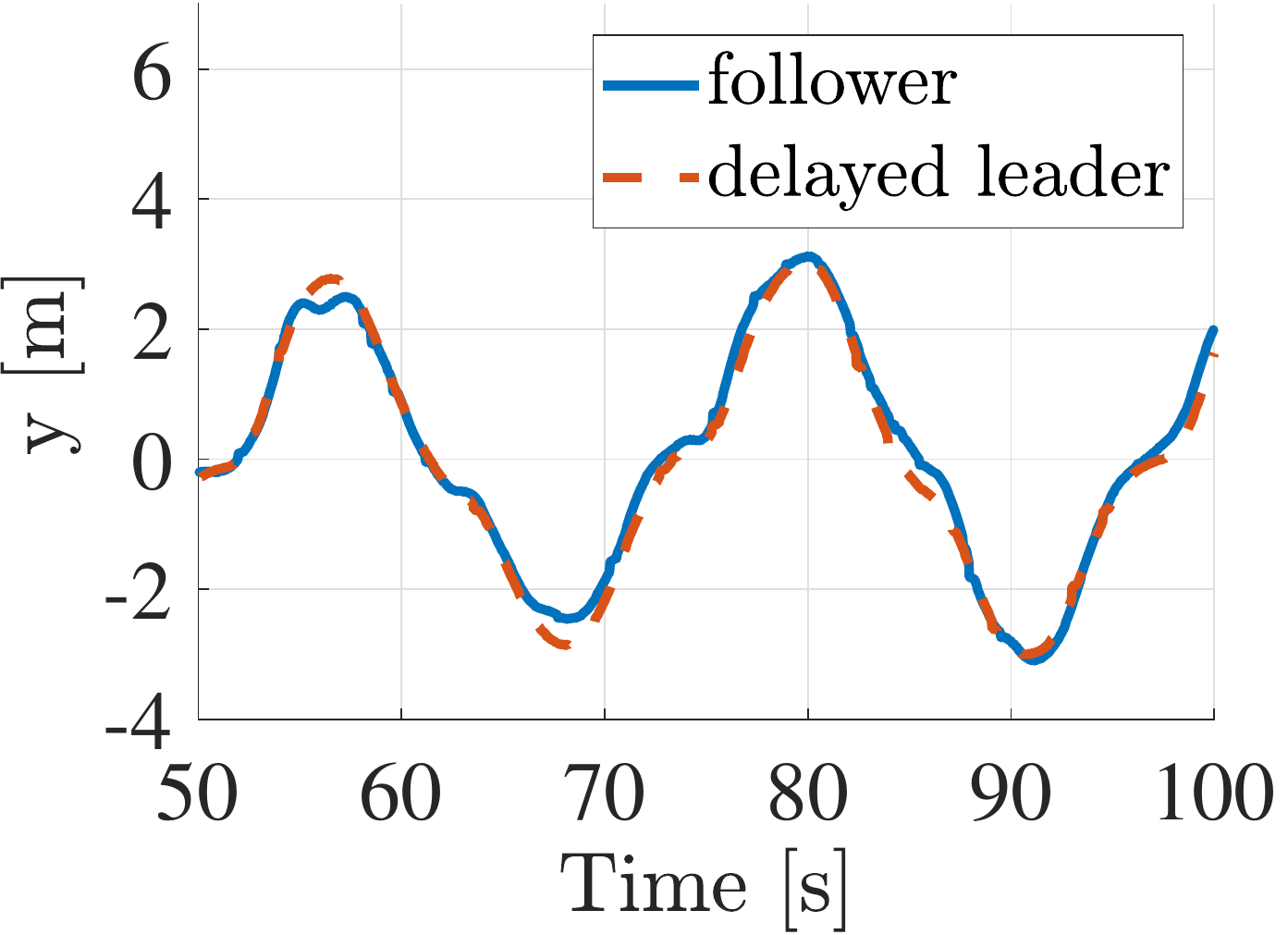}%
    \caption{}
\label{fig:NDI_GPS_y}
\end{subfigure}
\caption{The trajectory of the follower compared to the delayed trajectory of the leader for the experiment with MCS height and velocity.}
\label{fig:NDI_GPS}
  \end{figure}

\begin{figure}[t!]
\centering
\includegraphics[width=0.9\columnwidth]{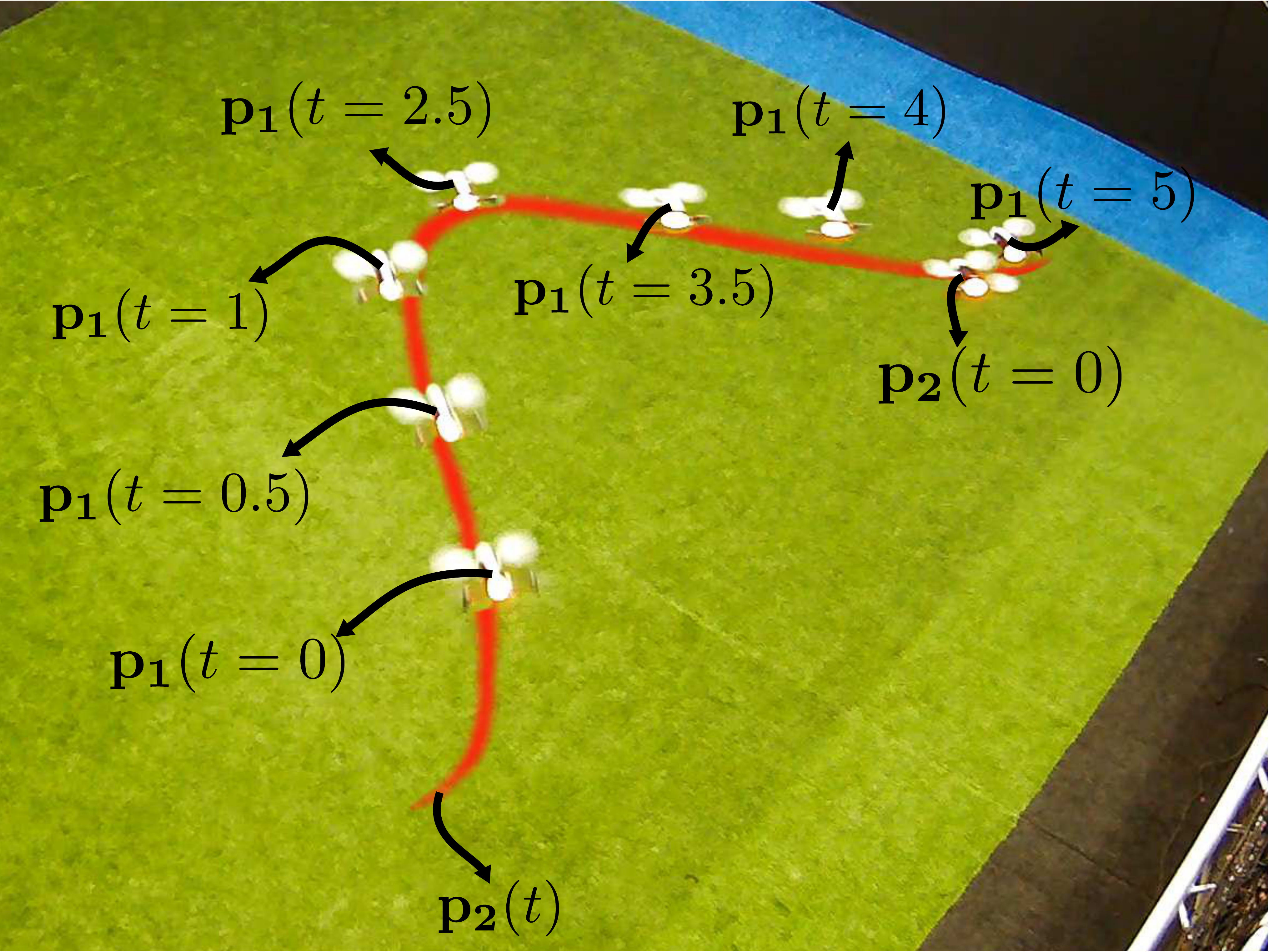}
\caption{Time composition of overhead camera images of leader and follower MAV in time, for the experiment with MCS height and velocity. Indicated in orange and marked by $\mathbf{p_2}(t)$, is part of the leader's trajectory. The leader's final position is indicated by $\mathbf{p_2}(t=0)$. Seven points in time of the follower's trajectory are indicated in the image. According to the control objective, $\mathbf{p_1}(t=5)$ should equal $\mathbf{p_2}(t=0)$. }
\label{fig:NDI_GPS_camera}
\end{figure}

First, the case where velocity and height information is provided by the MCS is studied.
In \figref{fig:NDI_GPS_traj} the trajectory flown by the follower is compared to the trajectory of the leader.
The $x$ and $y$ coordinates are compared separately for part of the flight in \figref{fig:NDI_GPS_x} and \figref{fig:NDI_GPS_y}.
In \figref{fig:NDI_GPS_camera}, a time composition of overhead camera images is given for 5 seconds of flight as an illustration.
The follower's position is shown at seven time instances during these 5 seconds, and is compared to the leader's trajectory.

\begin{figure}[t!] 
\centering
\includegraphics[width=0.7\columnwidth]{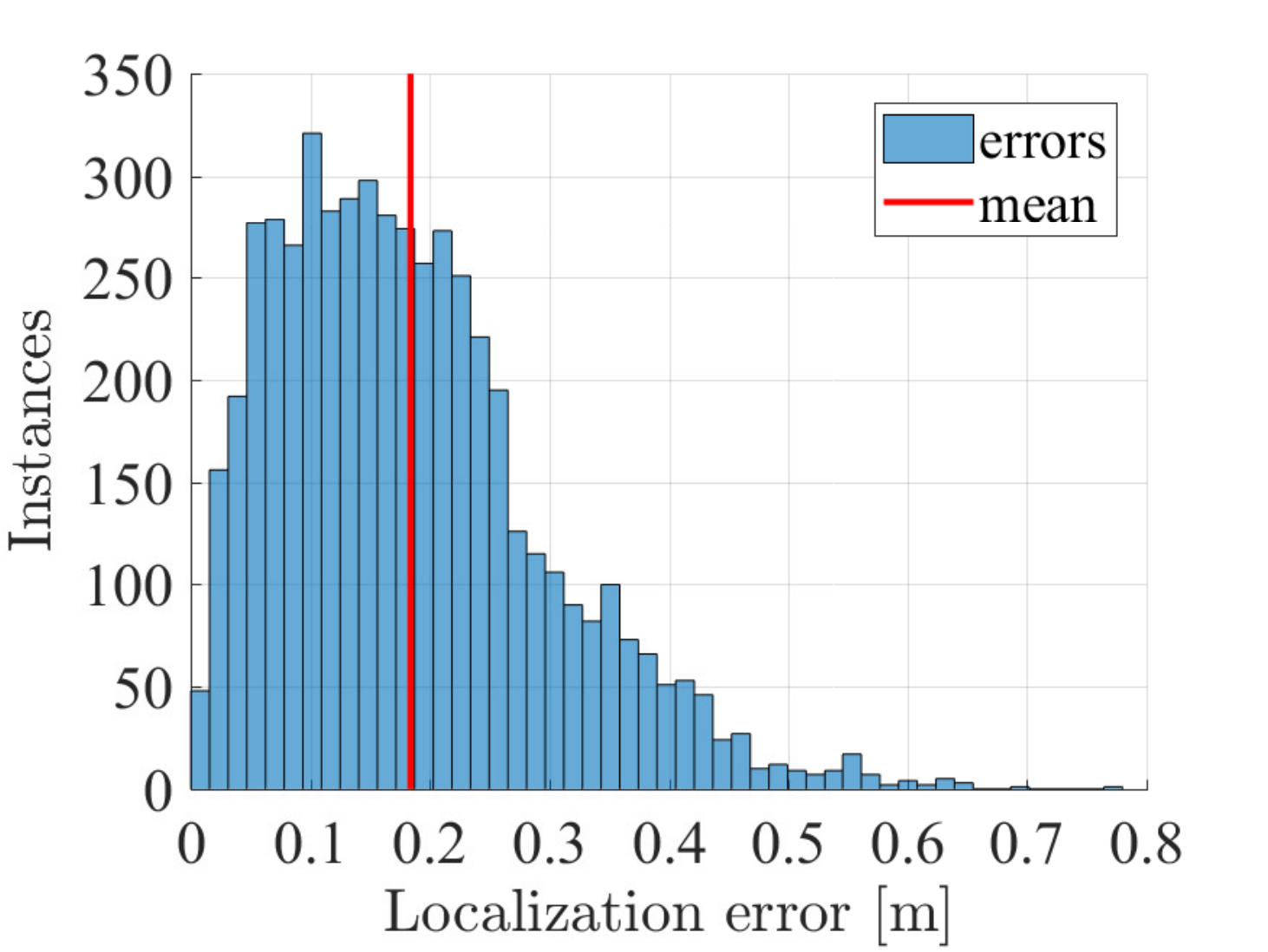}
\caption{Histogram of the localization error for the follower during experiment with MCS height and velocity}
\label{fig:NDI_GPS_plocerr}
\end{figure}

A total of 200 seconds of leader-follower flight were logged and will be analyzed here.
During this time, several laps of the designed trajectory were executed.
The trajectories in \figref{fig:NDI_GPS_traj} to \figref{fig:NDI_GPS_camera} indeed show that the follower is successfully tracking a delayed version of the leader's trajectory.
The actual error distribution for the norm of the relative location estimate $||\mathbf{p}||_2$ is shown in \figref{fig:NDI_GPS_plocerr}. The errors have a mean value of 18.4 cm and a maximum value of 77.5 cm, at maximum inter-MAV distances up to 5 m. 

\begin{figure}[t!] 
\centering
\includegraphics[width=0.7\columnwidth]{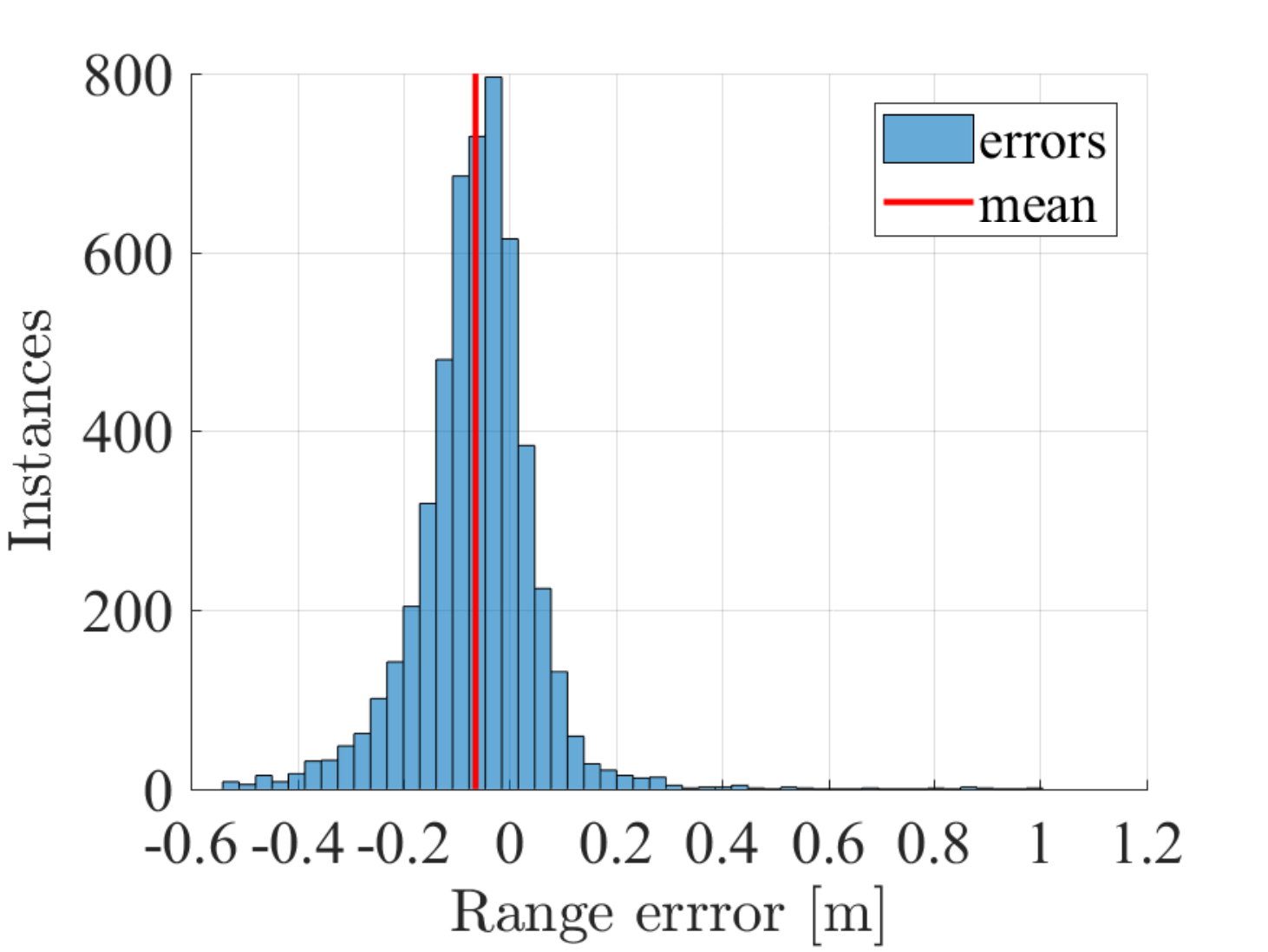}
\caption{Histogram of the ranging error during experiment with MCS height and velocity}
\label{fig:NDI_GPS_rangeerr}
\end{figure}

Since, in this experiment, the velocity and height measurements were provided with high accuracy by the MCS, one would expect the primary source for the localization error to be the ranging error from the UWB modules.
However, inspection of the ranging error actually shows a pretty favorable error distribution.
A histogram of the ranging error throughout the flight is given in \figref{fig:NDI_GPS_rangeerr}.
The mean of the ranging error is close to zero (about -6.4 cm) and the errors are well distributed around this mean.
This is therefore not the main cause of the occasionally higher relative localization errors.

The most clearly identifiable cause for the relative localization error is the occasional dropping of frames by the UWB modules.
The average update rate of the relative localization filter is about 25 Hz, corresponding to a time step of approximately 40 ms.
The update rate is established by the rate at which the UWB modules produce a new ranging result. However, the modules occasionally drop frames, causing the time step to spike up. The largest time step recorded during the flight is 470 ms, an order of magnitude larger than the average. It is not hard to imagine the unfavorable effect this can have for the relative localization estimate. It is therefore not coincidental that the largest localization error recorded during the flight also corresponds to one of those times where the UWB modules dropped frames.

\begin{figure}[t!] 
\centering
\includegraphics[width=0.7\columnwidth]{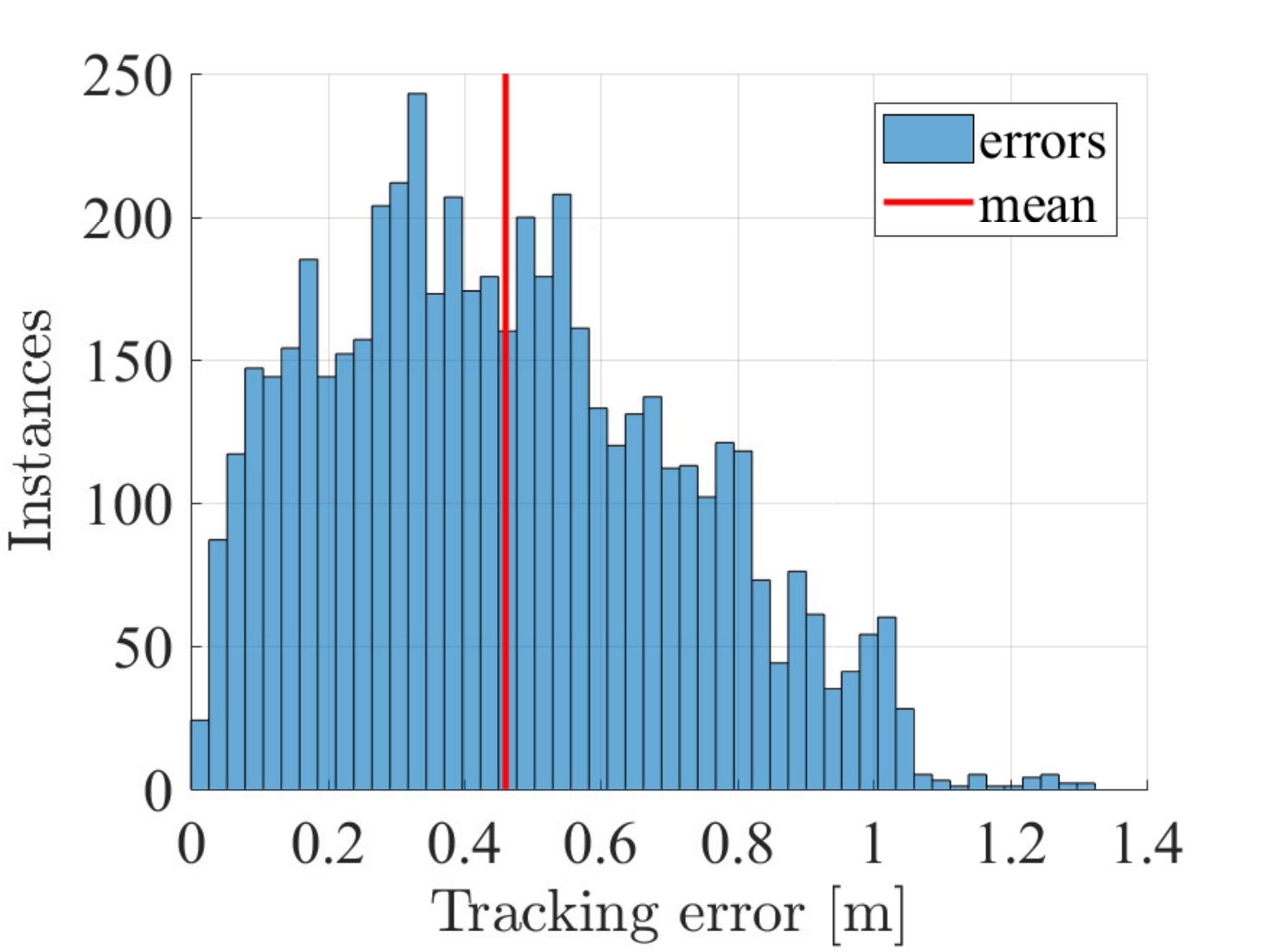}
\caption{Histogram of the tracking error $||\mathbf{e}||_2$ for the follower during experiment with MCS height and velocity}
\label{fig:NDI_GPS_ptrackerr}
\end{figure}

We now turn our attention to the tracking error of the follower MAV.
The tracking error distribution $||\mathbf{e}||_2$ is given in \figref{fig:NDI_GPS_ptrackerr}. The mean of the distribution is equal to 46.1 cm and the maximum error is 1.32 m.
Of course, part of this error is caused by a relative localization error from the follower's perspective, which will inevitably affect the tracking performance. However, since the relative localization error is considerably lower than the tracking error, there must be more sources to the error.

One source of error is the fact that the follower's response to a velocity command $\mathbf{v_{1c}}$ is modeled as a first order delay.
In reality, the MAV has some overshoot with respect to commands, which is not modeled by this first order delay.
This model mismatch by itself might not be that harmful to the performance, since the control law would respond with more aggressive velocity commands as a reaction to the MAV not behaving as modeled.
However, the control law's freedom is severely restricted by the command saturation at 1.5 m/s, which means that the follower cannot move as fast as the command law demands.
This argument is further supported by a qualitative analysis of the follower's trajectory with respect to the leader's trajectory in \figref{fig:NDI_GPS_traj}.
The trajectory of the follower often seems to take `shortcuts' with respect to the leader's trajectory.
This falls in line with the expected behavior due to the command saturation. The control law is designed not only to track the trajectory of the leader in space, but also in time.
As the follower starts lagging behind the leader more than the desired $\tau=5$ seconds, the follower starts to take shortcuts in the trajectory to catch up with the leader.
This error would be less prevalent if the command saturation were increased.

\subsubsection{Leader-follower flight with only on-board measurements}

We now demonstrate the workings of the proposed methods in this paper when \textit{only on-board sensing} is used.
In this set-up, the follower MAV does not use any MCS information.
Instead, the velocity information comes from Lucas-Kanade optical flow measurements while the height is derived from the on-board ultrasonic sensor.
Similarly, the leader MAV directly communicates optical flow velocities and ultrasonic height measurements 
(along with accelerations and yaw rate from the IMU)
to the follower MAV for use in the relative localization filter.
The MCS is only used to log ground truth data and for the leader to safely fly its trajectory.
No MCS data is used by the follower at all.
Again, 200 s of leader-follower flight with full on-board sensing took place successfully and will be analyzed here.

\begin{figure}[t!]
\centering
\includegraphics[width=0.7\columnwidth]{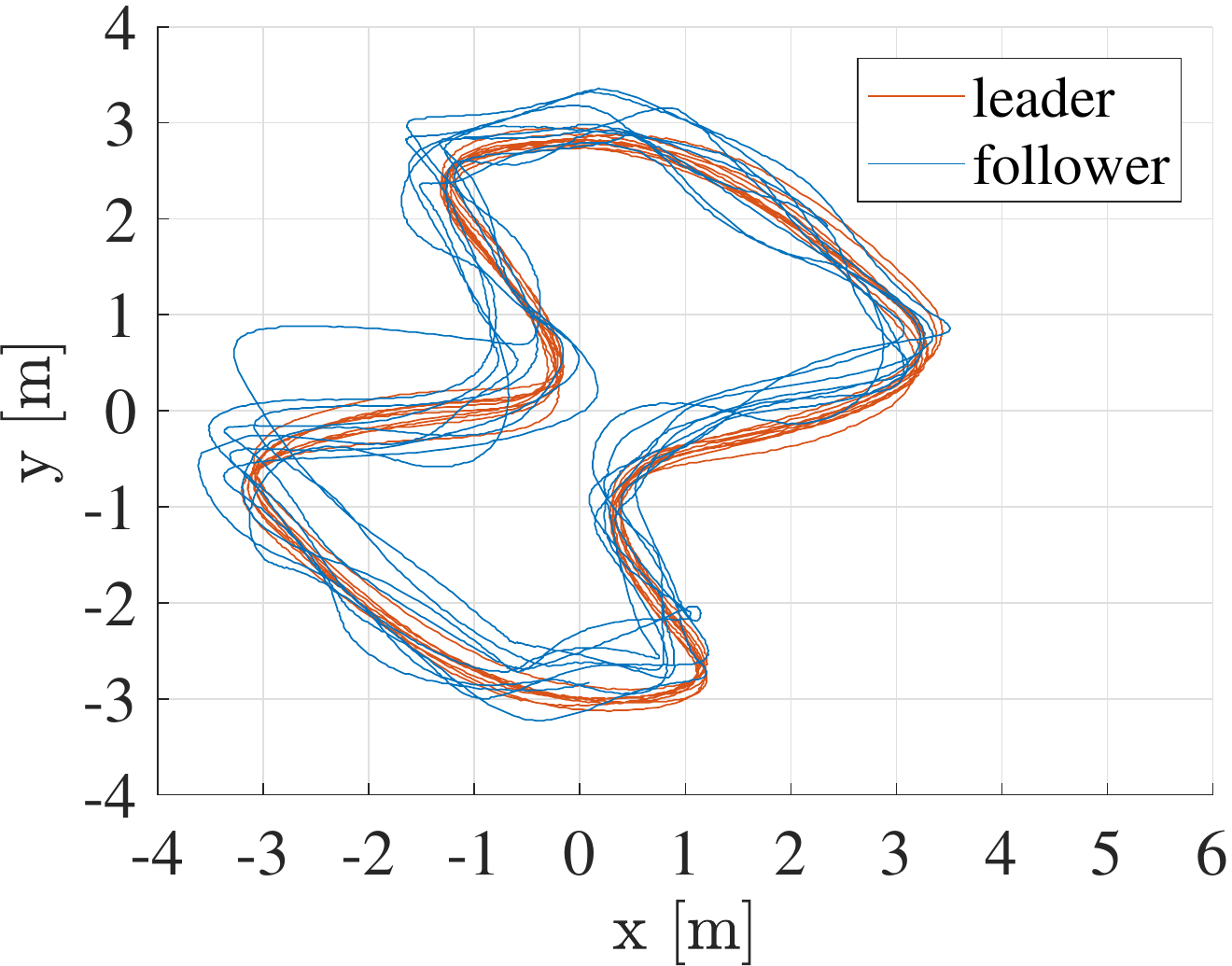}
\caption{Trajectory of leader and follower during experiment with only on-board sensing and processing}
\label{fig:NDI_onboard_traj}
\end{figure}

\begin{figure}[t!]
  \centering
  \begin{subfigure}[t]{39mm}
    \centering
    \includegraphics[width=\textwidth]{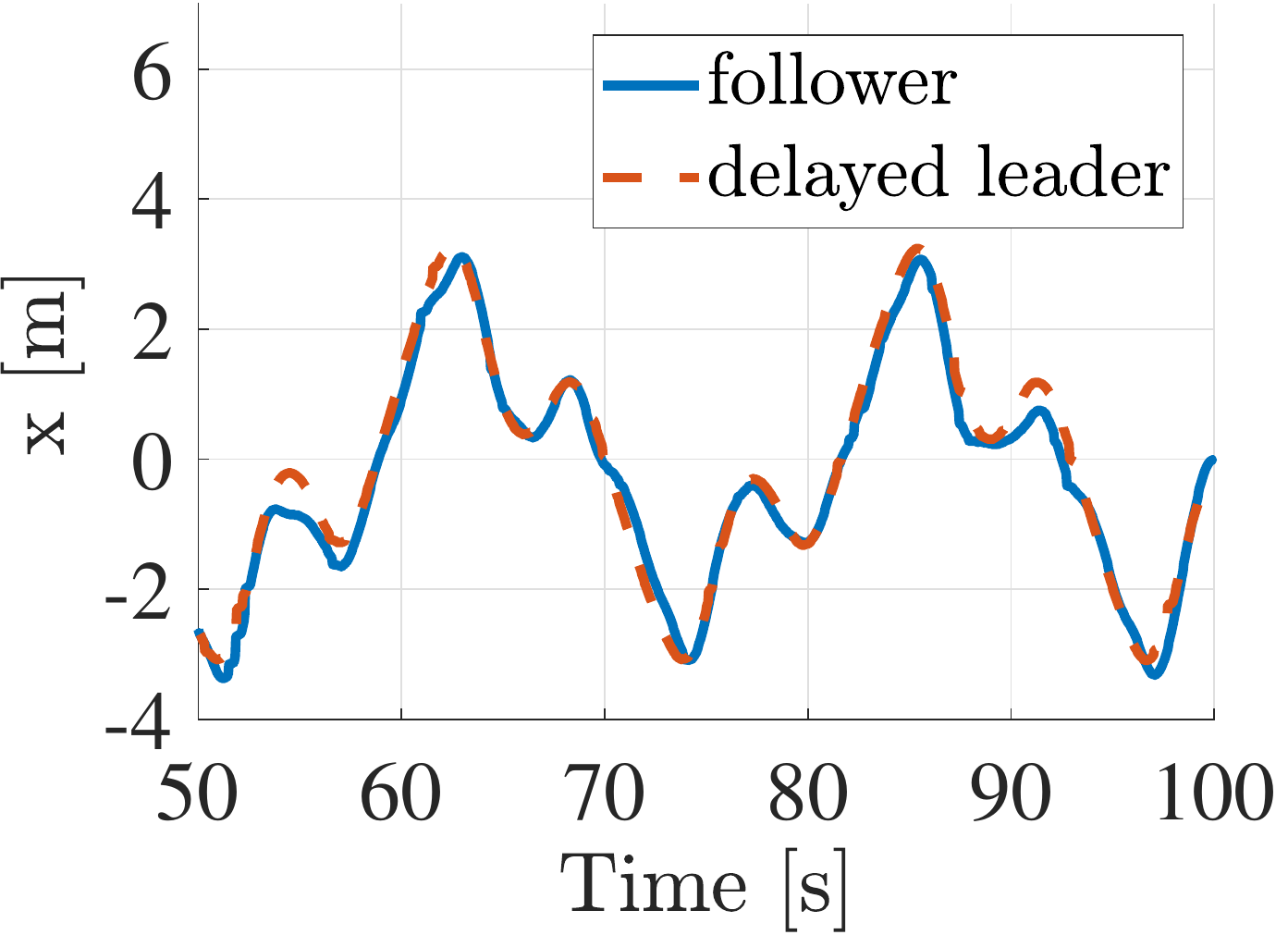}%
\label{fig:NDI_onboard_x}
  \end{subfigure}
  ~
  \begin{subfigure}[t]{39mm}
    \centering
    \includegraphics[width=\textwidth]{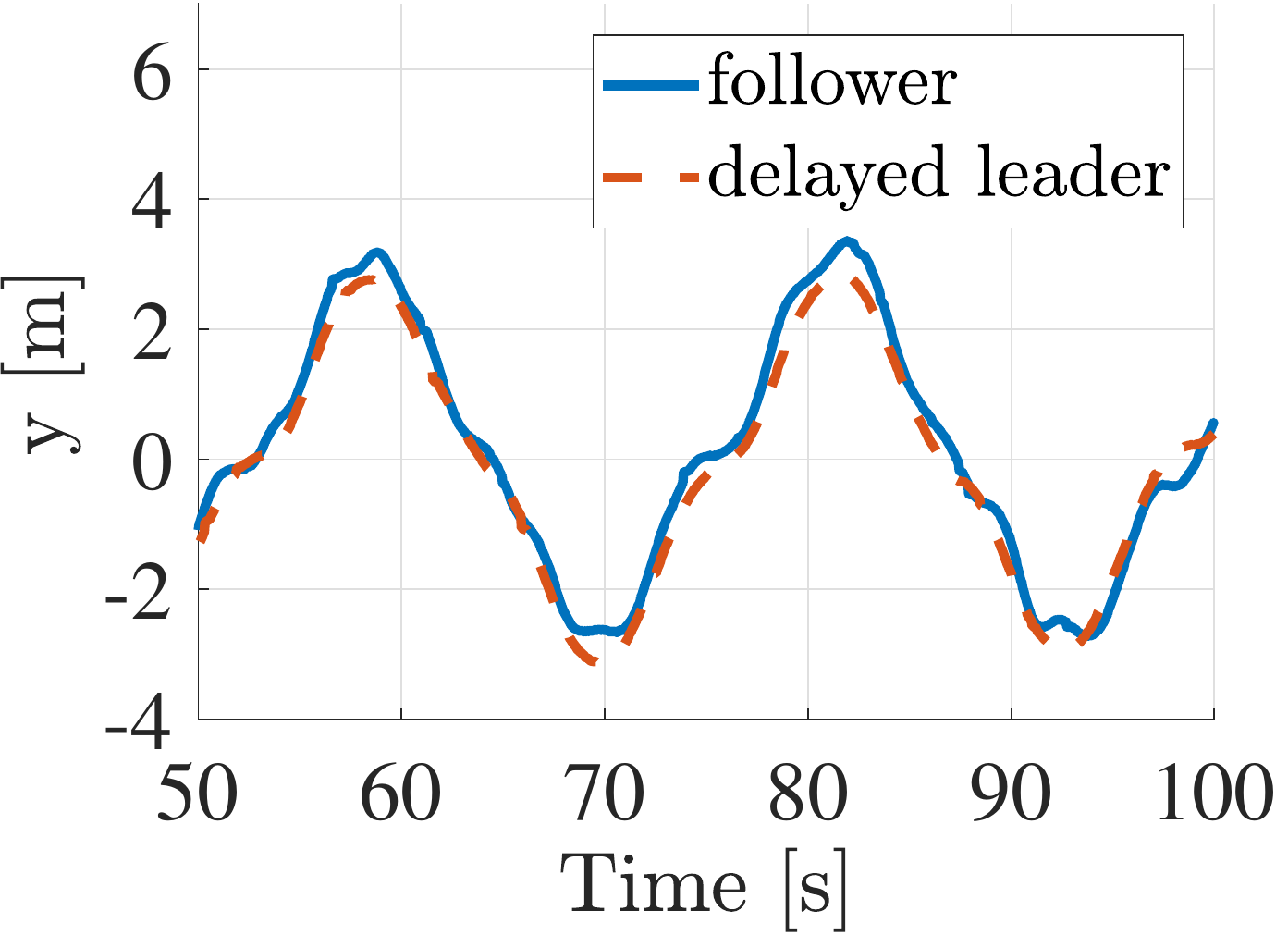}%
\label{fig:NDI_onboard_y}
\end{subfigure}
\caption{The trajectory of the follower compared to the delayed trajectory of the leader for the experiment with only on-board sensing.}
\label{fig:NDI_onboard}
  \end{figure}

\begin{figure}[t!]
\centering
\includegraphics[width=0.9\columnwidth]{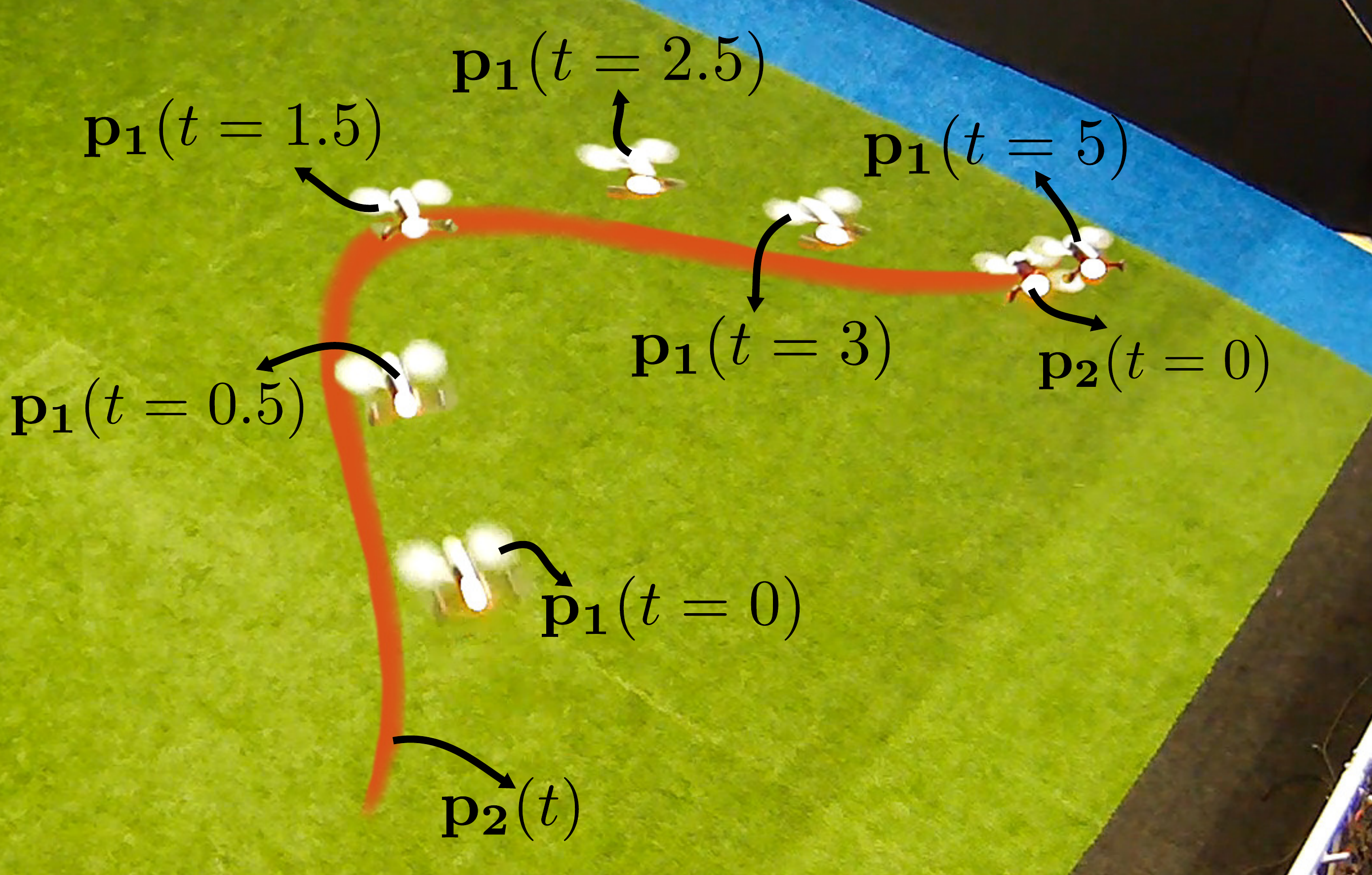}
\caption{Time composition of overhead camera images of leader and follower MAV in time, for the experiment with only on-board sensing. Indicated in orange and marked by $\mathbf{p_2}(t)$, is part of the leader's trajectory. The leader's final position is indicated by $\mathbf{p_2}(t=0)$. Six points in time of the follower's trajectory are indicated in the image. According to the control objective, $\mathbf{p_1}(t=5)$ should equal $\mathbf{p_2}(t=0)$. }
\label{fig:NDI_Onboard_camera}
\end{figure}

The trajectory of the follower with respect to the delayed leader's trajectory is compared in \figref{fig:NDI_onboard_traj} and \figref{fig:NDI_onboard}. Furthermore, another time composition for 5 seconds of flight where the follower is tracking the leader is given in \figref{fig:NDI_Onboard_camera}.

The main qualitative difference with respect to the situation where the MCS was still used for velocity and height information is the fact that the follower's trajectory appears less smooth.
Otherwise, the performance seems qualitatively similar.
The follower still appears to take `shortcuts' with respect to the leader's trajectory, although the increased disorder in the follower's trajectory makes this less apparent. 

%

\begin{figure}[t!] 
\centering
\includegraphics[width=0.7\columnwidth]{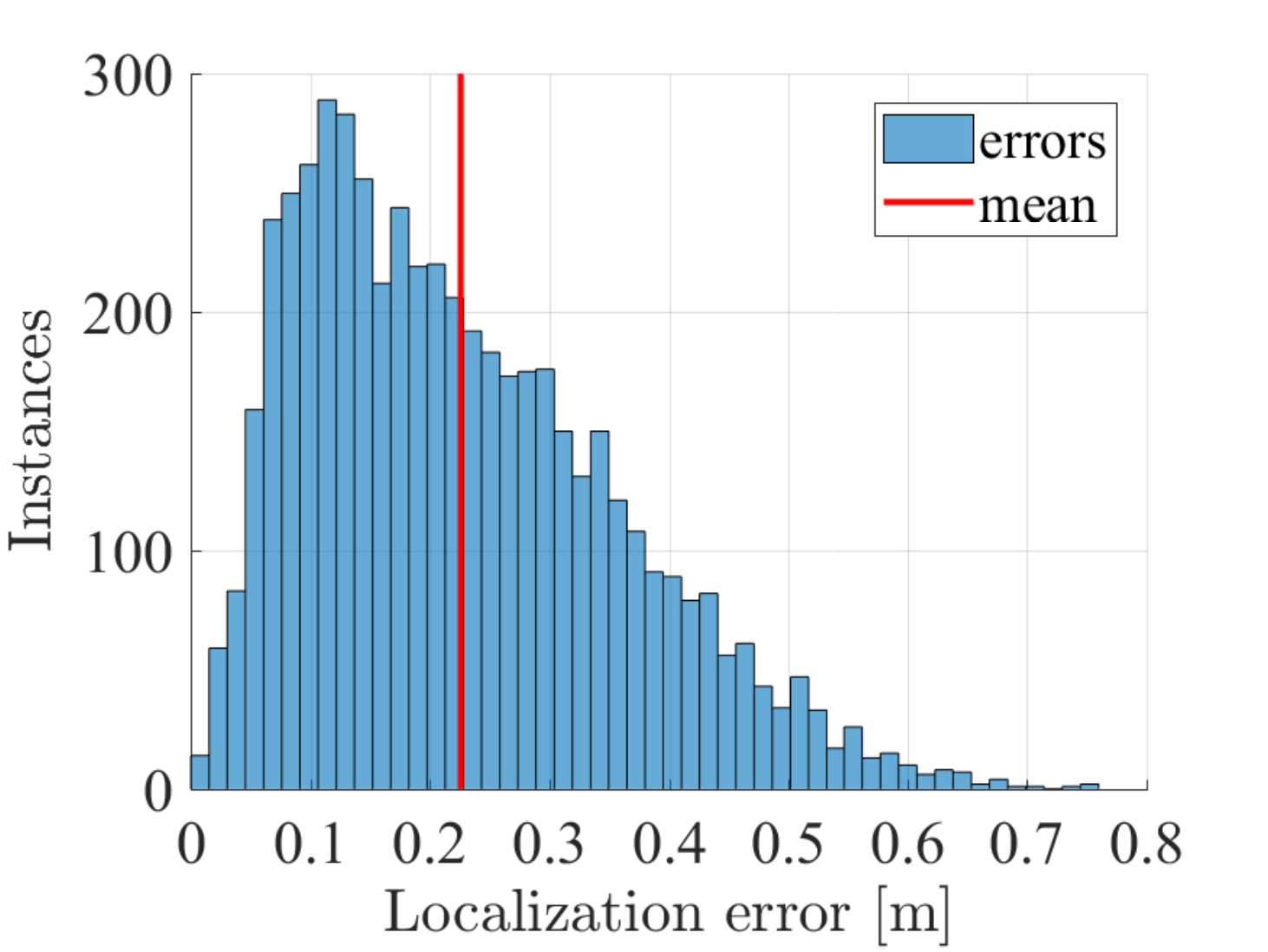}
\caption{Histogram of the localization error for the follower during experiment with only on-board sensing and processing}
\label{fig:NDI_onboard_plocerr}
\end{figure}

\begin{figure}[t!] 
\centering
\includegraphics[width=0.7\columnwidth]{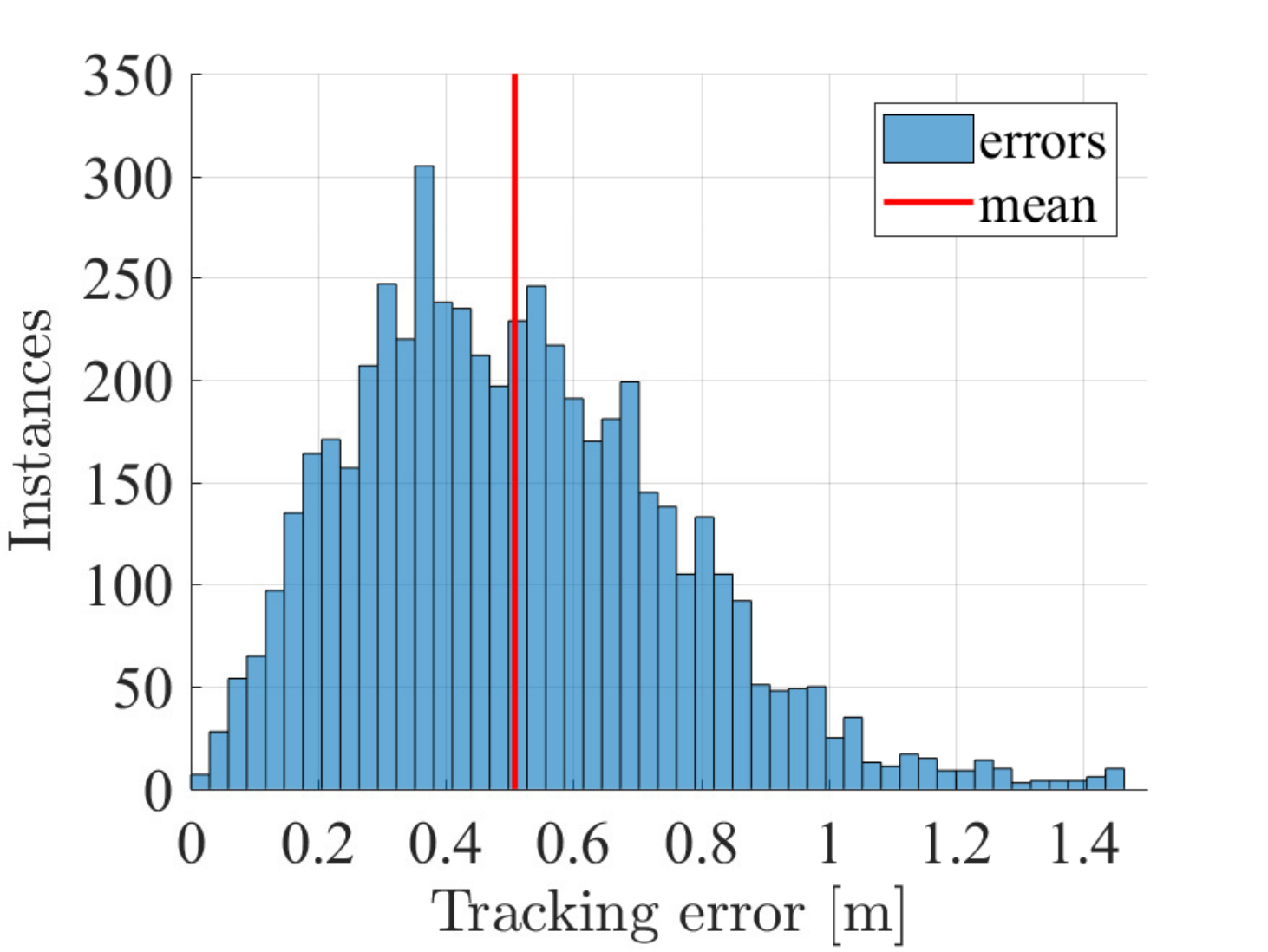}
\caption{Histogram of the tracking error $||\mathbf{e}||_2$ for the follower during experiment with only on-board sensing and processing}
\label{fig:NDI_onboard_ptrackerr}
\end{figure}

The tracking error distribution for the on-board sensing case is given in \figref{fig:NDI_onboard_ptrackerr}.
The mean tracking error is 50.8 cm and the maximum error is 1.47 m.
The relative localization error is given in \figref{fig:NDI_onboard_plocerr}. 
Here, the mean error is 22.6 cm and the maximum error is 75.8 cm, at maximum MAV distances up to 5.2 meters.

The performance when using only on-board sensing is very similar to when using the MCS for height and velocity data.
This can be mainly attributed to the fact that the measurements that have been replaced (the height and velocity of both MAVs) are actually also accurately measured on-board.

The primary reason as to why the trajectory of the follower with on-board sensors still seems slightly more disordered is the fact that the follower has difficulty to accurately control its altitude when using only on-board sensing.
The follower now purely relies on height measurements from its ultrasonic sensor.
The update rate of this sensor is low, and in between measurements the follower uses (noisy) accelerometer data to update its height.
This sometimes causes the follower to believe its altitude is different than it really is, causing it to rapidly ascend or descend.
This takes up thrust, restricting the follower's ability to maneuverer accurately in the horizontal plane due to thrust saturation.

%

\subsection{Leader-follower flight with two followers}
\label{sec:threefollowers}

To demonstrate that the methods in this paper can also scale to more than one follower, the leader-follower flight is also performed with two follower MAVs instead of one.
This is done both with MCS height and velocity data and with only on-board sensing.

For this purpose, The UWB messaging protocol is adapted to allow every MAV to perform ranging with every other MAV.
The MAVs also communicate a unique (pre-assigned) identification number within the UWB messages.
The followers can use this identification number to determine which messages originate from the leader so that they individually keep track of the leader as before.
The main consequence of the increased messages is a drop in the UWB range update rate, which is reduced from about 25 Hz with 2 MAVs, to about 16 Hz with 3 MAVs. 

This time, due to the lack of space available, there is no initialization flight procedure to give the EKFs of the followers time to converge.
Instead, the MAVs are placed in starting positions and orientations that roughly match with what EKFs on-board the MAVs are initialized to.
Although this placement is done purely by eye, it is proven to be sufficient to safely start the leader-follower flight.

The leader flies the same trajectory as before.
The first follower follows this trajectory with a $\tau=4$ second delay, and the second follower follows it with an $\tau=8$ second delay.
Once again, 200 seconds of successful flight data is logged and analyzed. 

\begin{figure}[t!] 
\centering
\includegraphics[width=0.9\columnwidth]{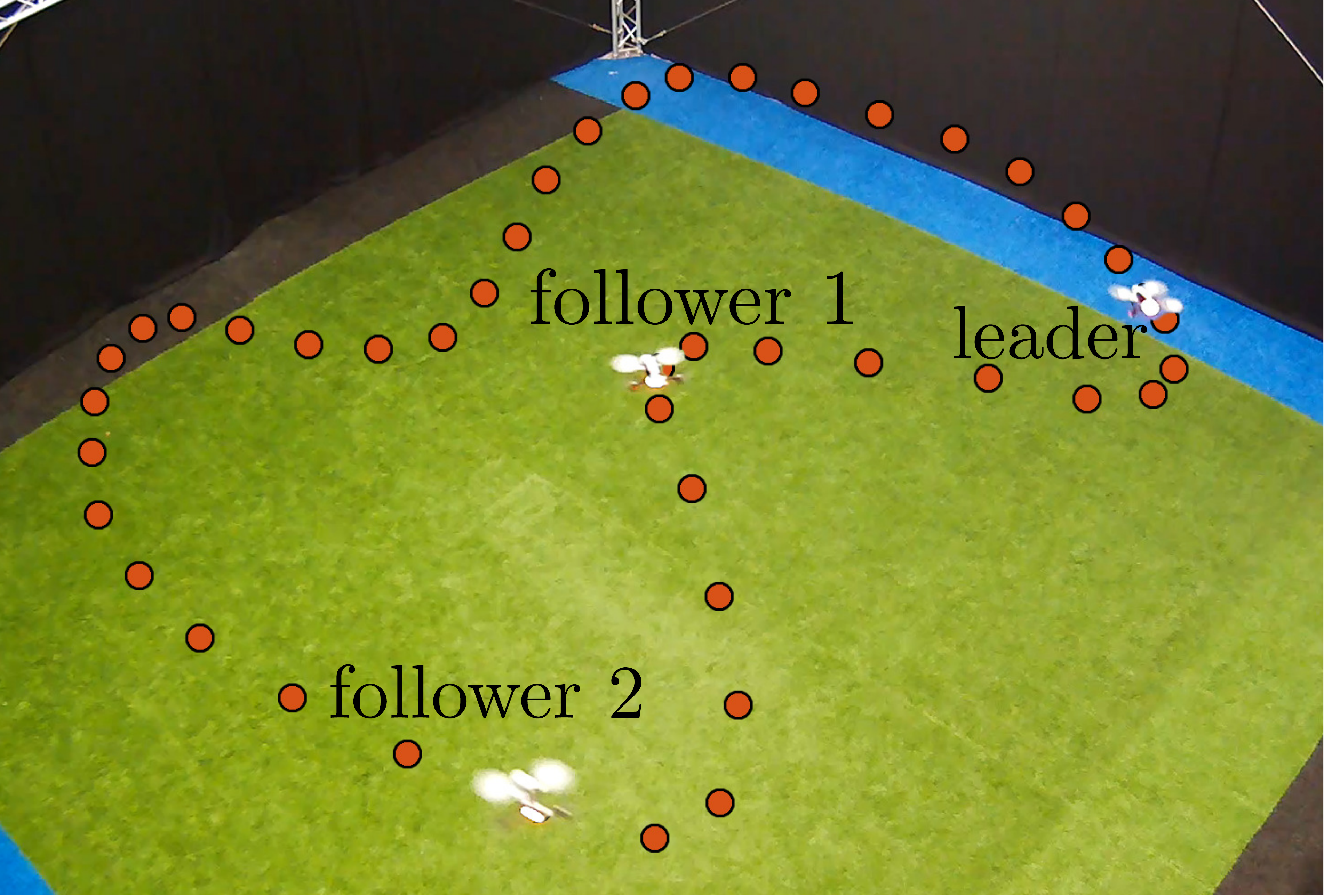}
\caption{Overhead camera image of leader and two followers using MCS height and velocity. In orange is the leader's trajectory marked at 0.5 second intervals.}
\label{fig:multi_NDI_GPS_camera}
\end{figure}

\begin{figure}[t!] 
\centering
\includegraphics[width=0.7\columnwidth]{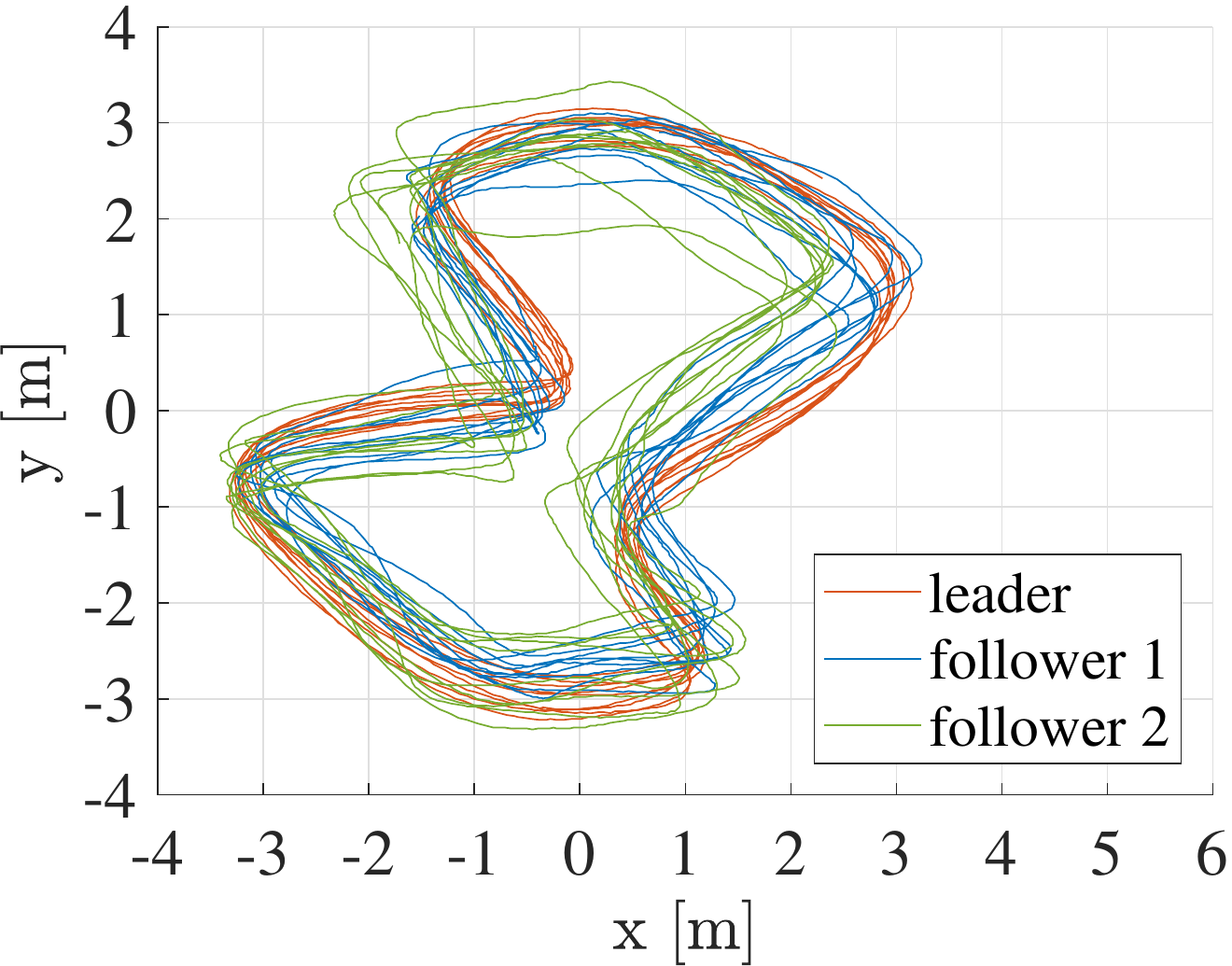}
\caption{Trajectory of leader and two followers using MCS height and velocity}
\label{fig:multi_NDI_traj_GPS}
\end{figure}

\begin{figure}[t!]
\centering
\includegraphics[width=0.7\columnwidth]{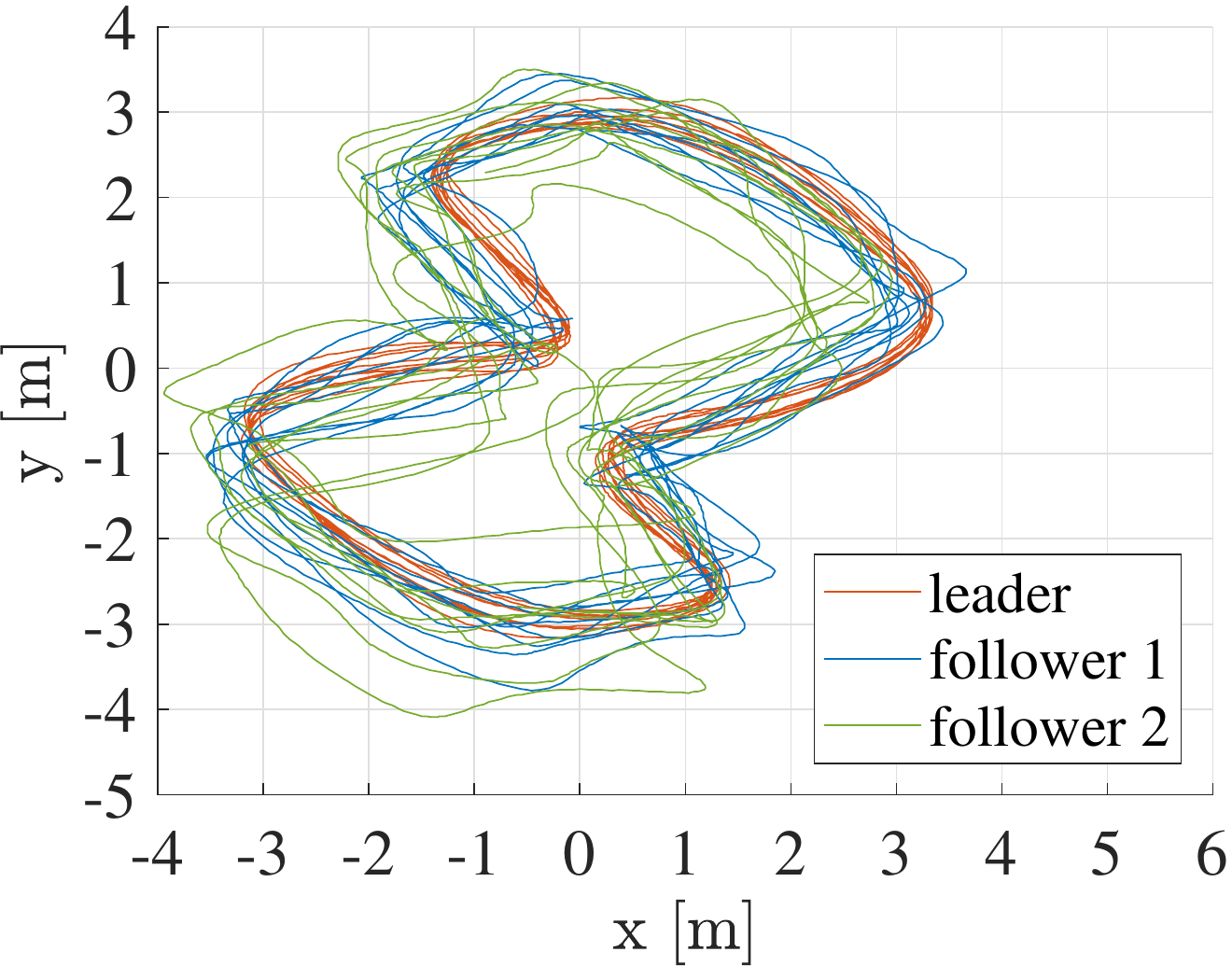}
\caption{Trajectory of leader and two followers using only on-board information}
\label{fig:multi_NDI_traj_onboard}
\end{figure}

An overhead camera image for the flight with MCS height and velocity data is presented in \figref{fig:multi_NDI_GPS_camera}, giving an idea of how the experiment looked like.
The trajectories for this flight are displayed in \figref{fig:multi_NDI_traj_GPS} for the leader and two followers. For the flights with only on-board information, the trajectories are shown in \figref{fig:multi_NDI_traj_onboard}.

As for the case with just one follower, we see that the followers tend to take shortcuts with respect to the leader's trajectory.
Furthermore, the flights using only on-board information are less smooth than those with MCS height and velocity information.
For the flight with MCS data, follower 1 has a MAE for the relative localization error of only 15.8 cm. By comparison, follower 2 has a MAE of 43.9 cm. Furthermore, followers 1 and 2 have MAE for the tracking of 42.9 cm and 70.3 cm, respectively.
The flight with only on-board sensing resulted in a relative localization MAE of 51.8 and 53.6 cm. The tracking MAE this time was 58.6 and 98.4 cm.

\subsection{Comparison of flights}
In this section we present the relative localization and tracking MAE of the various flights that were executed.
We also discuss in more detail the most noteworthy differences between experiments.

\begin{table}[t!]
\centering
\caption{Comparison of mean localization (loc.) errors and mean tracking (track.) errors for all performed experimental flights, both for MCS and fully on-board (on-b.) flights.  }
\label{tab:flightcomp}
\setlength\tabcolsep{2pt}
\begin{tabular}{l|cc|cccc}
\multirow{2}{*}{} & \multicolumn{2}{c|}{1 follower} & \multicolumn{4}{c}{2 followers}  \\
& \multicolumn{1}{c}{MCS} & \multicolumn{1}{c|}{on-b.} & \multicolumn{1}{c}{MCS 1} & \multicolumn{1}{c}{MCS 2} & \multicolumn{1}{c}{on-b. 1} & \multicolumn{1}{c}{on-b. 2} \\ \hline
Loc. error {[}cm{]}   & 18.4 & 22.6 & 15.8 & 43.9 & 51.8 & 53.6 \\
Track. error {[}cm{]} & 46.1 & 50.8 & 42.9 & 70.3 & 58.6 & 98.4 \\ \hline  
\end{tabular}
\end{table}
All the errors are presented in \tabref{tab:flightcomp}.
The first noteworthy observation is the fact that, for the experiment with two followers, the tracking performance of the second follower is worse than for the first follower in both the MCS and fully on-board case.
This is a byproduct of the fact that the proposed leader-follower control method inherently relies on integration of velocity information in time.
As the delay with which the follower must follow the leader increases, so does the period of time over which the follower must integrate its velocity.
This is subject to drift, which shows in the tracking performance.
This effect is more noticeable in the fully on-board case, since the velocity estimates from optical flow methods are less accurate than the ones computed by the MCS.

\begin{figure}[t!]
  \centering
  \begin{subfigure}[t]{39mm}
    \centering
    \includegraphics[width=\textwidth]{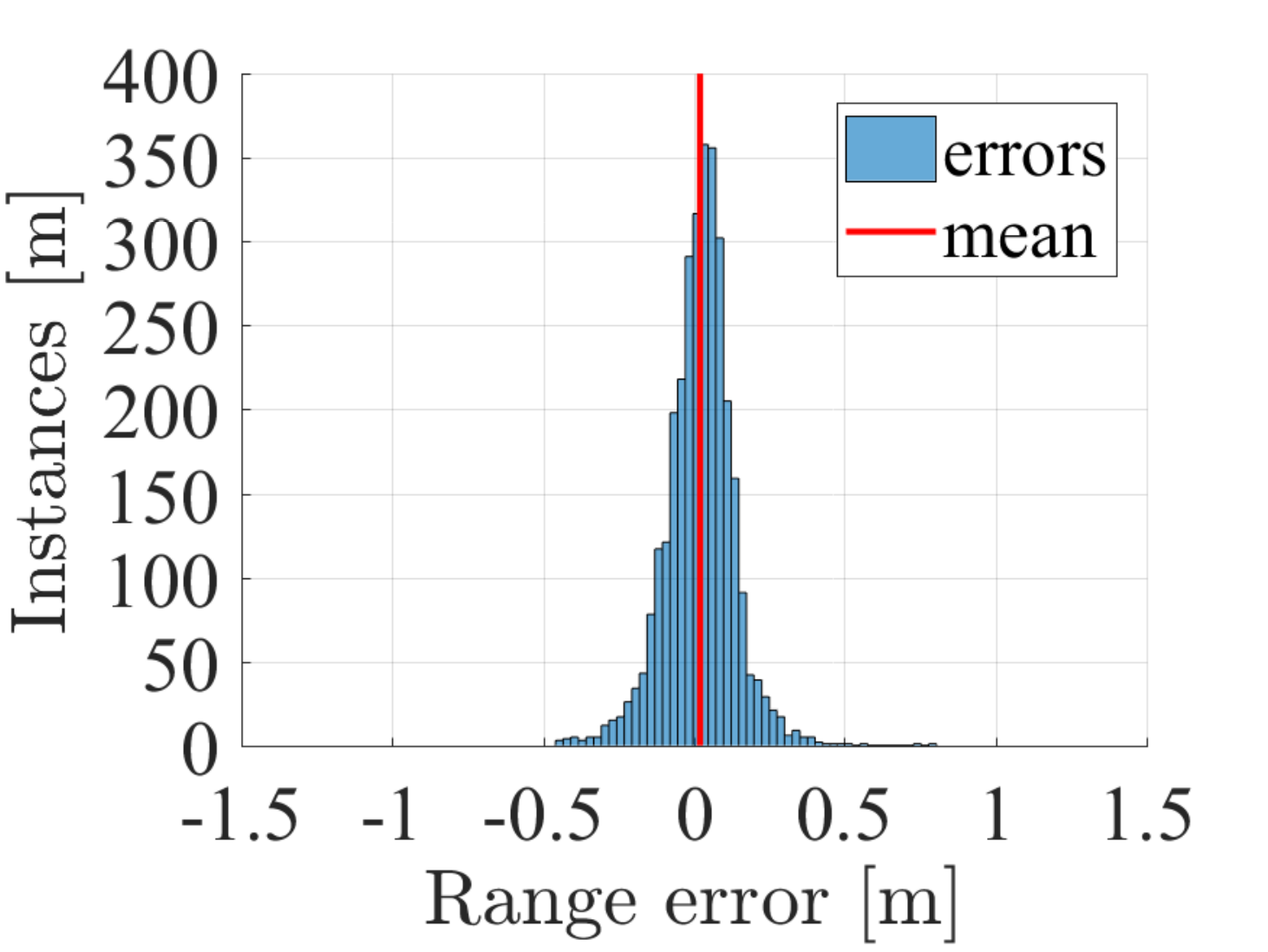}%
    \caption{Follower 1}
    \label{fig:unobshistGPS}
  \end{subfigure}
  ~
  \begin{subfigure}[t]{39mm}
    \centering
    \includegraphics[width=\textwidth]{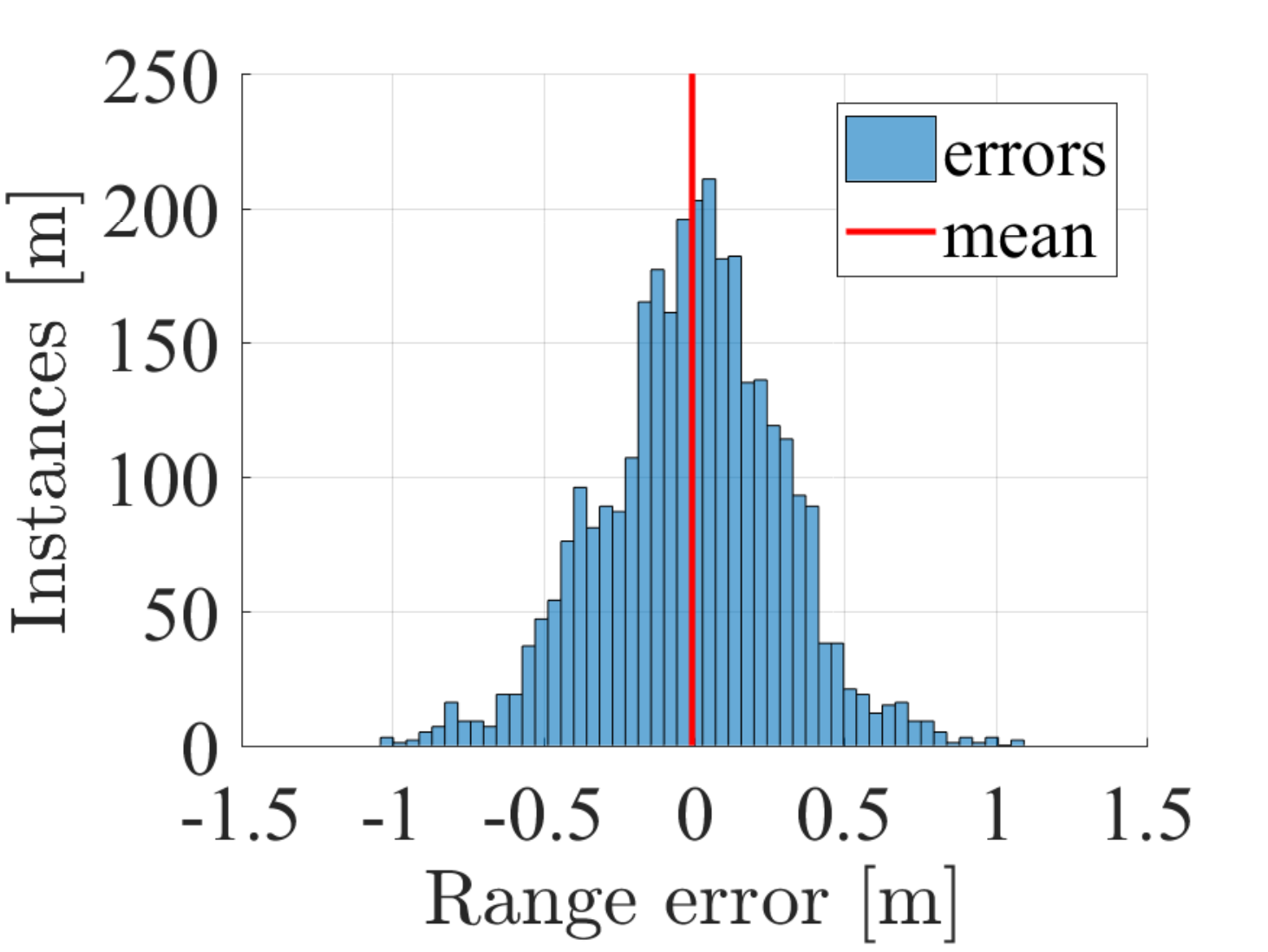}%
    \caption{Follower 2}
\label{fig:unobshistOnboard}
  \end{subfigure}
\caption{Comparison between ranging error distributions for follower 1 and 2 for the flight with MCS height and velocity data.}
\label{fig:rangeerr12_GPS}
  \end{figure}

Another result is that the localization error for follower 2 in the MCS case is higher than for the first follower.
This can be explained, in part, by the fact that follower 2 has a larger mean range with respect to the leader than follower 1 does (4.2 m compared to 2.9 m).
To inspect this deeper, we looked at the logged range between the MAVs.
It was found that follower 2 had substantially larger ranging errors with the leader than follower 1.
This can be appreciated in \figref{fig:rangeerr12_GPS}, where the ranging error distributions are compared.
In both cases, the mean is close to zero, yet the distribution for follower 2 is significantly wider.
At this stage, it is not clear what the primary cause for this drop in range error is.
It shall be studied further in future work when the scalability of the system is addressed in more detail.

A final result that stands out is that both followers 1 and 2 have substantially higher localization errors in the on-board case than was found for the on-board experiment with a single follower.
This result appears to be due to a combination of factors.
The increased communication traffic caused a decrease in the filter update rate and also resulted in an increase in ranging frames dropped.
Follower 2, as mentioned above, showed a worse ranging performance than follower 1.
Follower 1, in turn, had slightly less accurate optical flow velocity estimates than were obtained with the single follower flight (21 cm/s MAE compared to 15 cm/s before) and also slightly higher ranging errors than for the single follower flight (15 cm MAE compared to 8 cm before). All factors combined, both followers suffered a comparable degradation in localization performance.

\section{Discussion}
\label{sec:Discussion}
In this section we revisit the observability analysis from \secref{sec:Obs} with the obtained experimental data.
We also present some remarks on the scalability of this methodology to larger groups of MAVs.

\subsection{Remarks on observability}

In \secref{sec:obsernonorth} showed that for a specific set of velocity, accelerations and relative positions for both MAVs, the system will become unobservable.
To directly integrate the full observability condition in the design of a leader-follower system is difficult due to its high dimensionality.
By having followers fly a delayed version of the leader's trajectory, it is possible to naturally vary the relative positions between leader and follower, as long as the leader's velocity changes in time.
Given the sparsity of unobservable relative positions, we therefore postulated that this control behavior would be sufficient to limit unobservable situations.
Furthermore, even if an unobservable situation were to occur, this would only be for a short period of time, as the relative position continuously changes and the system automatically transitions back to being observable.

Having performed the experiments and collected all the ground truth data, it is now possible to test whether this assumption is valid.
All the parameters needed to evaluate \eqnref{eq:sys2cond} have been logged during the experiments and can be inserted into \eqnref{eq:sys2cond} to check the observability of the relative localization filter in time.
In line with our previous analysis, the measure of observability of the system is represented by the cross product between the left hand side of \eqnref{eq:sys2cond} and the relative position vector $\mathbf{p}$.
Once more, we shall take a threshold of 1, meaning that an observability value between -1 and 1 is considered unobservable.
Although theoretically only a value of 0 would indicate an unobservable system, the higher threshold is chosen to account for noise in the data.



With the chosen threshold, the unobservable data points for the MCS and the on-board flight are 4.76\% and 4.75\% of all the data points, respectively. 
The unobservable points are spread in time, thus giving the system ample observable data in between to recover from the short periods of unobservability.
Furthermore, isolated events of unobservability are not expected to cause issues.
Instead, they can gradually cause an increase in the localization error in time.
This has also been confirmed by the simulations in \secref{sec:Simulation}.

\begin{figure}[t!] %
\centering
\includegraphics[width=0.7\columnwidth]{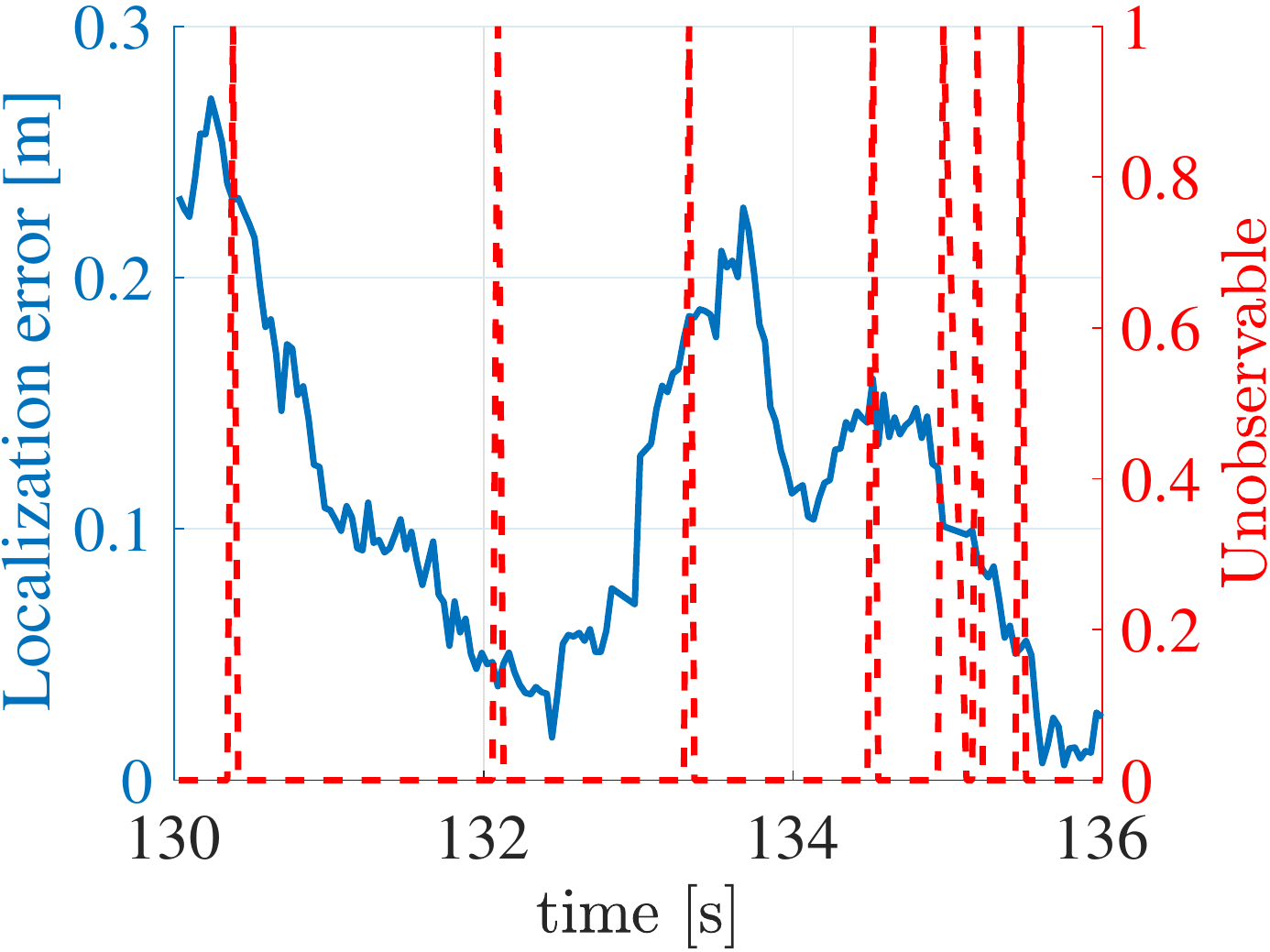}
\caption{Comparison between localization error and the observability of the filter. An unobservable value of '1' means the observability measure is within the threshold of unobservability (between -1 and 1)}
\label{fig:obscorr}
\end{figure}

Further qualitative inspection of the data does not show a correlation between the unobservable regions of the flight and the relative localization error.
To demonstrate this, the localization error is compared to the observability of the filter in \figref{fig:obscorr} for a small segment of the flight with MCS information.
For easier comparison, the observability has been reduced to a binary value, where a value of `1' indicates that the system is within the threshold of unobservability at that time.
It can be seen that there is no apparent correlation between the two parameters.

\subsection{Remarks on scalability}
The experimental results in \secref{sec:Experiment} show that the methods in this paper can successfully scale to two followers that follow a leader in a confined area. Even when full on-board sensing is used by the followers, more than three minutes of successful autonomous flight were demonstrated, with no pilot input.

Despite the successful results, analysis of the data does  show a substantial rise in localization and tracking errors when scaling up to two MAVs. This raises the question of what would happen if even more MAVs are added to the experiment; would this be viable?

One of the results we found is that there is a correlation between the tracking performance of the follower and the time delay with which it follows the leader's trajectory.
The follower that tracked with a time-delay of eight seconds showed consistently larger tracking errors than the followers with four or five second delays.
An alternative solution to the two follower problem is having one follower follow the leader and the other following the first follower.
With such an arrangement, both followers could follow another MAV with the same time delay. This setup has not yet been studied in this work, but could prove to be a better alternative to explore in future research.

Another result we found is that the update rate reduces when flying with two followers instead of one.
It is to be expected that adding more MAVs requires additional data communication, yet a drop from 25 Hz to 16 Hz is quite significant for adding just one more MAV.
The main remark to make here is that this reduction in update rate is very much dependent on the software and hardware used for these experiments. It should be possible to significantly increase the update rate to allow for more MAVs without sacrificing the update rate to a large extent.

As an example, in these experiments we operated the UWB modules on the lowest data rate settings (110 kbps).
Furthermore, every message contains a lengthy preamble of 2048 bits, resulting in substantial protocol overhead for every transmitted message (the actual payload of the UWB messages is less than 200 bits).
This should theoretically help to improve the ranging accuracy, but in practice will most likely not make a big difference at the small inter-MAV ranges occurring in these experiments \citep{DWM1000UM}.
The maximum data rate that the UWB modules support is actually 6.8 Mbps and the preamble can be as short as 64 bits.
These would allow for much higher update rates, even with three or more MAVs. One would, however, need to examine what such a change would have on ranging accuracy and stability.

\section{Conclusion}
\label{sec:Conclusion}


The work in this paper has shown the feasibility of heading-independent range-based relative localization on MAVs.
We now know that removing the dependency on a common heading between MAVs has two main disadvantages: 
the motion of agents must meet more stringent conditions to be observable and 
the relative localization becomes more susceptible to noise on the range measurements. 
The clear advantage, on the other hand, is that the filter is no longer affected by local disturbances in Earth's magnetic field.
As shown by our simulations, small magnetic perturbations can already lead to a large negative impact, showing how a heading-independent method can actually perform better than the heading-dependent method.

The results of our observability analysis have shown that leader-follower flight is a difficult task when using the proposed relative localization method.
Fixed geometry formation flight is not possible. Instead, we developed a method that allows one MAV to follow another MAV's trajectory with a certain time delay.
This approach has been shown to stay sufficiently clear from unobservable conditions, which has allowed us to successfully demonstrate leader-follower flight in practice.

Using only on-board sensory information, one MAV can localize another MAV with a mean error of just 22.6 cm over 200 seconds of leader-follower flight.
This consequently allows the MAV to track another MAV's trajectory with a mean error of 50.8 cm.
The method has been demonstrated to work also with two followers tracking the same leader.


\section{Future work}
\label{sec:futurework}
There are plenty of opportunities to research within the domain of range based relative localization.
Certainly, one such opportunity is the initial convergence behavior of the filter.
The initial estimate of the EKF is important to quickly converge to a correct estimate of the relative location of another MAV.
If the initial condition is too different from the real situation, the filter have difficulties to converge.
One primary problem is that there exist ambiguous states where the EKF can converge to and from which it is difficult to then escape.
In the future, it would thus be interesting to research methods to address this problem.
Examples solutions could be alternative filters (e.g., a particle filter), or running multiple filters in known ambiguous states to identify the correct state more easily.

Furthermore, the current leader-follower implementation flight uses a large amount of past data values, and directly uses state values like the velocities of the two MAVs to implement its control method.
It would be interesting to research other methods of accomplishing this type of leader-follower flight.
For example, it might be possible to perform real time polynomial data fitting on the relative positions of the leader.
The resulting polynomial trajectories could instead be used to obtain the velocities and accelerations through analytical derivations of the polynomials.
This might result in less data that needs to be stored and smoother trajectories.

Finally, considering the hardware used in the experiments, the importance of consistent, high frequency communication and ranging has become apparent.
It would be valuable to further optimize the frequency and consistency with which ranging messages are exchanged. 

\section*{Videos}
Videos of the experiments can be found at:\\
\sloppy
\url{https://www.youtube.com/playlist?list=PL_KSX9GOn2P--aEr4JtFl7SV3LO5QZY4q}

\appendix

\bibliographystyle{spbasic}   
\bibliography{bibliography}   

\end{document}